\documentclass[10pt,twocolumn,letterpaper]{article}

\makeatletter
\@namedef{ver@everyshi.sty}{}
\makeatother
\usepackage{iccv}
\usepackage{times}
\usepackage{epsfig}
\usepackage{graphicx}
\usepackage{amsmath}
\usepackage{amssymb}
\usepackage{caption}
\usepackage{subcaption}

\usepackage[breaklinks=true,bookmarks=false]{hyperref}

\iccvfinalcopy 

\ificcvfinal\fi
\newcommand{\ours}{{VL-MoE}}

\usepackage{multirow}
\usepackage{amsmath}
\usepackage{capt-of}
\usepackage{tabularx}
\usepackage{epsfig}
\usepackage{amssymb}
\usepackage{amsfonts}
\usepackage{booktabs}
\usepackage{scalerel}
\usepackage[inline]{enumitem}
\usepackage{listings}
\usepackage{varwidth}
\usepackage[export]{adjustbox}
\usepackage{tikz}
\usetikzlibrary{tikzmark}

\usepackage{stmaryrd}
\usepackage{bbm}
\usepackage{wrapfig}
\usepackage{pifont}

\newcommand{\sptk}[1]{\texttt{[#1]}}

\definecolor{deepblue}{rgb}{0,0,0.5}
\definecolor{officeblue}{RGB}{0,102,204}
\definecolor{deepred}{rgb}{0.6,0,0}
\definecolor{deepgreen}{rgb}{0,0.5,0}
\definecolor{mybrickred}{RGB}{182,50,28}

\definecolor{fillcolor}{RGB}{216,217,252}

\usepackage{amsmath,amsfonts,bm}

\def\eqref#1{equation~\ref{#1}}

\def\1{\bm{1}}

\def\vzero{{\bm{0}}}

\def\vv{{\bm{v}}}
\def\vw{{\bm{w}}}
\def\vx{{\bm{x}}}

\def\mA{{\bm{A}}}

\def\mI{{\bm{I}}}

\def\mW{{\bm{W}}}
\def\mX{{\bm{X}}}

\DeclareMathAlphabet{\mathsfit}{\encodingdefault}{\sfdefault}{m}{sl}
\SetMathAlphabet{\mathsfit}{bold}{\encodingdefault}{\sfdefault}{bx}{n}

\newcommand{\R}{\mathbb{R}}

\usepackage{amsmath}
\usepackage{amsfonts}
\usepackage{amssymb}
\usepackage{wrapfig}
\usepackage{subcaption}
\usepackage{multirow}
 \usepackage{mathtools} 

\usepackage{verbatim}

\usepackage{anyfontsize}

\usepackage{microtype}
\usepackage{graphicx}
\usepackage{booktabs} 
\usepackage{multirow}
\usepackage{amsmath,amssymb}
\usepackage{booktabs}
\usepackage{caption,subcaption}

\usepackage{xcolor}
\definecolor{mygreen}{HTML}{3cb44b}
\definecolor{skyblue}{HTML}{beffff}
\definecolor{lightgreen}{HTML}{90ee90}

\usepackage{tcolorbox}
\usepackage{enumitem}
\setitemize{itemsep=10pt,topsep=0pt,parsep=0pt,partopsep=0pt}
\usepackage{adjustbox}

\newcommand{\RN}[1]{%
	\textup{\lowercase\expandafter{\it \romannumeral#1}}%
}
\usepackage{tabu}

\usepackage{amsmath}

\newcommand{\beq}{\vspace{0mm}\begin{equation}}
\newcommand{\eeq}{\vspace{0mm}\end{equation}}
\newcommand{\beqs}{\vspace{0mm}\begin{eqnarray}}
\newcommand{\eeqs}{\vspace{0mm}\end{eqnarray}}
\newcommand{\barr}{\begin{array}}
\newcommand{\earr}{\end{array}}

\usepackage{color, colortbl}
\definecolor{Gray}{gray}{0.93}

\usepackage{lipsum}

\newcommand\blfootnote[1]{%
  \begingroup
  \renewcommand\thefootnote{}\footnote{#1}%
  \addtocounter{footnote}{-1}%
  \endgroup
}

\usepackage{xcolor,amsmath}
\usepackage[linesnumbered,ruled,vlined]{algorithm2e}
\DontPrintSemicolon

\usepackage{color, colortbl}

\definecolor{emerald}{rgb}{0.31, 0.78, 0.37}
\definecolor{coralred}{rgb}{1.0, 0.25, 0.25}

\newcommand{\MyColorBox}[2][red]%
{%
    \settowidth{\Width}{#2}%
    \colorbox{#1}%
    {%
        \raisebox{-\DepthReference}%
        {%
                \parbox[b][\HeightReference+\DepthReference][c]{\Width}{\centering#2}%
        }%
    }%
}

\definecolor{codegray}{gray}{0.9}

\SetKwComment{Comment}{\color{green!50!black}\# }{}

\SetKwProg{Function}{def}{:}{}

\SetKwProg{For}{for}{:}{}
\SetKwProg{If}{if}{:}{}

\definecolor{demphcolor}{RGB}{144,144,144}
\newcommand{\demph}[1]{\textcolor{demphcolor}{#1}}
\definecolor{mygray}{gray}{0.4}

\usepackage{pifont}
\newcommand{\cmark}{{\color{blue}\ding{51}}}%
\newcommand{\xmark}{{\color{red}\ding{55}}}%
\usepackage{color, colortbl}
\definecolor{Gray}{gray}{0.93}

\newcommand\vlmo{\textsc{VLMo}}
\newcommand\mome{\textsc{MoME}}
\newcommand\beit{\textsc{BEiT}}
\newcommand\vlbeit{\textsc{BEiT-3}}
\newcommand\vmoe{\textsc{V-MoE}}
\newcommand\limoe{\textsc{LIMoE}}

\newcommand{\tblidx}[1]{{\small \texttt{[#1]}}}

\begin{document}

\title{Scaling Vision-Language Models with Sparse Mixture of Experts}

\author{
    Sheng Shen$^{\dagger\S*}$ \,\,\, Zhewei Yao$^{\ddag*}$\,\, Chunyuan Li$^{\ddag*}$ \,\,  
    {Trevor Darrell}$^\dagger$ \,\, {Kurt Keutzer}$^\dagger$ \,\, {Yuxiong He}$^\ddag$ \\
      $^\dagger$UC Berkeley~~~~~ $^\ddag$Microsoft \\
  \texttt{\small sheng.s@berkeley.edu, \{zheweiyao,chunyl\}@microsoft.com} \\
}

\maketitle
\ificcvfinal\thispagestyle{empty}\fi

\begin{abstract}
   The field of natural language processing (NLP) has made significant strides in recent years, particularly in the development of large-scale vision-language models (VLMs). These models aim to bridge the gap between text and visual information, enabling a more comprehensive understanding of multimedia data. However, as these models become larger and more complex, they also become more challenging to train and deploy. One approach to addressing this challenge is the use of sparsely-gated mixture-of-experts (MoE) techniques, which divide the model into smaller, specialized sub-models that can jointly solve a task. 
In this paper, we explore the effectiveness of MoE in scaling vision-language models, demonstrating its potential to achieve state-of-the-art performance on a range of benchmarks over dense models of equivalent computational cost. 
Our research offers valuable insights into stabilizing the training of MoE models, understanding the impact of MoE on model interpretability, and balancing the trade-offs between compute performance when scaling VLMs. 
We hope our work will inspire further research into the use of MoE for scaling large-scale vision-language models and other multimodal machine learning applications. 
\blfootnote{${*}$~equal contribution; $\S$ work initiated during an internship at Microsoft.}
\end{abstract}
\section{Introduction}
\label{sec:introduction}


%
\begin{figure*}[t!]
	\vspace{-0mm}\centering
	\begin{tabular}{c c c c}
		\hspace{-3mm}
		\includegraphics[height=4.5cm]{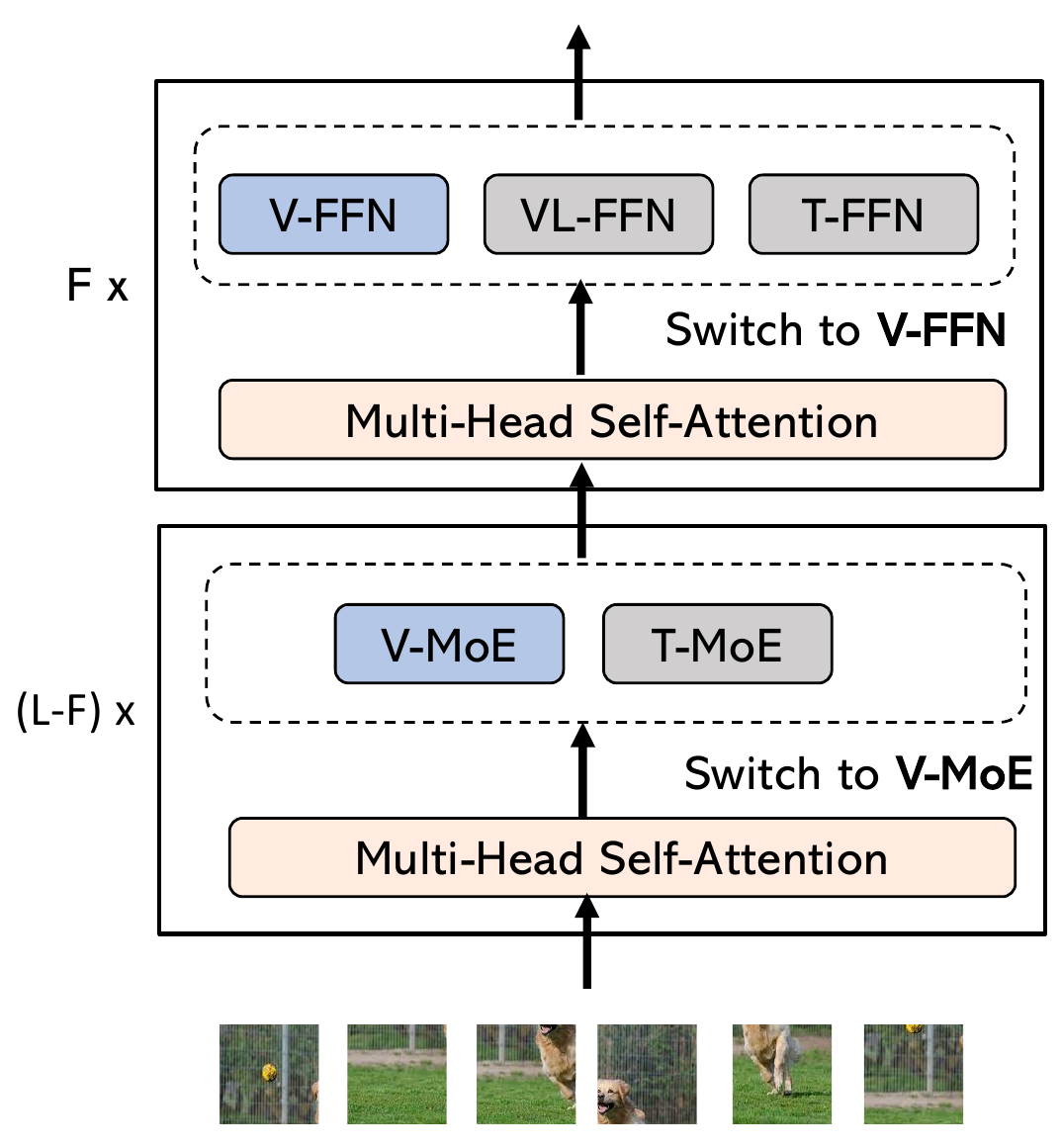}  & 
		\hspace{-4mm}
		\includegraphics[height=4.5cm]{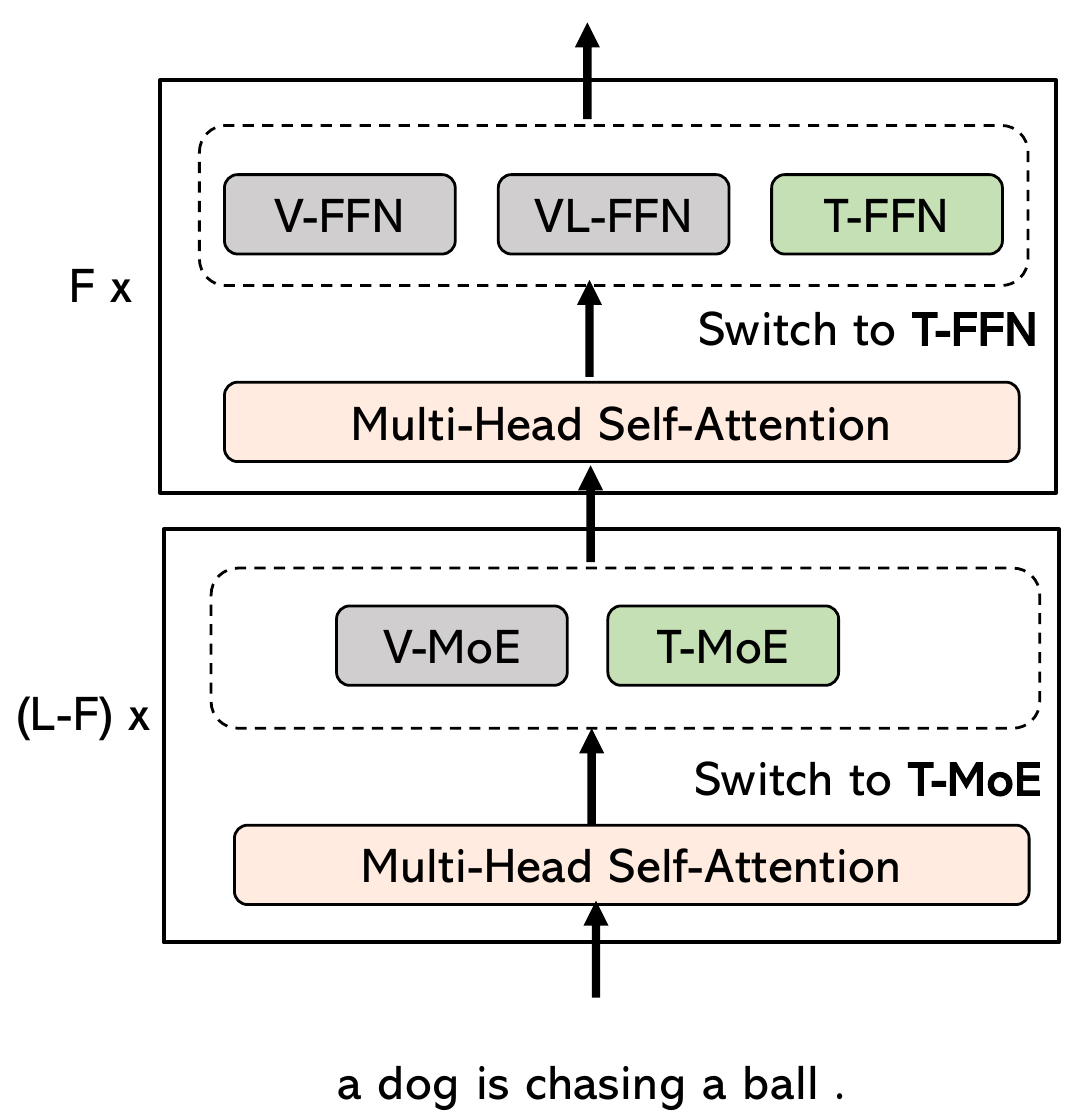}  &
		\hspace{-4mm}
		\includegraphics[height=4.6cm]{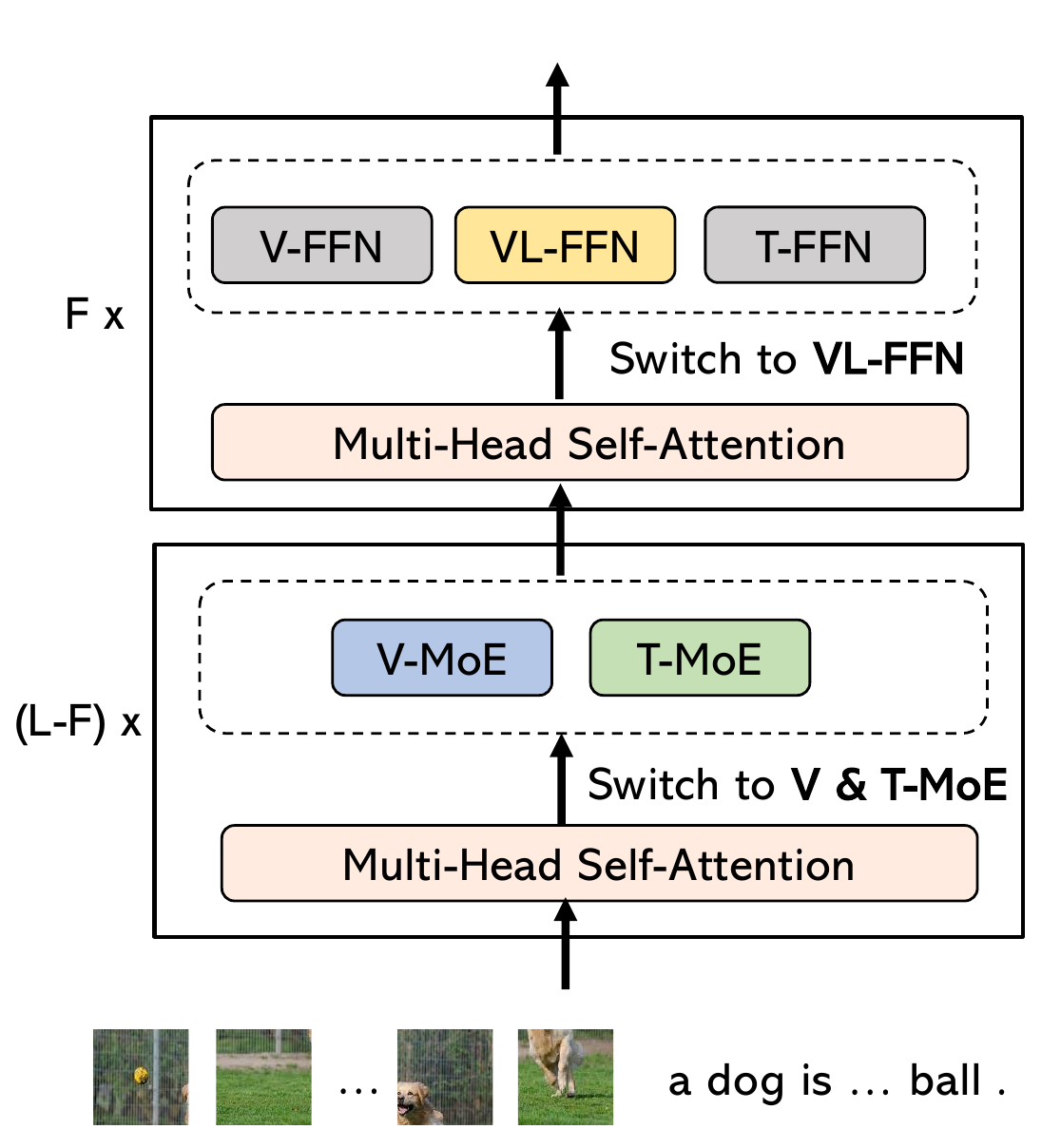}  &
		\hspace{-4mm}
		\includegraphics[height=4.5cm]{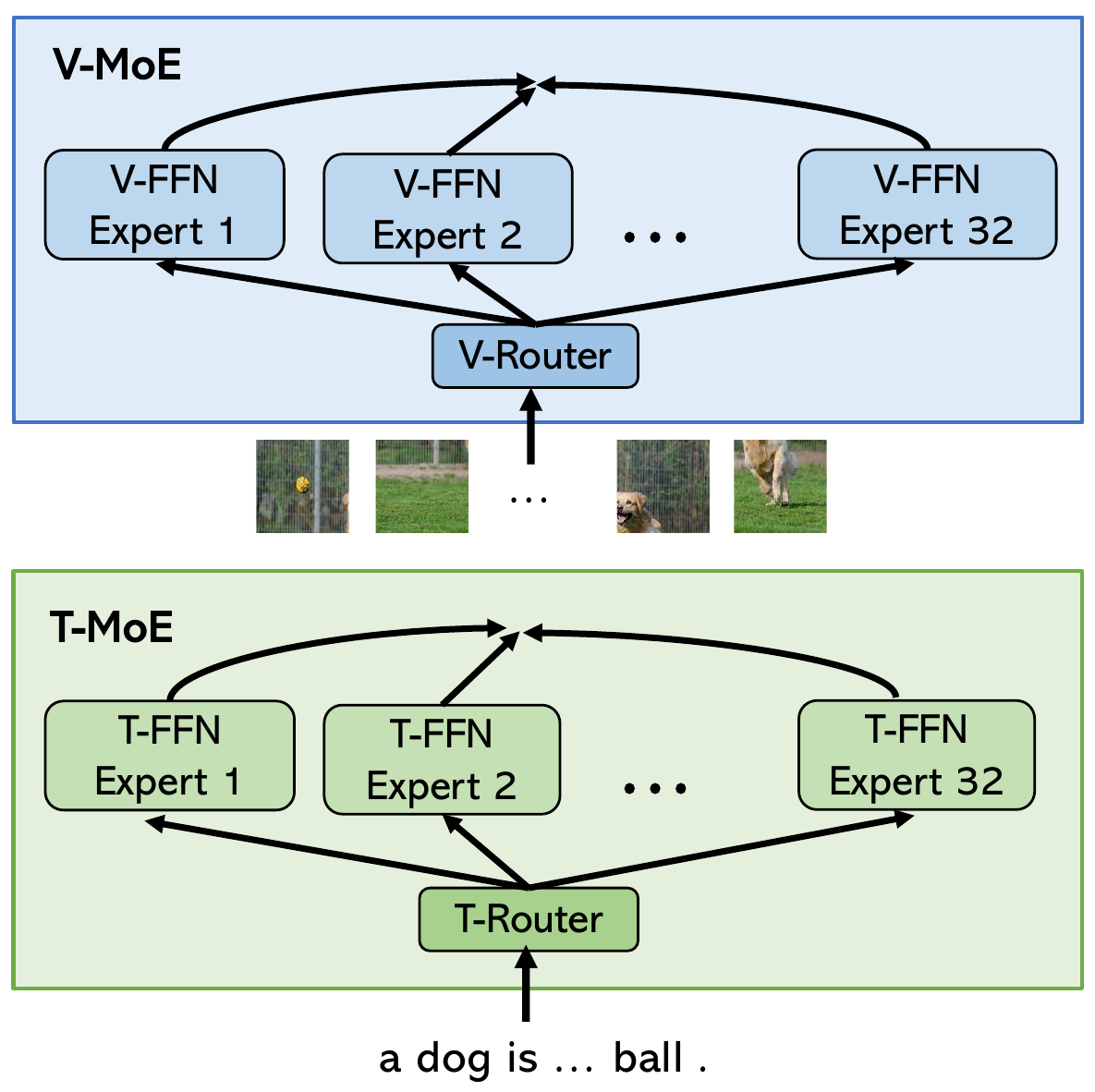} \\
		(a) Encode Image Only \vspace{0mm} & 
		(b) Encode Text Only
 \hspace{-0mm} & 
		(c) Encode Image-Text Pair  \hspace{0mm} & 
		(d) V-MoE \& T-MoE\hspace{-0mm} \\ 
	\end{tabular}
	\vspace{-3mm}
	\caption{The encoding process of \ours{} for various modality inputs, for which gray and colored blocks indicate non-activated and activated modules, respectively. 
    (a) For image input only, the encoding process switches to V-MoE or V-FFN  
    (b) For text input only, the encoding process switches T-MoE or T-FFN. 
    (c) For image-Text Pair input, the encoding process switches, V-MoE \& T-MoE and VL-FFN.
    (d) For the early layers, we scale the V-FFN and T-FFN with Sparse Mixture-of-Experts as V-MoE and T-MoE, respectively. VL-MoE will utilize conditional computation to allocate tokens in a modality-specific fashion. V/T-MoE converts multiple V/T-FFNs as experts, where the image/text input will be conditionally routed by V/T-Router Network.
	 }
	\vspace{-4mm}
	\label{fig:overview}
\end{figure*}

The ability to understand and generate natural language from visual information is a critical component of many real-world applications, including visual question answering (VQA), visual reasoning, and multimodal information retrieval. 
In recent years, the success of deep learning in natural language processing (NLP) has led to the development of large-scale vision-language models (VLMs)~\cite{lxmert,uniter,unimo,villa,vilt,flamingo,simvlm,vilclip,albef,klite,align,blip,coca} that leverage powerful neural network architectures to encode and decode multimodal information. 
However, state-of-the-art vision-language models like Flamingo-80B~\cite{flamingo}, \vlbeit-1.9B~\cite{vlbeit}, and PaLI-17B~\cite{pali} can be computationally expensive and difficult to train, which has motivated researchers to explore ways of improving their efficiency and effectiveness.

Recently, sparsely activated \textit{Mixture of Experts (MoE)} models have been successfully employed to scale both vision~\cite{vmoe,mixermoe,limoe} and text models~\cite{moe,gshard,stmoe,glam}. These models are motivated by the need to increase model parameters while controlling compute costs. 
In addition, these models provide other advantages, including sparsity that can mitigate catastrophic forgetting in continual learningg~\cite{routingcontinual,sparseupcycle}, and an inductive bias that can enhance performance in multitask learningg~\cite{mmoe,taskmoe,zeromoe}. 
Overall, the use of MoEs has proven to be a promising strategy for scaling deep learning models across various domains.

Building on the success of MoEs in individual domains and applying the intuition that sparse models may better handle different tasks versus dense counterparts, we investigate the potential of MoEs for vision-language modeling. 
To this end, we take the first step in this direction and explore models that can process both images and text for vision-language tasks. 
One similar effort has been studied in \limoe~\cite{limoe}, where the authors proposed a modal-agnostic CLIP-style~\cite{clip} multimodal MoEs architecture, but their focus is mainly on the contrastive pre-training objective and vision-only downstream tasks. There are two limitations in this setting: (1) The increasing model capacity of MoEs under the the simple contrastive objective can easily lead to over-fitting issues. (2) The vision-only benchmarking does not reveal the full power of scaling up multimodal models.
Alternatively, our goal is to demonstrate the effectiveness of MoEs under generative modeling for vision-language tasks and provide a more comprehensive foundation for future research in this area.

Specifically, we propose a novel VLM architecture that employs MoE to scale both the text-based and vision-based feed-forward networks (T-FFN and V-FFN, respectively) in a unified framework. 
Our approach divides the model into multiple sub-models, each of which is responsible for processing a modal-specific subset of the input data. 
The text and vision input representations are then aligned via three mask data modeling objectives~\cite{vlbeit}. 

We train a range of \ours{} models and evaluate the model on vision-language classification, vision-language retrieval, vision-only and language-only tasks, 
Our experiments demonstrate that MoE can significantly improve the efficiency and effectiveness of VLMs, enabling them to handle large-scale, real-world multimedia data. 
We scale \textsc{base}-size model up to a $2$B parameter \ours$_\textsc{base/32E}$, which only applies 180M parameters per token and achieves competitive performance with dense models that make use of similar or more pre-training image-text pair data and apply 3-4$\times$ more parameters per token. 

In summary, our contributions are as follows:

\begin{itemize}
    \item We propose \ours{}, the first large-scale generative MoEs multimodal models for vision/langauge-only, as well as vision-and-language tasks.
    \item We explore various scaling strategies, including increasing dense model size, increasing expert numbers, and scaling either T-FFN or V-FFN alone, to investigate the trade-offs between model complexity and performance on various downstream tasks. 
    \item We present ablations to understand
     \ours{} model’s behavior, interpretability, and our design choices.
\end{itemize}

\section{Related Work}
\label{sec:related_work}

\paragraph{Vision-Language Modeling.} 
Vision-language pretraining~\cite{lxmert,vilbert,vl-bert,vinvl,clip,oscar,vilt,albef,simvlm,vlmo,ofa,flamingo,coca,vlbeit,blip,pali,clip,align,vilclip,klite,florence,flava,liu2023learning} involves developing model architecture and pretraining objectives to learn effective multimodal representations from large-scale image-text pairs. Two main approaches are encoding distinct modalities separately with different encoders. 

For model architecture, there are two main designs. 
The first design, utilized by models such as~\cite{clip,align,florence} separately encodes each modality with different encoders. While this approach performs well for image-text retrieval tasks, it struggles with complex vision-language tasks like visual reasoning. 
The second design, employed by models like ~\cite{lxmert,albef,vilbert,visualbert,vilt,pali,flamingo}, uses a complex fusion module with cross-modal attention to combine modalities and learn powerful multimodal representations. However, this design sacrifices efficiency for improved performance. 
Recently, a new design has emerged with the \mome{} Transformer used in both \vlmo{} and \vlbeit{}. 
This design unifies the dual-encoder and fusion-encoder models by introducing a mixture-of-modality-experts technique. 
With \mome{}, various modalities are encoded within a shared Transformer block, allowing for improved scalability and achieving state-of-the-art performance on vision-language tasks. There is an increasing interest to grow the VL model capacity with an affordable compute budget, including MoE~\cite{limoe} and the injection of new trainable modules on pre-trained models~\cite{flamingo,klite,liu2023learning,li2023gligen,li2023blip2,koh2023grounding}; the former remains less studied.

For pretraining objectives, multiple cross-modal pretraining objectives have been studied. They can be categorized into two classes: (1) {\it Discriminative modeling}, including image-text contrastive learning~\cite{clip,align}, image-text matching~\cite{lxmert,vilt,albef,vlmo} and word-patch/region alignment~\cite{uniter,vilt}; (2) {\it  Generative modeling}, including masked language modeling~\cite{lxmert,vl-bert,vilt} or prefix language modeling~\cite{simvlm}, masked region modeling~\cite{lxmert}, multimodal prefix language modeling~\cite{simvlm}.
Recently, 
\vlbeit{} shows strong scaling results by unifying the generative multimodal pretraining objective with masked data modeling, which comprises masked image modeling and masked language modeling on the monomodal encoders and masked multimodal modeling on the multimodal encoder.

In this paper, we perform MoE study, by adopting the \mome{} Transformer as the backbone dense network and generative  (masked data)  modeling as pretraining objectives given its simplicity and scaling ability.

\paragraph{Sparse Mixture of Experts models.} 
We build upon the concept of deep sparse MoEs, which have been studied independently in both Computer Vision~\cite{vmoe,mixermoe,limoe} and Natural Language Processing~\cite{vmoe,mixermoe,limoe,moe,gshard,switchtransformer,glam,stmoe,unifiedscaling,expertchoice,sparseupcycle,taskmoe} in the context of conditional computation. 
The goal of conditional computation is to increase the number of model parameters without a proportional increase in computational cost, which is achieved by selectively activating only relevant parts of the model based on input-dependent factors~\cite{bengio2013deep,chen1999improved,davis2013low}. 
MoE models use a learned gating mechanism that activates only a subset of $k$ experts out of $E\gg k$ for a given input, allowing an input to select either all experts~\cite{eigen2013learning} or only a sparse mixture thereof, as in recent massive language models~\cite{switchtransformer,glam}. 
While many works aim to improve the gating mechanism itself~\cite{hazimeh2021dselectk,lewis2021base,roller2021hash,expertchoice}, MoE models have also been studied for multitask learning~\cite{hazimeh2021dselectk,taskmoe} with per-task routers~\cite{mmoe}, although a shared pool of experts is typically used.

MoE models have been explored for multimodal learning as well, with \limoe~\cite{limoe} being most relevant to our work. 
However, their MoE scaling considers the CLIP-style contrast as the pre-training objective, and vision/retrieval tasks as the downstream evaluation. 
To the best of our knowledge, the proposed \ours{} is the first the MoE scaling study to consider the generative modeling objective in the VL pre-training, and we valuate its scaling performance in a more comprehensive manner, including vision/language-only, as well as vision-and-language tasks.
\section{Method}
\label{sec:method}


\begin{figure*}[t!]
	\vspace{-0mm}\centering
	\begin{tabular}{c c c c}
		&
		\hspace{-48mm}
  \includegraphics[height=0.54cm]{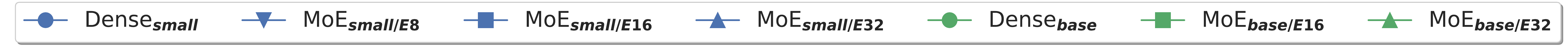}
  \hspace{-85mm}
  & & 
		 
		 \\
		\hspace{-3mm}
		\includegraphics[height=3.0cm]{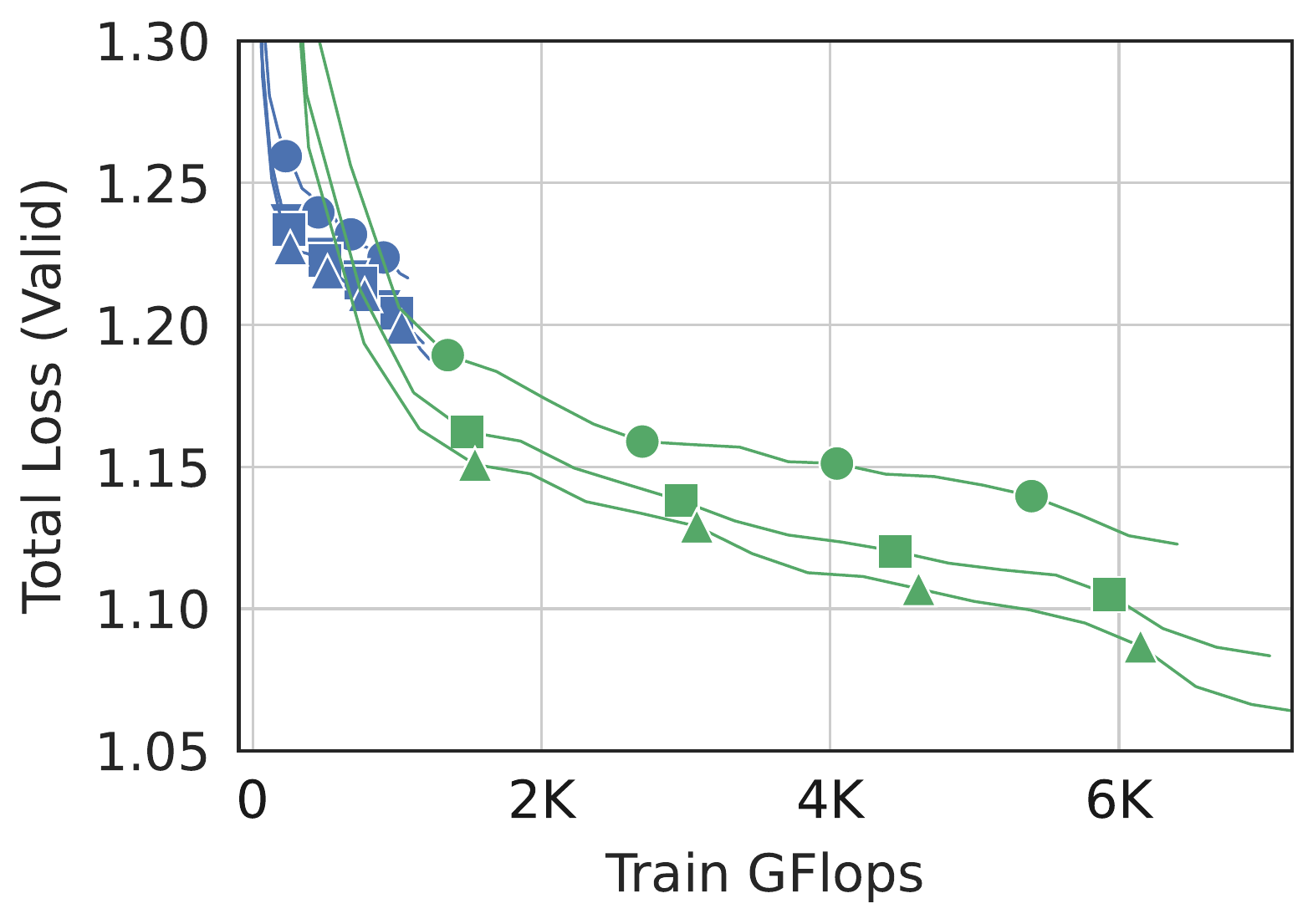}
		&
  \hspace{-6mm}
		\includegraphics[height=3.0cm]{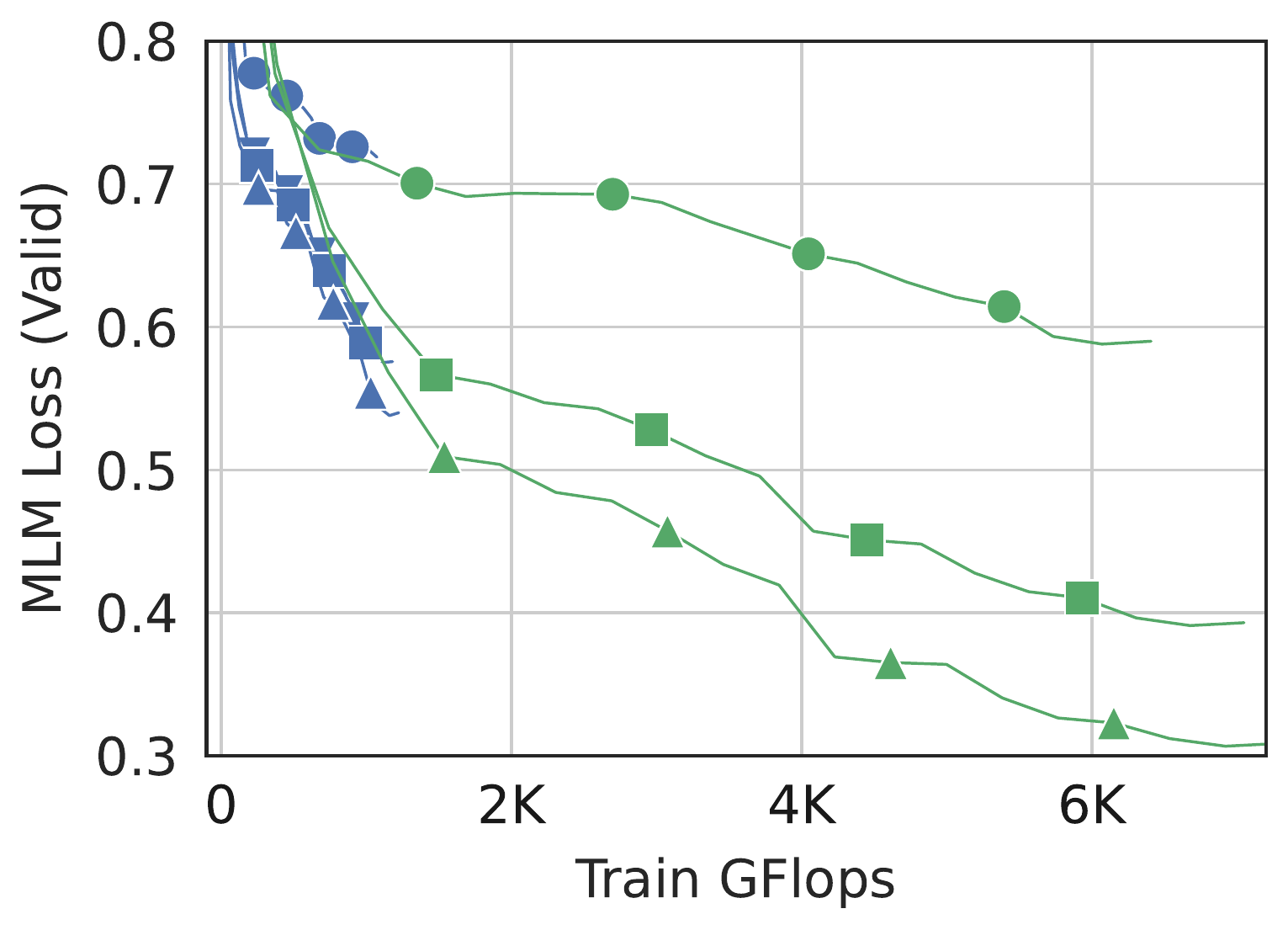}
 		&
  \hspace{-6mm}
		\includegraphics[height=3.0cm]{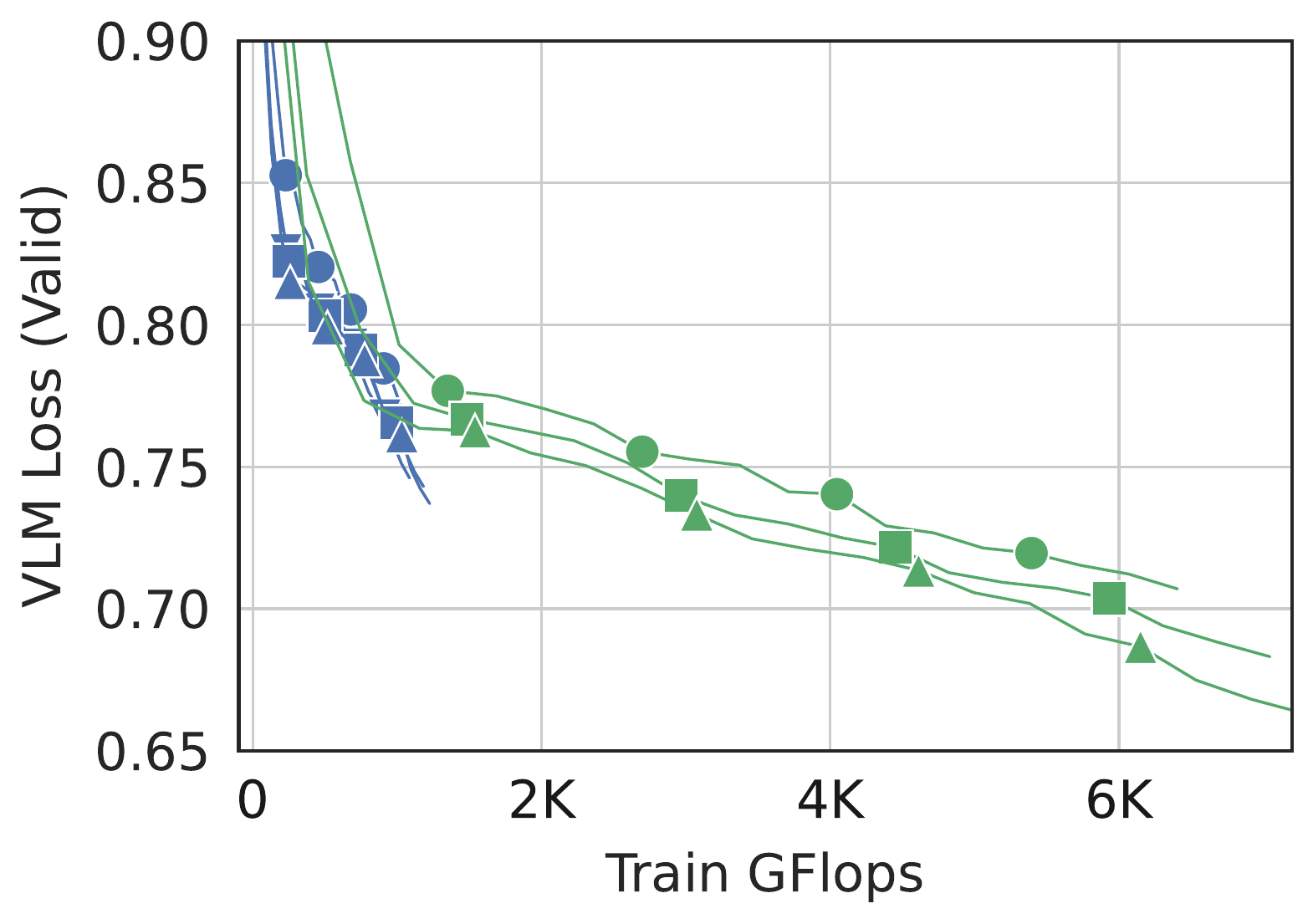} 
            &
            \hspace{-5mm}
  		\includegraphics[height=3.0cm]{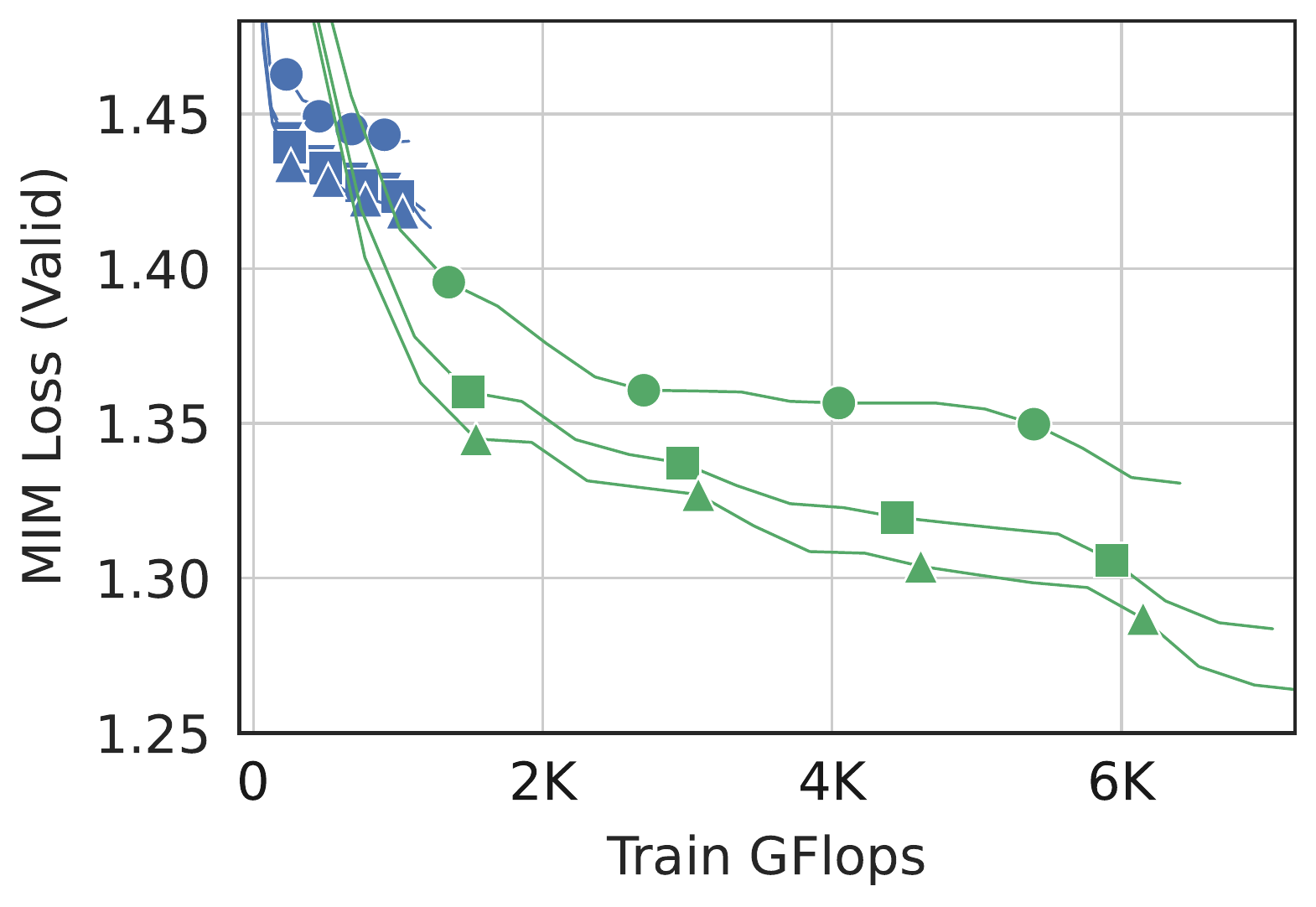} \\
		(a) Total Validation Loss \vspace{2mm} 
		&
		(b) MLM Validation Loss \vspace{2mm} 
		&
  		(c) VLM Validation Loss \vspace{2mm} 
		&
		(d) MIM Validation Loss \hspace{-0mm}  \\
	\end{tabular}
	\vspace{-6mm}
	\caption{Effect of \ours{} scaling on three mask language modeling (MLM), mask image modeling (MIM), and masked vision-language modeling (VLM) pre-training tasks across training flops. 
	 }
	\vspace{-2mm}
	\label{fig:train_loss_beit}
\end{figure*}

We first describe the masked data modeling pretraining objectives. 
We next discuss MoEs, sparse MoEs and present how we apply sparse MoEs methodology to vision-language models, before explaining our design choices for the routing algorithm and the implementation of \ours.

\subsection{Vision-Language Masked Data Modeling}
We utilized a unified masked data modeling objective~\cite{vlbeit} to pretrain \ours{} on monomodal (i.e., images and texts) and multimodal data (i.e., image-text pairs). 
This approach has been demonstrated to be scaling-friendly with small batch-sizes. 
Our pretraining process involved masked image modeling on monomodal image data, masked language modeling on monomodal text data, and masked vision-language modeling on multimodal image-text pairs. 

\paragraph{Masked Language Modeling}

We use masked language modeling~(MLM) to learn language representations from large-scale text-only data.
For MLM, 15\% of tokens in monomodal text data are randomly masked, and the model is trained to recover the masked tokens from the corrupted input text. 
Masked tokens are replaced by a \texttt{[MASK]} token 80\% of the time, a random token 10\% of the time, and kept the original tokens 10\% of the time, following BERT~\cite{bert}. 

\paragraph{Masked Image Modeling}

In addition to masked language modeling, \ours{} uses masked image modeling (MIM) to learn vision representations from large-scale image data.
For MIM, block-wise masking is applied to 40\% of image patches, and the pretraining objective is to reconstruct the discrete visual tokens of masked patches, following BEiT~\cite{beit}. The image tokenizer of \beit{}v2~\cite{beitv2} is used to obtain the discrete tokens as the reconstructed targets.

\paragraph{Masked Vision-Language Modeling}

To learn aligned vision-language representation, we use masked vision-language modeling (VLM), which extends masked language modeling and masked image modeling to multimodal data.
The task aims at recovering masked image patches and text tokens based on visual and linguistic clues.
In VLM, text tokens (with 50\% mask ratio) are randomly masked as in MLM, and the model is trained to recover the masked text tokens based on the joint image-text representations. 
Image patches are also masked with the same ratio as in MIM, and the corresponding visual tokens are predicted based on the image-text pair. 
The VLM task further encourages the model to learn alignments between image and text pairs.

\subsection{\ours{} Architecture}

\paragraph{Input Representation.}

To obtain text representations, the input text is tokenized and projected onto word embeddings ($\{\vw_i\}_{i=1}^{M}$), where $M$ is the length of the tokenized text sequence.
Two special tokens, a start-of-sequence token (\sptk{T\_CLS}) and a special boundary token (\sptk{T\_SEP}), are added to the sequence. 
Text representations are obtained by summing the word embeddings and text position embeddings, resulting in $\mathbf{H}^{w} = [ \vw_{\sptk{T\_CLS}} , \vw_{1} , \dots , \vw_{M} , \vw_{\sptk{T\_SEP}} ] + \mathbf{T}_{pos}$.

For image representations, the input 2D image $\vv \in \R^{H \times W \times C}$ is split and reshaped into $N={HW}/{P^2}$ patches $\vv^{p} \in \R^{N \times (P^2 C)}$, where $C$ is the number of channels, $(H, W)$ is height and width of the input image, and $P$ is the patch size. 
These patches are then flattened into vectors and linearly projected to obtain patch embeddings following vision Transformers~\cite{vit,deit,beit}. 
We prepend a learnable special token \sptk{I\_CLS} to the sequence. 
The resulting image input representations are given by $\mathbf{H}^{v} = [ \vv_{\sptk{I\_CLS}} , \vv_{1} , \dots , \vv_{N} ] + \mathbf{V}_{pos}$, where $\mathbf{H}^{v} \in \R^{(N+1) \times D}$, $\mathbf{V} \in \R^{(P^2 C) \times D}$ is a linear projection, $\mathbf{V}_{pos} \in \R^{(N+1) \times D}$ are learnable 1D position embeddings. 

To form image-text input representations, we concatenate image and text input vectors, resulting in $\mathbf{H}_0^{vl} = [\mathbf{H}_0^{w} ; \mathbf{H}_0^{v}]$.

\paragraph{Backbone Network.}

The dense backbone network of \ours{} is a shared multimodal Transformer, illustrated in Figure~\ref{fig:overview}. 
To encode different modalities, we utilize a mixture-of-modality-experts (\mome{}) Transformer\cite{vlmo,vlbeit}, which takes image and text representations of monomodal data, as well as representations of image-text pairs as input. 
The \mome{} Transformer comprises multiple layers of blocks, each consisting of a multi-head self-attention layer and a feed-forward expert layer. 
While the self-attention module is shared across modalities, each feed-forward expert layer contains a pool of modality-specific experts (V-FFN, T-FFN, or VL-FFN) that act as a substitute for the feed-forward network in standard Transformers. 
This allows for hard routing over the pool of feed-forward networks based on the modality of the input tokens. 

\paragraph{Conditional Computation with MoEs.}

The concept of conditional computation involves selectively activating different parts of a neural network based on the input~\cite{bengio2013deep}. One specific approach is to use a mixture-of-experts (MoE) model, where different ``experts" handle different regions of the input space~\cite{jacobs1991adaptive}.
In this paper, we adopt the MoE layer proposed in~\cite{moe}, which consists of $E$ experts and is defined as $\texttt{MoE}(\vx)= \sum_{i=1}^E g(\vx)_i \ e_i(\vx)$. 
Here, $\vx$ is the input to the layer, $e_i: \mathbb{R}^D\mapsto\mathbb{R}^D$ is the function computed by expert $i$, and $g: \mathbb{R}^D\mapsto\mathbb{R}^E$ is the ``routing" function that determines the input-dependent weights for the experts. Both $e_i$ and $g$ are implemented as neural networks. 
Although this formulation still involves a dense network, it can be made sparse by restricting $g$ to assign only $k \ll E$ non-zero weights, thereby eliminating the computation of unused experts. This approach allows for super-linear scaling of the number of model parameters in both training and inference.



\paragraph{\ours{}.}



%
We apply sparse MoE to vision-language models in the context of the \mome{}. 
As illustrated in Figure~\ref{fig:overview}, inputs from different modalities are routed to V-FFN and T-FFN in the first ($L-F$) layers, and V-FFN, T-FFN, or VL-FFN in the last $F$ layers. To avoid instability due to modality input imbalance when applying MoEs to modal-agnostic VL-modules in \vmoe~\cite{vmoe}, we only use MoE for V-FFN and T-FFN in the first ($L-F$) layers. 
V-FFN and T-FFN have two layers and a GeLU~\cite{gelu} non-linearity: V/T-$\texttt{FFN}(\vx)= \mathbf{W}_2 \ \sigma_\text{gelu}(\mathbf{W}_1 \vx)$. 
For \ours{}, we replace a subset of V-FFN and T-FFN with V-MoE and T-MoE layers, where each expert is an FFN with the same architecture $e_i(\mathbf{x})=\texttt{FFN}_{\theta_i}(\vx)$ but different weights $\theta_i = (\mathbf{W}_1^i, \mathbf{W}_2^i)$. 
This design pattern is similar to that of GShard~\cite{gshard} and \vmoe~\cite{vmoe} models. In V-MoE and T-MoE layers, each token $\vx \in\mathbb{R}^D$ is processed sparsely by $k$ out of $E$ available experts. To select which one, a lightweight V/T-Router predicts gating weights \textit{per token}: $g(\vx) = \texttt{softmax}(\mathbf{W}_g\vx) \in \R^E$, where $\mathbf{W}_g \in \R^{D\times E}$ is learned. 
The $k$ activated experts' outputs are combined linearly according to the gating weights: $\texttt{MoE}(\vx) = \sum_{e=1}^k g(\vx)_e \cdot \texttt{FFN}_e(\vx)$. 

\begin{figure*}[t]
    \centering
    \includegraphics[width=0.9\textwidth]{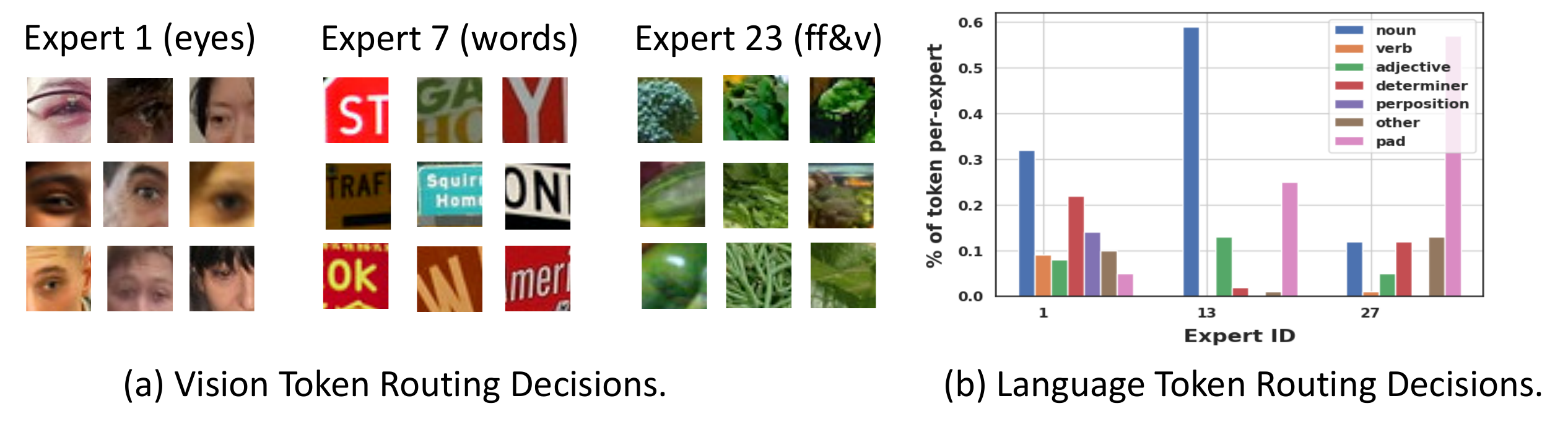}
    \vspace{-3mm}
    \caption{Token routing decisions on COCO. Examples of vision tokens routing decisions and breakdown of language token routing decisions at the V/T-MoE layer placed in the $6$-th encoder block –i.e. middle of the network– for \ours{}$_\textsc{base/32E}$.}
    \label{fig:overall_token_route}
\end{figure*}

To ensure computational efficiency and implementation constraints, each expert in \ours{} has a fixed buffer capacity, which determines the maximum number of tokens it can process. 
The assumption is that tokens are approximately balanced across experts. 
In case the capacity is exceeded, some tokens are not processed by the expert and are dropped, leading to a decrease in the success rate. 
This rate is a vital indicator of balanced routing and training stability. 
To mitigate this problem, we employ Batch Priority Routing (BPR)~\cite{vmoe,limoe}, which selectively skips tokens based on their routing weights. 
BPR prioritizes tokens with larger routing weights, as they are deemed more informative. 
Our results show that BPR is crucial for stable training of \ours{}. 
We further analyze token routing decisions in Section~\ref{sec:ablation} and dropped tokens in Appendix.


\section{Experiment}
\label{sec:experiment}

\subsection{Pretraining Setup}
\paragraph{Pretraining Data.} Our pretraining process uses both monomodal and multimodal data. The monomodal data comprises ImageNet-22K for images and English Wikipedia and BookCorpus~\cite{bookcorpus} for text. The multimodal data combines four datasets of image-text pairs: Conceptual Captions~\cite{gcc}, SBU Captions~\cite{sbu}, COCO~\cite{coco}, and Visual Genome~\cite{vg}, containing a total of $4$ million images and $10$ million image-text pairs.

\paragraph{Pretraining Setting.} 
For the base-size model, we employ a $12$-layer Transformer network with $768$ hidden size and $12$ attention heads, following VIT~\cite{vit}, BEiT~\cite{beit}, and VLMO~\cite{vlmo}. 
The use of VL-FFN starts at $10$th layer. 
The small-size model is an $8$-layer Transformer network with $384$ hidden size and $6$ attention heads, where VL-FFN is used in $8$th layer. 
We randomly initialize the model parameters using the method described in BEiT~\cite{beit}. The image resolution is set to $224\times224$, and the patch size is $16\times16$. 
The maximum sequence length for text is $128$. 
We use a batch size of $6,144$ and train the model from scratch for $200$k steps, which is equivalent to $40$ epochs of the image-text pairs. Each batch contains $2,048$ images, $2,048$ texts, and $2,048$ image-text pairs. 
We perform image augmentation using random resized cropping, horizontal flipping, and color jittering, following the same method as BEiT~\cite{beit}. The text data is tokenized using a SentencePiece~\cite{sentencepiece} tokenizer with a vocabulary size of 64k. We use the Adam optimizer~\cite{adam}  with $\beta_1=0.9$ and $\beta_2=0.999$ to optimize the model. The peak learning rate is 2e-3, and we use linear warmup for the first $10,000$ steps and cosine learning rate decay. The weight decay is $0.05$, and we disable dropout and use stochastic depth~\cite{drop_path} with a rate of $0.1$. 
The three pretrain losses are equally weighted as in \vlbeit{}~\cite{vlbeit}.

\paragraph{MoE Setting.} 
For the default setting of MoEs in \ours{}$_\textsc{base/32E}$, we use $E=32$ experts for T-FFN and V-FFN, respectively. 
All \ours{}s activate $k=1$ expert per token, similar to Switch Transformer~\cite{switchtransformer} and LIMoE~\cite{limoe}. 
We replace every second dense T-FFN or V-FFN sublayer with MoE sublayer following GShard~\cite{gshard} and Switch Transformer~\cite{switchtransformer}. 
We use BPR for stability in V-MoE~\cite{vmoe}. 
For auxiliary loss, we use loading loss in~\cite{moe} for T-FFN's MoE and averaged loading loss and importance loss in V-MoE~\cite{vmoe} for V-FFN's MoE.
The combination ratio for auxiliary loss is set as $0.01$ in all our experiments
We use $32$ expert parallelism and \textsc{tutel}~\cite{tutel} for fast routing and computation. 
All the models are based on DeepSpeed~\cite{deepspeed}. 
Pre-training experiments are done on 32 Nvidia
Tesla V100-32GB GPUs. 
Following ST-MoE~\cite{stmoe}, we {\it freeze} all the MoE modules (router and expert network) during finetuning process. 
The capacity factor $C$ is set to be 1.05 during training and 1 during inference following~\cite{vmoe}. 

\paragraph{\ours{} in Pretraining.} 

We present the validation performance of \ours{} on the three pretraining tasks across different scales. The results show that the cost-performance tradeoff of \ours{} in terms of pretraining flops dominates the dense models by a wide margin, indicating that \ours{} offers significant improvements across all scales, from \textsc{small/8E} to \textsc{base/32E}. We also provide a wall-clock time versus validation performance figure in the Appendix, which shows a similar scaling trend of \ours{}. Thanks to careful kernel optimization and expert parallelism in DeepSpeed~\cite{deepspeed}, the maximum wall-clock overhead of \ours{}$_\textsc{base/32E}$ compared to dense counterparts can be reduced to only $13$\%.

\begin{table*}[t]
\small
\centering
\begin{tabular}{@{}lccccccccccc}
\toprule
\multirow{2}{*}{\bf Model} & \bf \# Pretrained & \bf \# Pretrained & \bf \# Params & \multicolumn{2}{c}{\bf VQA} & \multicolumn{2}{c}{\bf NLVR2} & \multicolumn{2}{c}{\bf COCO}  & \multicolumn{2}{c}{\bf Flickr30K} \\
 & \bf images & \bf Steps & \bf per token & test-dev & test-std & dev & test-P & TR & IR & TR & IR \\
\midrule
\multicolumn{12}{l}{
\textit{Base-size models pretrained in the similar settings}} \\
UNITER$_{\text{BASE}}$~\cite{uniter} & 4M & 200k & 86M & 72.70 & 72.91 & 77.18 & 77.85 & 64.4 & 50.3 & 85.9 & 72.5 \\
VILLA$_{\text{BASE}}$~\cite{villa} & 4M & 200k & 86M & 73.59 & 73.67 & 78.39 & 79.30 & - & - & 86.6 & 74.7 \\
UNIMO$_{\text{BASE}}$~\cite{unimo} & 4M & 500K & 120M & 73.79 & 74.02 & - & - & - & - & 89.7 & 74.7 \\
ViLT~\cite{vilt} & 4M & 200k & 120M & 71.26 & - & 75.70 & 76.13 & 61.5 & 42.7 & 83.5 & 64.4 \\
ALBEF$_{\text{BASE}}$~\cite{albef} & 4M & 240k & 210M & 74.54 & 74.70 & 80.24 & 80.50 & 73.1 & 56.8 & 94.3 & 82.8 \\
\vlmo{}$_{\text{BASE}}$~\cite{vlmo} & 4M & 200k & 180M & 76.64 & 76.89 & 82.77 & 83.34  & 74.8 & 57.2 & 92.3 & 79.3 \\
\vlbeit{}$_{\text{BASE}}$$^*$ & 4M & 200k & 180M & 76.21 & 76.75 & 84.93 & 85.76  & 78.7 & 60.3 & 95.3 & 83.8  \\
\rowcolor{Gray}
\ours{}$_{\text{BASE/32E}}$ & 4M & 200k & 180M & \textbf{78.23} & \textbf{78.65} & \textbf{85.54} & \textbf{86.77} & \textbf{79.4} & \textbf{61.2} & \textbf{96.1} & \textbf{84.9} \\
\midrule
\multicolumn{12}{l}{\textit{Pretained with more aggressive cost, including compute / data / model}} \\
\demph{VLMO$_{\text{LARGE}}$}~\cite{vlmo} & \demph{4M} & \demph{200k} & \demph{560M} & \demph{79.94} & \demph{79.98} & \demph{85.64} & \demph{86.86} & \demph{78.2} & \demph{60.6} & \demph{95.3} & \demph{84.5}  \\
\demph{ALBEF$_{\text{BASE}}$}~\cite{albef} & \demph{14M} & \demph{800k} & \demph{210M} & \demph{75.84} & \demph{76.04} & \demph{82.55} & \demph{83.14} & \demph{77.6} & \demph{60.7} & \demph{95.9} & \demph{85.6} \\
\demph{BLIP$_{\text{LARGE}}$~\cite{blip}} & \demph{129M} & \demph{1.26M} & \demph{427M} &  \demph{78.24} & \demph{78.17} & \demph{82.48} & \demph{83.08} & \demph{81.9} & \demph{64.3} & \demph{97.3} & \demph{87.3} \\
\demph{\textsc{SimVLM}$_{\text{BASE}}$~\cite{simvlm}} & \demph{1.8B} & \demph{1M} & \demph{230M} & \demph{77.87} & \demph{78.14} & \demph{81.72} & \demph{81.77} & \demph{-} & \demph{-} & \demph{-} & \demph{-} \\
\demph{\textsc{SimVLM}$_{\text{HUGE}}$}~\cite{simvlm} & \demph{1.8B} & \demph{1M} & \demph{1.7B} & \demph{80.03} & \demph{80.34} & \demph{84.53} & \demph{85.15} & \demph{-} & \demph{-} & \demph{-} & \demph{-} \\
\demph{\vlbeit{}$_{\text{HUGE}}$~\cite{vlbeit}} & \demph{21M} & \demph{1M} & \demph{1.9B} & \demph{84.19} & \demph{84.03} & \demph{91.51} & \demph{92.58} &  \demph{84.8} & \demph{67.2} & \demph{98.0} & \demph{90.3}  \\
\demph{\textsc{PaLI}$_{\text{HUGE}}$~\cite{vlbeit}} & \demph{1.6B} & \demph{1M} & \demph{17B} & \demph{84.30} & \demph{84.30} & \demph{-} & \demph{-} &  \demph{-} & \demph{-} & \demph{-} & \demph{-}  \\
\bottomrule
\end{tabular}
\caption{
Finetuning results of different models on vision-language classification tasks and image-text retrieval tasks.
We report vqa-score on VQA test-dev and test-standard split, accuracy for NLVR2 development and public test set (test-P) and top-1 recall for image retrieval (IR) and text retrieval (TR). ($^*$ denotes the model that is reproduced by us and trained with the same setting as \ours{.})
}
\label{tbl:results:vl_tasks}
\end{table*}

\subsection{Vision-and-Language Downstream Tasks}
In our study, we explore the performance of \ours{} on vision-and-language downstream tasks through fine-tuning experiments on three standard tasks: visual question answering~\cite{vqa}, natural language for visual reasoning~\cite{nlvr2}, and image-text retrieval~\cite{flickr30k,coco}. 
Following \vlbeit{}, we use $480\times480$ image resolution for VQA fine-tuning and $384\times384$ for the other tasks. 

\paragraph{Visual Question Answering (VQA).} 
For VQA, the task is to generate/choose the correct answer given a natural image and a question.
Following previous work~\cite{vilt,vlmo,vlbeit}, 
we utilize the VQA 2.0 dataset~\cite{vqa} and formulate it as a classification problem with $3,129$ most frequent answers.
We finetune \ours{} as a fusion network to encode both the image and question. 
We use the final encoding vector of the \sptk{T\_CLS} token as the representation of the image-question pair, and feed that into a classifier layer to predict the label.

\paragraph{Natural Language for Visual Reasoning (NLVR2).}

Visual reasoning task aims to predict whether a text description is true about a pair of images. 
We use NLVR2~\cite{nlvr2} dataset for evaluation.
Following OSCAR~\cite{oscar}, VinVL~\cite{vinvl} and \vlmo{}~\cite{vlmo}, we reformulate the triplet input into two image-text pairs, each containing the text description
and one image. 
We use \ours{} as a fusion network to jointly encode the image and text.
The concatenated final vector of \sptk{T\_CLS} token from the two pairs is then fed into a classification layer to predict the label.

\paragraph{Image-Text Retrieval.}
For image-text retrieval, it contains both image-to-text retrieval and text-to-image retrieval for different target modalities.
We use the widely used COCO~\cite{coco} and Flickr30K~\cite{flickr30k} datasets to evaluate the model, and adopt the Karpathy split~\cite{karpathysplit} following common practices. 
Noted that in the architecture of \ours{} and \vlbeit{}~\cite{vlbeit}, it does not involve the image-text matching module as existing in CLIP~\cite{clip}. 
To enable image-text matching, we further fine-tune \ours{} jointly with image-text contrastive and image-text matching with hard negative mining objectives as in \vlmo{}~\cite{vlmo} and \vlbeit{}.
During inference, \ours{} is  used to encode images and text separately and compute the matching scores by the dot product of image and text vectors to obtain the top-$k$ candidates. 
Table~\ref{tbl:results:vl_tasks} presents the results of our vision-language model on classification and retrieval tasks, including VQA, NLVR2, COCO, and Flickr30K. To ensure a fair comparison, we provide details on the amount of pretraining image-text pair data, pretraining steps, and the number of parameters per input token. Following \limoe~\cite{limoe}, we define the number of parameters per input token as the number of parameters that the model applies to each image-text token pair. Notably, \ours{}$_\textsc{base/32E}$ contains $2$ billion parameters in total, but only applies $180$ million parameters per token. Additionally, all routers combined account for less than $0.5$ million parameters. Our model outperforms previous base-size models on VQA, NLVR2, COCO, and Flickr30K by a significant margin, particularly when compared to a reproduced \vlbeit{}~\cite{vlbeit}, which was pretrained using the same settings as \ours{}. Moreover, to the best of our knowledge, \ours{} is the first to demonstrate that a mixture-of-experts architecture can successfully scale with a comparably modest architecture size and training counts, while achieving generalization performance on a range of tasks in the context of vision-language tasks. 
Interestingly, Switch Transformer~\cite{switchtransformer} struggles with generalization for language MoE, while \vmoe~\cite{vmoe} and \limoe~\cite{limoe} only evaluate on downstream vision tasks. 
Additionally, \ours{} even outperforms \vlmo{}$_\textsc{large}$ and ALBEF, which are pretrained with more image-text pair data and initialized from pretrained models, on COCO and Flickr30K and achieves competitive performance on VQA and NLVR2. 
We assume that this may be due to the fact that the capacity of VL-FFN has not been scaled in \ours{}, as reflected in the pretraining plot in Figure~\ref{fig:train_loss_beit} (the difference of VLM loss between \ours{} and dense \vlbeit{} model is smaller compared to that of MLM and MIM loss). We leave the scale of the VL-FFN module for future work, considering the increasing instability in modal-agnostic MoE architectures demonstrated in \limoe~\cite{limoe}. 

\begin{table}[t]
\centering
\begin{tabular}{p{2.9cm}@{}|p{1.2cm}@{}p{1.2cm}@{}|p{1.2cm}@{}p{1.0cm}}
\toprule
\multirow{2}{*}{\bf Models} & \multicolumn{2}{c}{\bf Pretraining}  & \multicolumn{2}{c}{\bf Tasks} \\ 
& \scriptsize \bf \# Images & \scriptsize \bf \# Steps & \scriptsize  \bf  ImageNet &  \scriptsize  \bf MNLI-m\\
\midrule
\multicolumn{3}{l}{\textit{Vision Pretraining}} \\
\textsc{ViT}$_\text{B/16}$~\cite{vit} & 300M & 500k & 83.6 & - \\ 
\textsc{BEiT}$_\text{B/16}$~\cite{beit} & 1.2M & 500k & 85.2 & - \\
\vmoe$_\text{B/16-32E}$~\cite{vmoe} & 300M & 500k & \bf 85.3 & - \\
\midrule
\multicolumn{3}{l}{\textit{Vision-Language Pretraining}} \\
\textsc{SimVLM}$_\textsc{base}$ & 1.8B & 1M & 80.6 & 64.4 \\
\vlbeit{}$_\textsc{base}^*$ & 4M & 200k & 83.2 & 67.0  \\
\rowcolor{Gray}
\ours{}$_\textsc{base/32E}$ &  4M &  200k & 84.5 & \bf 68.1 \\
\bottomrule
\end{tabular}
\caption{Results of base-size models on image classification (ImageNet-1K) and natural language inference (MNLI-m).
We report top-$1$ accuracy for ImageNet and MNLI-m.
}
\label{tbl:vision_and_language_only_tasks}
\vspace{-2mm}
\end{table}

\subsection{Vision/Language-Only Downstream Tasks}




\paragraph{Image Classification.}

We use the image classification task to evaluate the model on the vision-only downstream task, where the objective of this task is to categorize an input image into its corresponding class. 
We employ the ILSVRC-2012 ImageNet dataset~\cite{imagenet}, which consists of $1.3$M images with $1$k classes.
Following \beit{}~\cite{beit} and \vlmo{}~\cite{vlmo}, we perform average pooling over the final vectors and feed the resulting vector into a linear classifier layer to predict the label. 

\paragraph{Natural Language Inference.}

We use the natural language inference task to evaluate the model on the language-only downstream task. 
The task involves determining the relationship between two pieces of text. 
In this task, a model is given a premise sentence and a hypothesis sentence, and it needs to determine whether the hypothesis is true, false, or undetermined based on the information provided in the premise. 
We use Multi-Genre Natural Language Inference (MNLI)~\cite{mnli2017} dataset, which contains 433k sentence pairs annotated with textual entailment information. 
We evaluate on matched (MLM-m) setting only. 

As shown Table~\ref{tbl:vision_and_language_only_tasks}, we compare \ours{} with two base-size vision Transformers and \vmoe-B/16-E32 on image classification.
For \beit{}, \vlbeit{}$_\textsc{base}$ and \ours{}$_\textsc{base/32E}$, we perform intermediate finetuning on ImageNet-22k to compare with \textsc{ViT} pretrained on ImageNet-22k. 
The model performs competitively with previous state-of-the-art supervised and self-supervised models on ImageNet-1k.
Besides the dense counterpart \vlbeit{}$_\textsc{base}$, \ours{} also outperforms other strong vision-language models (\textsc{SimVLM}) pretrained with more data and more steps on MNLI-m.

\begin{figure}[t]
    \centering
    \includegraphics[width=0.46\textwidth]{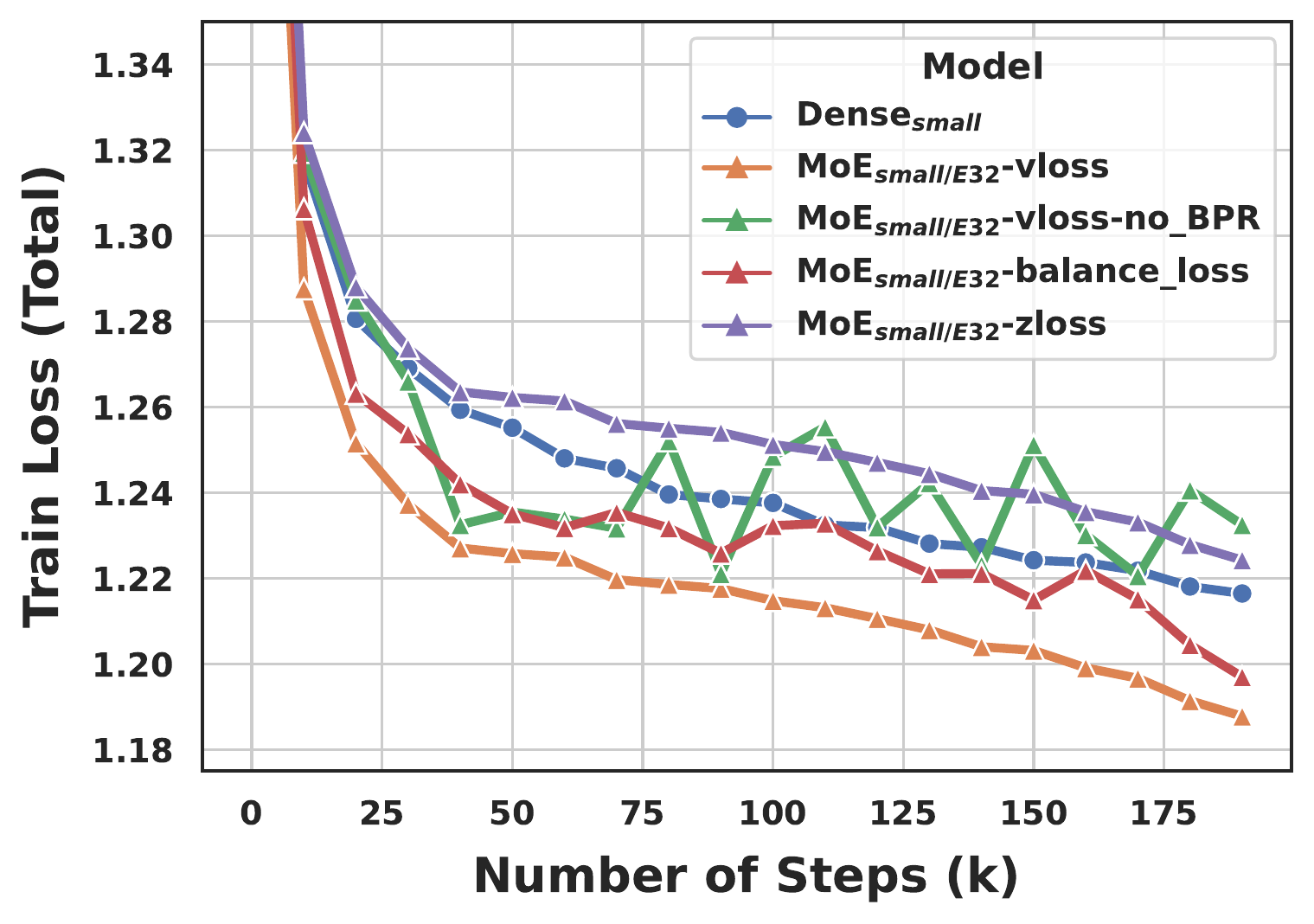}
    \vspace{-2mm}
    \caption{Effect of auxiliary loss on training stability.}
    \label{fig:ablate:loss}
    \vspace{-4mm}
\end{figure}
\section{Discussions}
\label{sec:ablation}
\begin{table*}[t]
\centering
\begin{tabular}{lcc|ccccccc}
\toprule
& \multicolumn{2}{c}{\textbf{Scaling Strategy}}&
\multicolumn{2}{c}{\textbf{NLVR2}} & \multicolumn{2}{c}{\textbf{Flickr30k}} & \textbf{ImageNet} & \textbf{MNLI-m} & \multirow{2}{*}{\bf Avg.} \\
& T-MoE & V-MoE &  dev & test-P & TR R@1 & IR R@1 & Acc@1 & Acc &  \\
\midrule
\tblidx{1} & \xmark  &\xmark & 67.42 & 68.21 & 80.4 & 61.7 & 67.2 & 54.3  & 66.5 \\
\tblidx{2} & \cmark  &\xmark & 72.42 & 72.73 & 83.2 & 64.7 & 67.8 & \textbf{58.3} & 69.9 \\
\tblidx{3} & \xmark  &\cmark & 71.19 & 72.23 & 82.9 & 64.5 & \textbf{69.2} & 55.2 & 69.2 \\
\tblidx{4} & \cmark  &\cmark & \textbf{72.98} & \textbf{73.34} & \textbf{84.7} & \textbf{65.3}  & 69.0 & 58.1 & \textbf{70.6} \\
\bottomrule
\end{tabular}
\caption{
Ablation studies of scaling strategies (all the results are based on \ours{}$_\textsc{small/E32}$ models). All the *-MoE uses 32 experts (where T stands for applying MoE on the T-FFN and V stands for applying MoE on the V-FFN).  
}
\label{tbl:ablation:scaling_strategy}
\vspace{-2mm}
\end{table*}

We conduct ablation studies to analyze the contributions of Mixture-of-Experts module used in \ours{} from different perspectives.
We evaluate the models on visual reasoning (NLVR2), image-text retrieval (Flickr30k), image classification (ImageNet-1k) and natural language inference (MNLI-m).

\paragraph{Scaling Strategy.} 
In addition to scaling both T-FFN and V-FFN, we have also explored different scaling strategies by applying Mixture-of-Experts (MoEs) modules for either T-FFN or V-FFN alone. The results of our experiments are presented in Table~\ref{tbl:ablation:scaling_strategy}. 
Our findings indicate that scaling a single modality can improve the downstream performance on the corresponding modality as well as overall vision-language tasks. 
However, we observed that scaling both vision and language modalities leads to the most balanced performing model with $70.6$\% averaged performance. 
This may be attributed to the fact that we employ three different pretraining objectives for each modality, and scaling each modality contributes to better optimization of the specific modality pretraining loss as well as the VLM loss. 
For further evidence, we include the pre-training loss in Appendix.

\begin{figure}[t]
    \centering
    \includegraphics[width=0.4\textwidth]{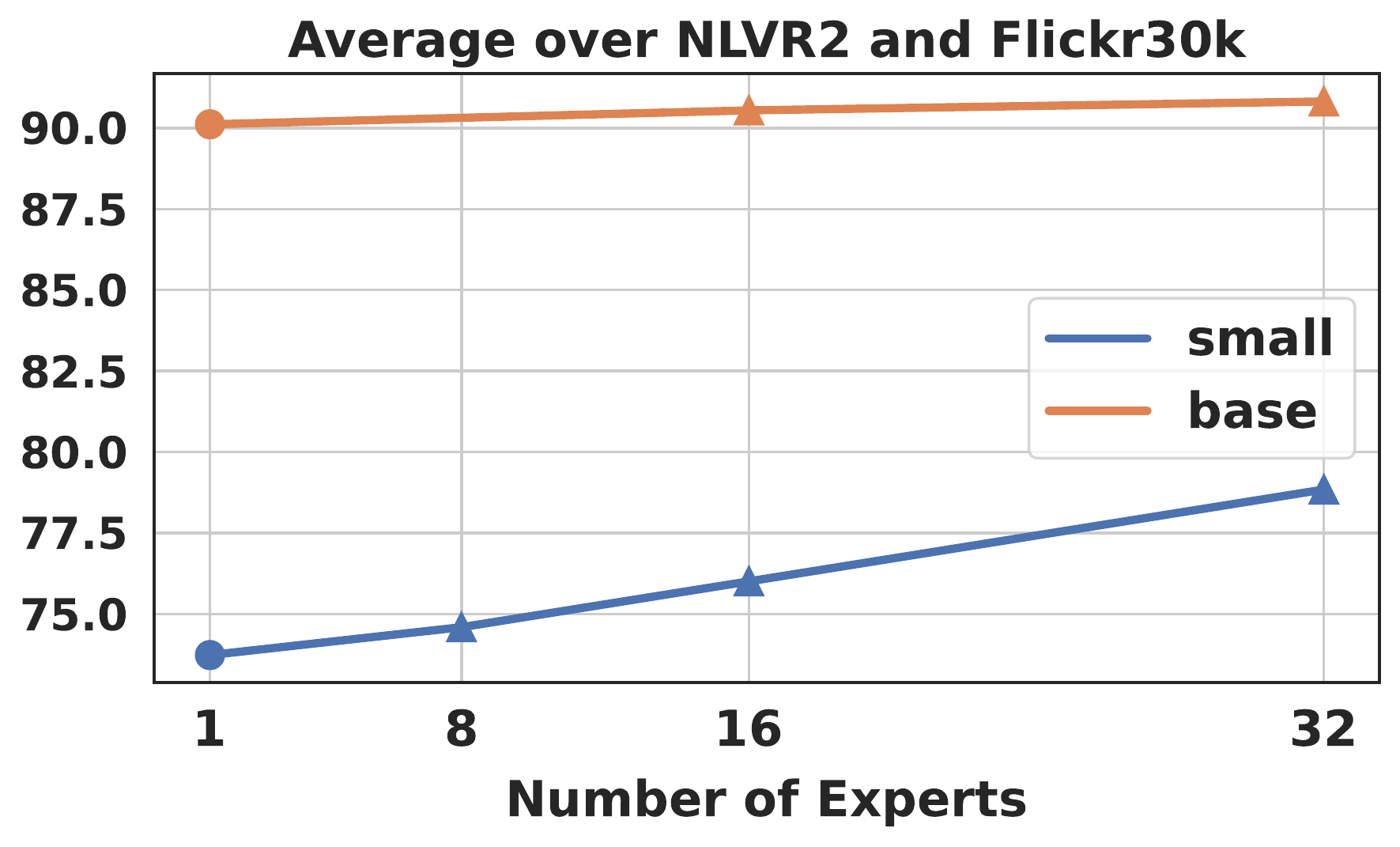}
    \caption{Effect of Experts Number on Downstream tasks.
    }
    \label{fig:ablate:expert_number}
\end{figure}

\paragraph{Number of Experts.}
The optimal number of experts in Mixture-of-Experts (MoEs) is still a topic of debate, as there is no agreement on the ideal number. Previous NLP research has experimented with a wide range of expert numbers, ranging from thousands in early studies~\cite{moe,switchtransformer}, to as low as 32 or 64 in more recent research~\cite{stmoe,glam,expertchoice}, which has become the standard for vision models~\cite{vmoe,limoe}. In Figure~\ref{fig:ablate:expert_number}, we investigate this further with \ours{}, and our findings suggest that larger expert pools consistently yield performance improvements.

\paragraph{Effects of the Auxiliary Losses.} 

As previously mentioned, experts in MoEs have a fixed buffer capacity, and without intervention, top-$k$ MoEs tend to collapse, leading to poor performance as most tokens are dropped~\cite{moe,expertchoice}. 
To prevent this, prior research has employed auxiliary losses to promote balanced routing ~\cite{vmoe,stmoe,expertchoice,limoe}. 
However, as shown in \limoe~\cite{limoe}, in multimodal settings, new challenges emerge, such as modality misbalance, where one data type may be more prevalent than the other. 
We design \ours{} in a modal-specific fashion to prevent the instability caused by imbalance of multimodal data and  experiment with different auxiliary losses for V-MoE: loading balance loss~\cite{moe}, averaged loading balance and important loss (``vloss'')~\cite{vmoe}, z-loss~\cite{stmoe}). \footnote{We find that the T-MoE is quite stable using different auxiliary losses, and resort to the most common loading balance loss in~\cite{moe} for T-MoE. We detail the formula of each auxiliary loss in the Appendix.}
We present the results on \ours{}$_\textsc{small/E32}$ in Figure~\ref{fig:ablate:loss}, which suggest that Z-loss presents to hurt the vision-and-lanaguage pretraininig of \ours{} and using loading balance loss only will introduce unstable training and underperforming models. 
The ``vloss'' turns out to lead to most stable training, which is consistent with \vmoe~\cite{vmoe} and \limoe~\cite{limoe}. 
BPR also helps in stablizing training.

\paragraph{Token Routing Examples in \ours{}.} 

In Figure~\ref{fig:overall_token_route}, we provide a qualitative analysis of token routing decisions on COCO. 
For vision tokens, their specialization is clear, as they are routed to specific experts such as food and vegetable experts, eyes experts, OCR experts, etc. 
On the other hand, language tokens show signs of syntax specialization, with some experts processing mostly padding tokens, while others focus on nouns and adjectives (and some padding), excluding prepositions, determiners, or verbs.

\section{Conclusion}
\label{sec:conclusion}

In this paper, we have explored the use of Mixture-of-Experts (MoE) for scaling vision-language models. Our experiments demonstrate that MoE can be a promising technique for improving the efficiency and effectiveness of vision-language models. Specifically, we have shown that dividing a large vision-language model into smaller, specialized sub-models through MoE can achieve state-of-the-art performance on several benchmarks while reducing computational costs. 
Our experiments have also shown that larger expert pools yield consistent performance improvements. 
Furthermore, we have 
explored the impact of MoE on model interpretability and found it can improve the interpretability of vision-language models by providing better insights into how the model processes different inputs.

In conclusion, our findings suggest that MoE is a valuable technique for scaling vision-language models, enabling them to handle large-scale, real-world multimedia data. 
Our work opens up new research directions for exploring the effectiveness of MoEs in other vision-language tasks, such as visual question answering, visual reasoning and image-text retrieval, and we hope our findings will inspire further investigations into this research area.
\section*{Acknowledgements}
We gratefully acknowledge Wenhui Wang, Li Dong, Furu Wei for the early insightful discussions on the implementation details of \mome{}, Mengchen Liu for the MoEs models for unified contrastive learning.
SS, KK and TD are partly supported by Samsung SAIT, Intel corporation, Intel VLAB team, Intel One-API center of excellence, DARPA’s LwLL, PTG, SemaFor, as well as funding through BDD and BAIR. 

{
\bibliographystyle{ieee_fullname}
\bibliography{egbib}

\begin{thebibliography}{10}\itemsep=-1pt

\bibitem{flamingo}
Jean{-}Baptiste Alayrac, Jeff Donahue, Pauline Luc, Antoine Miech, Iain Barr,
  Yana Hasson, Karel Lenc, Arthur Mensch, Katie Millican, Malcolm Reynolds,
  Roman Ring, Eliza Rutherford, Serkan Cabi, Tengda Han, Zhitao Gong, Sina
  Samangooei, Marianne Monteiro, Jacob Menick, Sebastian Borgeaud, Andrew
  Brock, Aida Nematzadeh, Sahand Sharifzadeh, Mikolaj Binkowski, Ricardo
  Barreira, Oriol Vinyals, Andrew Zisserman, and Karen Simonyan.
\newblock Flamingo: a visual language model for few-shot learning.
\newblock {\em CoRR}, abs/2204.14198, 2022.

\bibitem{beit}
Hangbo Bao, Li Dong, Songhao Piao, and Furu Wei.
\newblock {BEiT}: {BERT} pre-training of image transformers.
\newblock In {\em ICLR}, 2022.

\bibitem{vlmo}
Hangbo Bao, Wenhui Wang, Li Dong, Qiang Liu, Owais~Khan Mohammed, Kriti
  Aggarwal, Subhojit Som, Songhao Piao, and Furu Wei.
\newblock Vlmo: Unified vision-language pre-training with
  mixture-of-modality-experts.
\newblock In {\em Advances in Neural Information Processing Systems}.

\bibitem{bengio2013deep}
Yoshua Bengio.
\newblock Deep learning of representations: Looking forward.
\newblock In {\em Statistical Language and Speech Processing: First
  International Conference, SLSP 2013, Tarragona, Spain, July 29-31, 2013.
  Proceedings 1}, pages 1--37. Springer, 2013.

\bibitem{chen1999improved}
Ke Chen, Lei Xu, and Huisheng Chi.
\newblock Improved learning algorithms for mixture of experts in multiclass
  classification.
\newblock {\em Neural networks}, 12(9):1229--1252, 1999.

\bibitem{pali}
Xi Chen, Xiao Wang, Soravit Changpinyo, AJ Piergiovanni, Piotr Padlewski,
  Daniel Salz, Sebastian Goodman, Adam Grycner, Basil Mustafa, Lucas Beyer,
  et~al.
\newblock Pali: A jointly-scaled multilingual language-image model.
\newblock {\em arXiv preprint arXiv:2209.06794}, 2022.

\bibitem{uniter}
Yen{-}Chun Chen, Linjie Li, Licheng Yu, Ahmed~El Kholy, Faisal Ahmed, Zhe Gan,
  Yu Cheng, and Jingjing Liu.
\newblock {UNITER:} universal image-text representation learning.
\newblock In Andrea Vedaldi, Horst Bischof, Thomas Brox, and Jan{-}Michael
  Frahm, editors, {\em Computer Vision - {ECCV} 2020 - 16th European
  Conference, Glasgow, UK, August 23-28, 2020, Proceedings, Part {XXX}}, volume
  12375 of {\em Lecture Notes in Computer Science}, pages 104--120. Springer,
  2020.

\bibitem{unifiedscaling}
Aidan Clark, Diego De~Las~Casas, Aurelia Guy, Arthur Mensch, Michela Paganini,
  Jordan Hoffmann, Bogdan Damoc, Blake Hechtman, Trevor Cai, Sebastian
  Borgeaud, et~al.
\newblock Unified scaling laws for routed language models.
\newblock In {\em ICML}, pages 4057--4086. PMLR, 2022.

\bibitem{routingcontinual}
Mark Collier, Efi Kokiopoulou, Andrea Gesmundo, and Jesse Berent.
\newblock Routing networks with co-training for continual learning.
\newblock {\em arXiv preprint arXiv:2009.04381}, 2020.

\bibitem{davis2013low}
Andrew Davis and Itamar Arel.
\newblock Low-rank approximations for conditional feedforward computation in
  deep neural networks.
\newblock {\em arXiv preprint arXiv:1312.4461}, 2013.

\bibitem{bert}
Jacob Devlin, Ming{-}Wei Chang, Kenton Lee, and Kristina Toutanova.
\newblock {BERT:} pre-training of deep bidirectional transformers for language
  understanding.
\newblock In Jill Burstein, Christy Doran, and Thamar Solorio, editors, {\em
  Proceedings of the 2019 Conference of the North American Chapter of the
  Association for Computational Linguistics: Human Language Technologies,
  {NAACL-HLT} 2019, Minneapolis, MN, USA, June 2-7, 2019, Volume 1 (Long and
  Short Papers)}, pages 4171--4186. Association for Computational Linguistics,
  2019.

\bibitem{vit}
Alexey Dosovitskiy, Lucas Beyer, Alexander Kolesnikov, Dirk Weissenborn,
  Xiaohua Zhai, Thomas Unterthiner, Mostafa Dehghani, Matthias Minderer, Georg
  Heigold, Sylvain Gelly, et~al.
\newblock An image is worth 16x16 words: Transformers for image recognition at
  scale.
\newblock {\em preprint arXiv:2010.11929}, 2020.

\bibitem{glam}
Nan Du, Yanping Huang, Andrew~M Dai, Simon Tong, Dmitry Lepikhin, Yuanzhong Xu,
  Maxim Krikun, Yanqi Zhou, Adams~Wei Yu, Orhan Firat, et~al.
\newblock Glam: Efficient scaling of language models with mixture-of-experts.
\newblock In {\em ICML}, pages 5547--5569. PMLR, 2022.

\bibitem{eigen2013learning}
David Eigen, Marc'Aurelio Ranzato, and Ilya Sutskever.
\newblock Learning factored representations in a deep mixture of experts.
\newblock {\em arXiv preprint arXiv:1312.4314}, 2013.

\bibitem{switchtransformer}
William Fedus, Barret Zoph, and Noam Shazeer.
\newblock Switch transformers: Scaling to trillion parameter models with simple
  and efficient sparsity.
\newblock {\em CoRR}, abs/2101.03961, 2021.

\bibitem{villa}
Zhe Gan, Yen{-}Chun Chen, Linjie Li, Chen Zhu, Yu Cheng, and Jingjing Liu.
\newblock Large-scale adversarial training for vision-and-language
  representation learning.
\newblock In Hugo Larochelle, Marc'Aurelio Ranzato, Raia Hadsell,
  Maria{-}Florina Balcan, and Hsuan{-}Tien Lin, editors, {\em Advances in
  Neural Information Processing Systems 33: Annual Conference on Neural
  Information Processing Systems 2020, NeurIPS 2020, December 6-12, 2020,
  virtual}, 2020.

\bibitem{vqa}
Yash Goyal, Tejas Khot, Douglas Summers{-}Stay, Dhruv Batra, and Devi Parikh.
\newblock Making the {V} in {VQA} matter: Elevating the role of image
  understanding in visual question answering.
\newblock In {\em 2017 {IEEE} Conference on Computer Vision and Pattern
  Recognition, {CVPR} 2017, Honolulu, HI, USA, July 21-26, 2017}, pages
  6325--6334. {IEEE} Computer Society, 2017.

\bibitem{hazimeh2021dselectk}
Hussein Hazimeh, Zhe Zhao, Aakanksha Chowdhery, Maheswaran Sathiamoorthy, Yihua
  Chen, Rahul Mazumder, Lichan Hong, and Ed~H. Chi.
\newblock Dselect-k: Differentiable selection in the mixture of experts with
  applications to multi-task learning.
\newblock In {\em Advances in Neural Information Processing Systems 34: Annual
  Conference on Neural Information Processing Systems 2021, NeurIPS 2021,
  December 6-14, 2021, virtual}, 2021.

\bibitem{gelu}
Dan Hendrycks and Kevin Gimpel.
\newblock Gaussian error linear units ({GELU}s).
\newblock {\em arXiv preprint arXiv:1606.08415}, 2016.

\bibitem{drop_path}
Gao Huang, Yu Sun, Zhuang Liu, Daniel Sedra, and Kilian~Q. Weinberger.
\newblock Deep networks with stochastic depth.
\newblock In Bastian Leibe, Jiri Matas, Nicu Sebe, and Max Welling, editors,
  {\em Computer Vision - {ECCV} 2016 - 14th European Conference, Amsterdam, The
  Netherlands, October 11-14, 2016, Proceedings, Part {IV}}, volume 9908 of
  {\em Lecture Notes in Computer Science}, pages 646--661. Springer, 2016.

\bibitem{tutel}
Changho Hwang, Wei Cui, Yifan Xiong, Ziyue Yang, Ze Liu, Han Hu, Zilong Wang,
  Rafael Salas, Jithin Jose, Prabhat Ram, et~al.
\newblock Tutel: Adaptive mixture-of-experts at scale.
\newblock {\em arXiv preprint arXiv:2206.03382}, 2022.

\bibitem{jacobs1991adaptive}
Robert~A Jacobs, Michael~I Jordan, Steven~J Nowlan, and Geoffrey~E Hinton.
\newblock Adaptive mixtures of local experts.
\newblock {\em Neural computation}, 3(1):79--87, 1991.

\bibitem{align}
Chao Jia, Yinfei Yang, Ye Xia, Yi{-}Ting Chen, Zarana Parekh, Hieu Pham,
  Quoc~V. Le, Yun{-}Hsuan Sung, Zhen Li, and Tom Duerig.
\newblock Scaling up visual and vision-language representation learning with
  noisy text supervision.
\newblock In Marina Meila and Tong Zhang, editors, {\em Proceedings of the 38th
  International Conference on Machine Learning, {ICML} 2021, 18-24 July 2021,
  Virtual Event}, volume 139 of {\em Proceedings of Machine Learning Research},
  pages 4904--4916. {PMLR}, 2021.

\bibitem{karpathysplit}
Andrej Karpathy and Li Fei{-}Fei.
\newblock Deep visual-semantic alignments for generating image descriptions.
\newblock In {\em {IEEE} Conference on Computer Vision and Pattern Recognition,
  {CVPR} 2015, Boston, MA, USA, June 7-12, 2015}, pages 3128--3137. {IEEE}
  Computer Society, 2015.

\bibitem{vilt}
Wonjae Kim, Bokyung Son, and Ildoo Kim.
\newblock {ViLT}: Vision-and-language transformer without convolution or region
  supervision.
\newblock In Marina Meila and Tong Zhang, editors, {\em Proceedings of the 38th
  International Conference on Machine Learning, {ICML} 2021, 18-24 July 2021,
  Virtual Event}, volume 139 of {\em Proceedings of Machine Learning Research},
  pages 5583--5594. {PMLR}, 2021.

\bibitem{zeromoe}
Young~Jin Kim, Ammar~Ahmad Awan, Alexandre Muzio, Andres Felipe~Cruz Salinas,
  Liyang Lu, Amr Hendy, Samyam Rajbhandari, Yuxiong He, and Hany~Hassan
  Awadalla.
\newblock Scalable and efficient moe training for multitask multilingual
  models.
\newblock {\em arXiv preprint arXiv:2109.10465}, 2021.

\bibitem{adam}
Diederik~P. Kingma and Jimmy Ba.
\newblock Adam: {A} method for stochastic optimization.
\newblock In Yoshua Bengio and Yann LeCun, editors, {\em 3rd International
  Conference on Learning Representations, {ICLR} 2015, San Diego, CA, USA, May
  7-9, 2015, Conference Track Proceedings}, 2015.

\bibitem{koh2023grounding}
Jing~Yu Koh, Ruslan Salakhutdinov, and Daniel Fried.
\newblock Grounding language models to images for multimodal generation.
\newblock {\em arXiv preprint arXiv:2301.13823}, 2023.

\bibitem{sparseupcycle}
Aran Komatsuzaki, Joan Puigcerver, James Lee-Thorp, Carlos~Riquelme Ruiz, Basil
  Mustafa, Joshua Ainslie, Yi Tay, Mostafa Dehghani, and Neil Houlsby.
\newblock Sparse upcycling: Training mixture-of-experts from dense checkpoints.
\newblock {\em arXiv preprint arXiv:2212.05055}, 2022.

\bibitem{vg}
Ranjay Krishna, Yuke Zhu, Oliver Groth, Justin Johnson, Kenji Hata, Joshua
  Kravitz, Stephanie Chen, Yannis Kalantidis, Li{-}Jia Li, David~A. Shamma,
  Michael~S. Bernstein, and Li Fei{-}Fei.
\newblock Visual genome: Connecting language and vision using crowdsourced
  dense image annotations.
\newblock {\em Int. J. Comput. Vis.}, 123(1):32--73, 2017.

\bibitem{sentencepiece}
Taku Kudo and John Richardson.
\newblock {S}entence{P}iece: A simple and language independent subword
  tokenizer and detokenizer for neural text processing.
\newblock In {\em Proceedings of the 2018 Conference on Empirical Methods in
  Natural Language Processing: System Demonstrations}, pages 66--71, Brussels,
  Belgium, Nov. 2018. Association for Computational Linguistics.

\bibitem{taskmoe}
Sneha Kudugunta, Yanping Huang, Ankur Bapna, Maxim Krikun, Dmitry Lepikhin,
  Minh-Thang Luong, and Orhan Firat.
\newblock Beyond distillation: Task-level mixture-of-experts for efficient
  inference.
\newblock In {\em Findings of the Association for Computational Linguistics:
  EMNLP 2021}, pages 3577--3599, 2021.

\bibitem{gshard}
Dmitry Lepikhin, HyoukJoong Lee, Yuanzhong Xu, Dehao Chen, Orhan Firat, Yanping
  Huang, Maxim Krikun, Noam Shazeer, and Zhifeng Chen.
\newblock Gshard: Scaling giant models with conditional computation and
  automatic sharding.
\newblock {\em arXiv preprint arXiv:2006.16668}, 2020.

\bibitem{lewis2021base}
Mike Lewis, Shruti Bhosale, Tim Dettmers, Naman Goyal, and Luke Zettlemoyer.
\newblock {BASE} layers: Simplifying training of large, sparse models.
\newblock In {\em ICML}. {PMLR}, 2021.

\bibitem{li2023blip2}
Junnan Li, Dongxu Li, Silvio Savarese, and Steven Hoi.
\newblock Blip-2: Bootstrapping language-image pre-training with frozen image
  encoders and large language models.
\newblock {\em arXiv preprint arXiv:2301.12597}, 2023.

\bibitem{blip}
Junnan Li, Dongxu Li, Caiming Xiong, and Steven Hoi.
\newblock Blip: Bootstrapping language-image pre-training for unified
  vision-language understanding and generation.
\newblock In {\em ICML}, pages 12888--12900. PMLR, 2022.

\bibitem{albef}
Junnan Li, Ramprasaath Selvaraju, Akhilesh Gotmare, Shafiq Joty, Caiming Xiong,
  and Steven Chu~Hong Hoi.
\newblock Align before fuse: Vision and language representation learning with
  momentum distillation.
\newblock In {\em Advances in neural information processing systems},
  volume~34, pages 9694--9705, 2021.

\bibitem{visualbert}
Liunian~Harold Li, Mark Yatskar, Da Yin, Cho{-}Jui Hsieh, and Kai{-}Wei Chang.
\newblock Visualbert: {A} simple and performant baseline for vision and
  language.
\newblock {\em CoRR}, abs/1908.03557, 2019.

\bibitem{unimo}
Wei Li, Can Gao, Guocheng Niu, Xinyan Xiao, Hao Liu, Jiachen Liu, Hua Wu, and
  Haifeng Wang.
\newblock {UNIMO:} towards unified-modal understanding and generation via
  cross-modal contrastive learning.
\newblock In Chengqing Zong, Fei Xia, Wenjie Li, and Roberto Navigli, editors,
  {\em Proceedings of the 59th Annual Meeting of the Association for
  Computational Linguistics and the 11th International Joint Conference on
  Natural Language Processing, {ACL/IJCNLP} 2021, (Volume 1: Long Papers),
  Virtual Event, August 1-6, 2021}, pages 2592--2607. Association for
  Computational Linguistics, 2021.

\bibitem{oscar}
Xiujun Li, Xi Yin, Chunyuan Li, Pengchuan Zhang, Xiaowei Hu, Lei Zhang, Lijuan
  Wang, Houdong Hu, Li Dong, Furu Wei, Yejin Choi, and Jianfeng Gao.
\newblock Oscar: Object-semantics aligned pre-training for vision-language
  tasks.
\newblock In Andrea Vedaldi, Horst Bischof, Thomas Brox, and Jan{-}Michael
  Frahm, editors, {\em Computer Vision - {ECCV} 2020 - 16th European
  Conference, Glasgow, UK, August 23-28, 2020, Proceedings, Part {XXX}}, volume
  12375 of {\em Lecture Notes in Computer Science}, pages 121--137. Springer,
  2020.

\bibitem{li2023gligen}
Yuheng Li, Haotian Liu, Qingyang Wu, Fangzhou Mu, Jianwei Yang, Jianfeng Gao,
  Chunyuan Li, and Yong~Jae Lee.
\newblock Gligen: Open-set grounded text-to-image generation.
\newblock {\em CVPR}, 2023.

\bibitem{coco}
Tsung{-}Yi Lin, Michael Maire, Serge~J. Belongie, James Hays, Pietro Perona,
  Deva Ramanan, Piotr Doll{\'{a}}r, and C.~Lawrence Zitnick.
\newblock Microsoft {COCO:} common objects in context.
\newblock In David~J. Fleet, Tom{\'{a}}s Pajdla, Bernt Schiele, and Tinne
  Tuytelaars, editors, {\em Computer Vision - {ECCV} 2014 - 13th European
  Conference, Zurich, Switzerland, September 6-12, 2014, Proceedings, Part
  {V}}, volume 8693 of {\em Lecture Notes in Computer Science}, pages 740--755.
  Springer, 2014.

\bibitem{liu2023learning}
Haotian Liu, Kilho Son, Jianwei Yang, Ce Liu, Jianfeng Gao, Yong~Jae Lee, and
  Chunyuan Li.
\newblock Learning customized visual models with retrieval-augmented knowledge.
\newblock {\em CVPR}, 2023.

\bibitem{mixermoe}
Yuxuan Lou, Fuzhao Xue, Zangwei Zheng, and Yang You.
\newblock Cross-token modeling with conditional computation.
\newblock {\em arXiv preprint arXiv:2109.02008}, 2021.

\bibitem{vilbert}
Jiasen Lu, Dhruv Batra, Devi Parikh, and Stefan Lee.
\newblock {ViLBERT}: Pretraining task-agnostic visiolinguistic representations
  for vision-and-language tasks.
\newblock In {\em Advances in Neural Information Processing Systems 32: Annual
  Conference on Neural Information Processing Systems 2019, NeurIPS 2019,
  December 8-14, 2019, Vancouver, BC, Canada}, pages 13--23, 2019.

\bibitem{mmoe}
Jiaqi Ma, Zhe Zhao, Xinyang Yi, Jilin Chen, Lichan Hong, and Ed~H. Chi.
\newblock Modeling task relationships in multi-task learning with multi-gate
  mixture-of-experts.
\newblock In {\em Proceedings of the 24th {ACM} {SIGKDD} International
  Conference on Knowledge Discovery {\&} Data Mining, {KDD} 2018, London, UK,
  August 19-23, 2018}. {ACM}, 2018.

\bibitem{limoe}
Basil Mustafa, Carlos Riquelme, Joan Puigcerver, Rodolphe Jenatton, and Neil
  Houlsby.
\newblock Multimodal contrastive learning with limoe: the language-image
  mixture of experts.
\newblock {\em arXiv preprint arXiv:2206.02770}, 2022.

\bibitem{sbu}
Vicente Ordonez, Girish Kulkarni, and Tamara~L. Berg.
\newblock Im2text: Describing images using 1 million captioned photographs.
\newblock In John Shawe{-}Taylor, Richard~S. Zemel, Peter~L. Bartlett, Fernando
  C.~N. Pereira, and Kilian~Q. Weinberger, editors, {\em Advances in Neural
  Information Processing Systems 24: 25th Annual Conference on Neural
  Information Processing Systems 2011. Proceedings of a meeting held 12-14
  December 2011, Granada, Spain}, pages 1143--1151, 2011.

\bibitem{beitv2}
Zhiliang Peng, Li Dong, Hangbo Bao, Qixiang Ye, and Furu Wei.
\newblock Beit v2: Masked image modeling with vector-quantized visual
  tokenizers.
\newblock {\em ArXiv}, abs/2208.06366, 2022.

\bibitem{flickr30k}
Bryan~A. Plummer, Liwei Wang, Chris~M. Cervantes, Juan~C. Caicedo, Julia
  Hockenmaier, and Svetlana Lazebnik.
\newblock Flickr30k entities: Collecting region-to-phrase correspondences for
  richer image-to-sentence models.
\newblock In {\em 2015 {IEEE} International Conference on Computer Vision,
  {ICCV} 2015, Santiago, Chile, December 7-13, 2015}, pages 2641--2649. {IEEE}
  Computer Society, 2015.

\bibitem{clip}
Alec Radford, Jong~Wook Kim, Chris Hallacy, Aditya Ramesh, Gabriel Goh,
  Sandhini Agarwal, Girish Sastry, Amanda Askell, Pamela Mishkin, Jack Clark,
  Gretchen Krueger, and Ilya Sutskever.
\newblock Learning transferable visual models from natural language
  supervision.
\newblock In Marina Meila and Tong Zhang, editors, {\em Proceedings of the 38th
  International Conference on Machine Learning, {ICML} 2021, 18-24 July 2021,
  Virtual Event}, volume 139 of {\em Proceedings of Machine Learning Research},
  pages 8748--8763. {PMLR}, 2021.

\bibitem{deepspeed}
Jeff Rasley, Samyam Rajbhandari, Olatunji Ruwase, and Yuxiong He.
\newblock Deepspeed: System optimizations enable training deep learning models
  with over 100 billion parameters.
\newblock In {\em Proceedings of the 26th ACM SIGKDD International Conference
  on Knowledge Discovery \& Data Mining}, pages 3505--3506, 2020.

\bibitem{vmoe}
Carlos Riquelme, Joan Puigcerver, Basil Mustafa, Maxim Neumann, Rodolphe
  Jenatton, Andr{\'e} Susano~Pinto, Daniel Keysers, and Neil Houlsby.
\newblock Scaling vision with sparse mixture of experts.
\newblock {\em Advances in Neural Information Processing Systems}, 2021.

\bibitem{roller2021hash}
Stephen Roller, Sainbayar Sukhbaatar, Arthur Szlam, and Jason Weston.
\newblock Hash layers for large sparse models.
\newblock In {\em Advances in Neural Information Processing Systems 34: Annual
  Conference on Neural Information Processing Systems 2021, NeurIPS 2021,
  December 6-14, 2021, virtual}, 2021.

\bibitem{imagenet}
Olga Russakovsky, Jia Deng, Hao Su, Jonathan Krause, Sanjeev Satheesh, Sean Ma,
  Zhiheng Huang, Andrej Karpathy, Aditya Khosla, Michael Bernstein, Alexander~C
  Berg, and Li Fei-Fei.
\newblock Imagenet large scale visual recognition challenge.
\newblock {\em IJCV}, 2015.

\bibitem{gcc}
Piyush Sharma, Nan Ding, Sebastian Goodman, and Radu Soricut.
\newblock Conceptual captions: {A} cleaned, hypernymed, image alt-text dataset
  for automatic image captioning.
\newblock In Iryna Gurevych and Yusuke Miyao, editors, {\em Proceedings of the
  56th Annual Meeting of the Association for Computational Linguistics, {ACL}
  2018, Melbourne, Australia, July 15-20, 2018, Volume 1: Long Papers}, pages
  2556--2565. Association for Computational Linguistics, 2018.

\bibitem{moe}
Noam Shazeer, Azalia Mirhoseini, Krzysztof Maziarz, Andy Davis, Quoc~V. Le,
  Geoffrey~E. Hinton, and Jeff Dean.
\newblock Outrageously large neural networks: The sparsely-gated
  mixture-of-experts layer.
\newblock In {\em ICLR}. OpenReview.net, 2017.

\bibitem{klite}
Sheng Shen, Chunyuan Li, Xiaowei Hu, Yujia Xie, Jianwei Yang, Pengchuan Zhang,
  Zhe Gan, Lijuan Wang, Lu Yuan, Ce Liu, et~al.
\newblock K-lite: Learning transferable visual models with external knowledge.
\newblock In {\em Advances in Neural Information Processing Systems}, 2022.

\bibitem{vilclip}
Sheng Shen, Liunian~Harold Li, Hao Tan, Mohit Bansal, Anna Rohrbach, Kai-Wei
  Chang, Zhewei Yao, and Kurt Keutzer.
\newblock How much can clip benefit vision-and-language tasks?
\newblock In {\em ICLR}, 2022.

\bibitem{flava}
Amanpreet Singh, Ronghang Hu, Vedanuj Goswami, Guillaume Couairon, Wojciech
  Galuba, Marcus Rohrbach, and Douwe Kiela.
\newblock {FLAVA:} {A} foundational language and vision alignment model.
\newblock {\em CoRR}, abs/2112.04482, 2021.

\bibitem{vl-bert}
Weijie Su, Xizhou Zhu, Yue Cao, Bin Li, Lewei Lu, Furu Wei, and Jifeng Dai.
\newblock {VL-BERT:} pre-training of generic visual-linguistic representations.
\newblock In {\em 8th International Conference on Learning Representations,
  {ICLR} 2020, Addis Ababa, Ethiopia, April 26-30, 2020}. OpenReview.net, 2020.

\bibitem{nlvr2}
Alane Suhr, Stephanie Zhou, Ally Zhang, Iris Zhang, Huajun Bai, and Yoav Artzi.
\newblock A corpus for reasoning about natural language grounded in
  photographs.
\newblock In Anna Korhonen, David~R. Traum, and Llu{\'{\i}}s M{\`{a}}rquez,
  editors, {\em Proceedings of the 57th Conference of the Association for
  Computational Linguistics, {ACL} 2019, Florence, Italy, July 28- August 2,
  2019, Volume 1: Long Papers}, pages 6418--6428. Association for Computational
  Linguistics, 2019.

\bibitem{lxmert}
Hao Tan and Mohit Bansal.
\newblock {LXMERT}: Learning cross-modality encoder representations from
  transformers.
\newblock In Kentaro Inui, Jing Jiang, Vincent Ng, and Xiaojun Wan, editors,
  {\em Proceedings of the 2019 Conference on Empirical Methods in Natural
  Language Processing and the 9th International Joint Conference on Natural
  Language Processing, {EMNLP-IJCNLP} 2019, Hong Kong, China, November 3-7,
  2019}, pages 5099--5110. Association for Computational Linguistics, 2019.

\bibitem{deit}
Hugo Touvron, Matthieu Cord, Matthijs Douze, Francisco Massa, Alexandre
  Sablayrolles, and Herv{\'e} J{\'e}gou.
\newblock Training data-efficient image transformers \& distillation through
  attention.
\newblock {\em preprint arXiv:2012.12877}, 2020.

\bibitem{ofa}
Peng Wang, An Yang, Rui Men, Junyang Lin, Shuai Bai, Zhikang Li, Jianxin Ma,
  Chang Zhou, Jingren Zhou, and Hongxia Yang.
\newblock Unifying architectures, tasks, and modalities through a simple
  sequence-to-sequence learning framework.
\newblock {\em CoRR}, abs/2202.03052, 2022.

\bibitem{vlbeit}
Wenhui Wang, Hangbo Bao, Li Dong, Johan Bjorck, Zhiliang Peng, Qiang Liu, Kriti
  Aggarwal, Owais~Khan Mohammed, Saksham Singhal, Subhojit Som, et~al.
\newblock Image as a foreign language: Beit pretraining for all vision and
  vision-language tasks.
\newblock {\em arXiv preprint arXiv:2208.10442}, 2022.

\bibitem{simvlm}
Zirui Wang, Jiahui Yu, Adams~Wei Yu, Zihang Dai, Yulia Tsvetkov, and Yuan Cao.
\newblock {SimVLM}: Simple visual language model pretraining with weak
  supervision.
\newblock In {\em ICLR}, 2022.

\bibitem{mnli2017}
Adina Williams, Nikita Nangia, and Samuel Bowman.
\newblock A broad-coverage challenge corpus for sentence understanding through
  inference.
\newblock In {\em Proceedings of the 2018 Conference of the North {A}merican
  Chapter of the Association for Computational Linguistics: Human Language
  Technologies}, pages 1112--1122, New Orleans, Louisiana, 2018.

\bibitem{coca}
Jiahui Yu, Zirui Wang, Vijay Vasudevan, Legg Yeung, Mojtaba Seyedhosseini, and
  Yonghui Wu.
\newblock Coca: Contrastive captioners are image-text foundation models.
\newblock {\em CoRR}, abs/2205.01917, 2022.

\bibitem{florence}
Lu Yuan, Dongdong Chen, Yi{-}Ling Chen, Noel Codella, Xiyang Dai, Jianfeng Gao,
  Houdong Hu, Xuedong Huang, Boxin Li, Chunyuan Li, Ce Liu, Mengchen Liu,
  Zicheng Liu, Yumao Lu, Yu Shi, Lijuan Wang, Jianfeng Wang, Bin Xiao, Zhen
  Xiao, Jianwei Yang, Michael Zeng, Luowei Zhou, and Pengchuan Zhang.
\newblock Florence: {A} new foundation model for computer vision.
\newblock {\em CoRR}, abs/2111.11432, 2021.

\bibitem{vinvl}
Pengchuan Zhang, Xiujun Li, Xiaowei Hu, Jianwei Yang, Lei Zhang, Lijuan Wang,
  Yejin Choi, and Jianfeng Gao.
\newblock {VinVL}: Revisiting visual representations in vision-language models.
\newblock In {\em {IEEE} Conference on Computer Vision and Pattern Recognition,
  {CVPR} 2021, virtual, June 19-25, 2021}, pages 5579--5588. Computer Vision
  Foundation / {IEEE}, 2021.

\bibitem{expertchoice}
Yanqi Zhou, Tao Lei, Hanxiao Liu, Nan Du, Yanping Huang, Vincent~Y Zhao,
  Andrew~M Dai, Zhifeng Chen, Quoc~V Le, and James Laudon.
\newblock Mixture-of-experts with expert choice routing.
\newblock In {\em Advances in Neural Information Processing Systems}.

\bibitem{bookcorpus}
Yukun Zhu, Ryan Kiros, Rich Zemel, Ruslan Salakhutdinov, Raquel Urtasun,
  Antonio Torralba, and Sanja Fidler.
\newblock Aligning books and movies: Towards story-like visual explanations by
  watching movies and reading books.
\newblock In {\em Proceedings of the IEEE international conference on computer
  vision}, pages 19--27, 2015.

\bibitem{stmoe}
Barret Zoph, Irwan Bello, Sameer Kumar, Nan Du, Yanping Huang, Jeff Dean, Noam
  Shazeer, and William Fedus.
\newblock St-moe: Designing stable and transferable sparse expert models.
\newblock {\em arXiv preprint arXiv:2202.08906}, 2022.

\end{thebibliography}
}

\appendix
\section{Appendix}
\label{sec:appendix}

\subsection{Further Analyses}
\paragraph{``Dropped'' Tokens.}

\begin{figure*}[h!]
	\vspace{-0mm}\centering
	\begin{tabular}{c c c c}
		&
		\hspace{-40mm}
  \includegraphics[height=0.54cm]{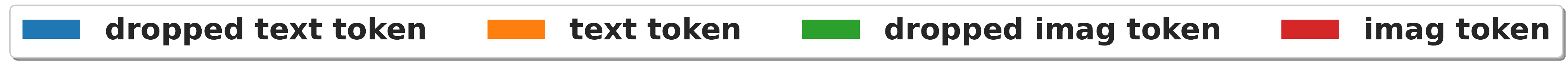}
  \hspace{-80mm}
  & & 
		 
		 \\
		\hspace{-3mm}
		\includegraphics[height=3.0cm]{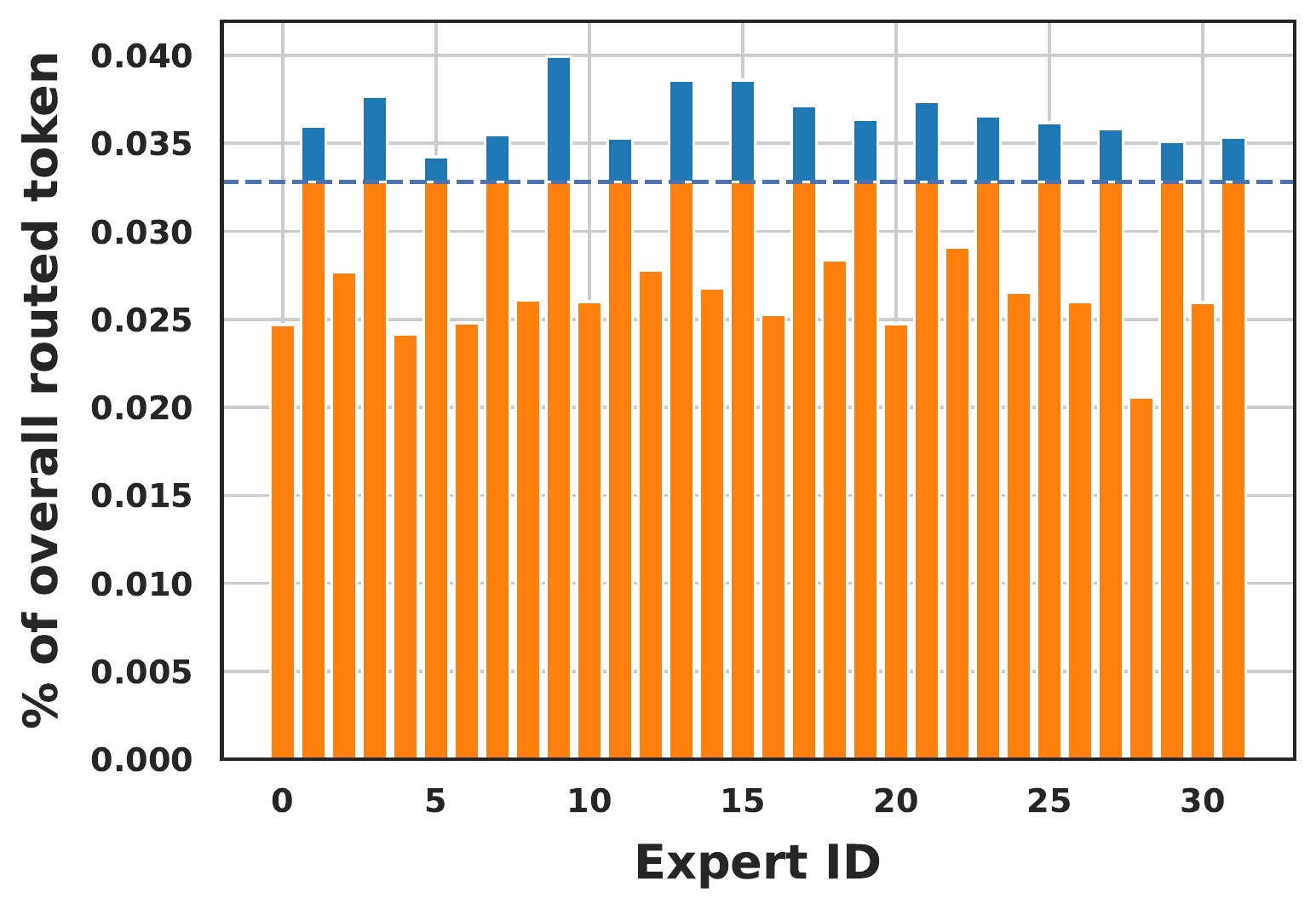}
		&
  \hspace{-4mm}
		\includegraphics[height=3.0cm]{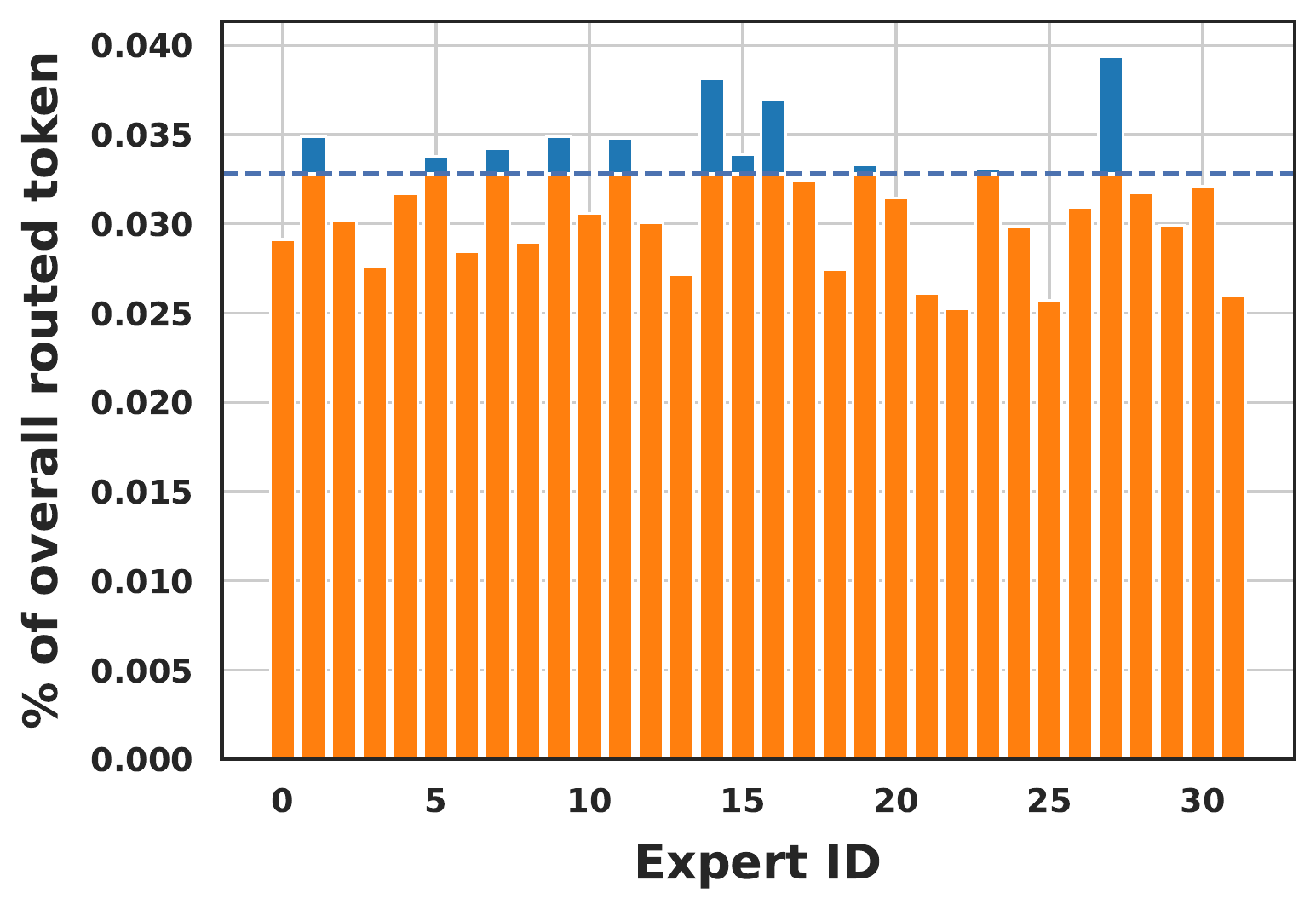}
 		&
  \hspace{-4mm}
		\includegraphics[height=3.0cm]{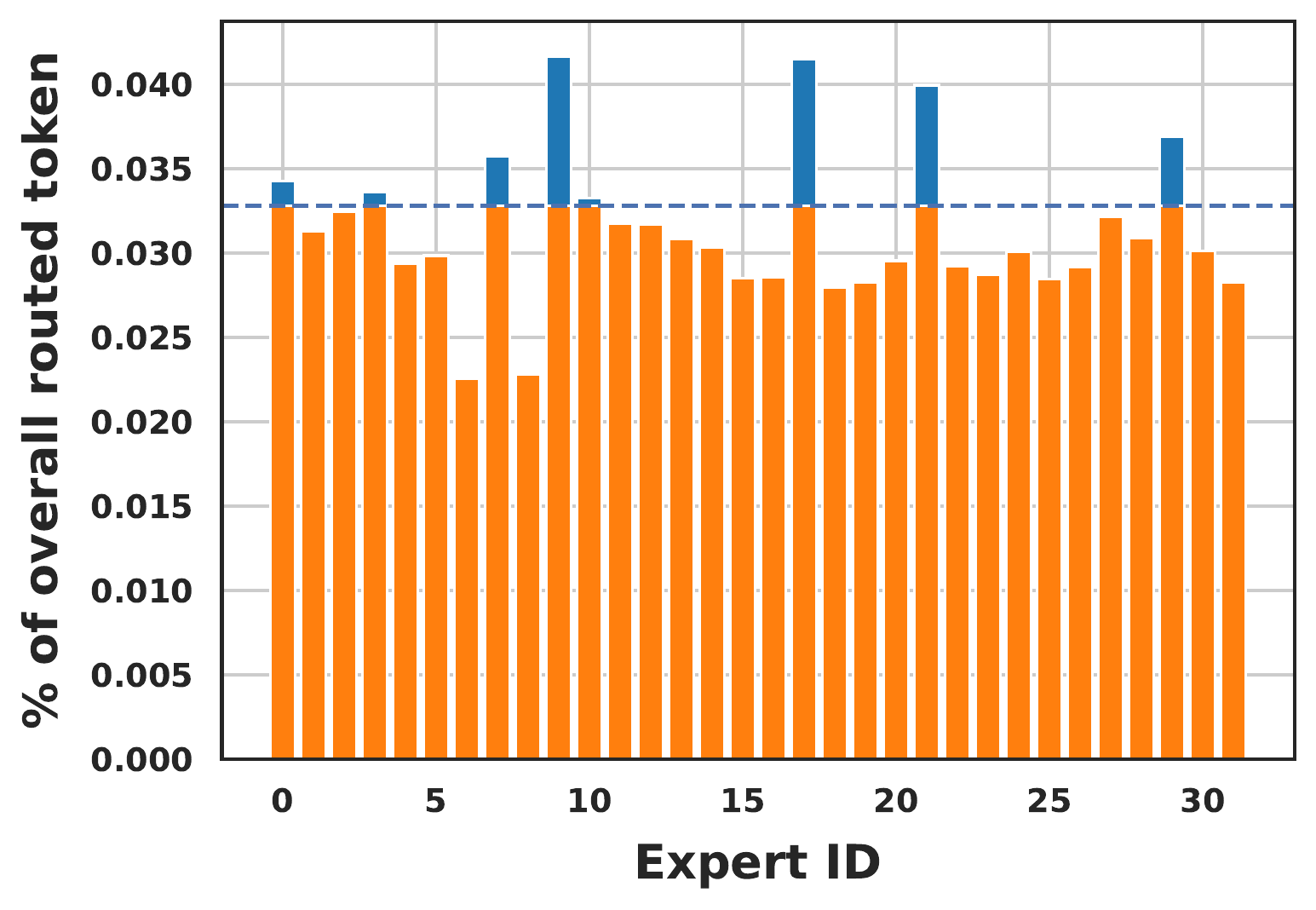} 
            &
            \hspace{-5mm}
  				\includegraphics[height=3.0cm]{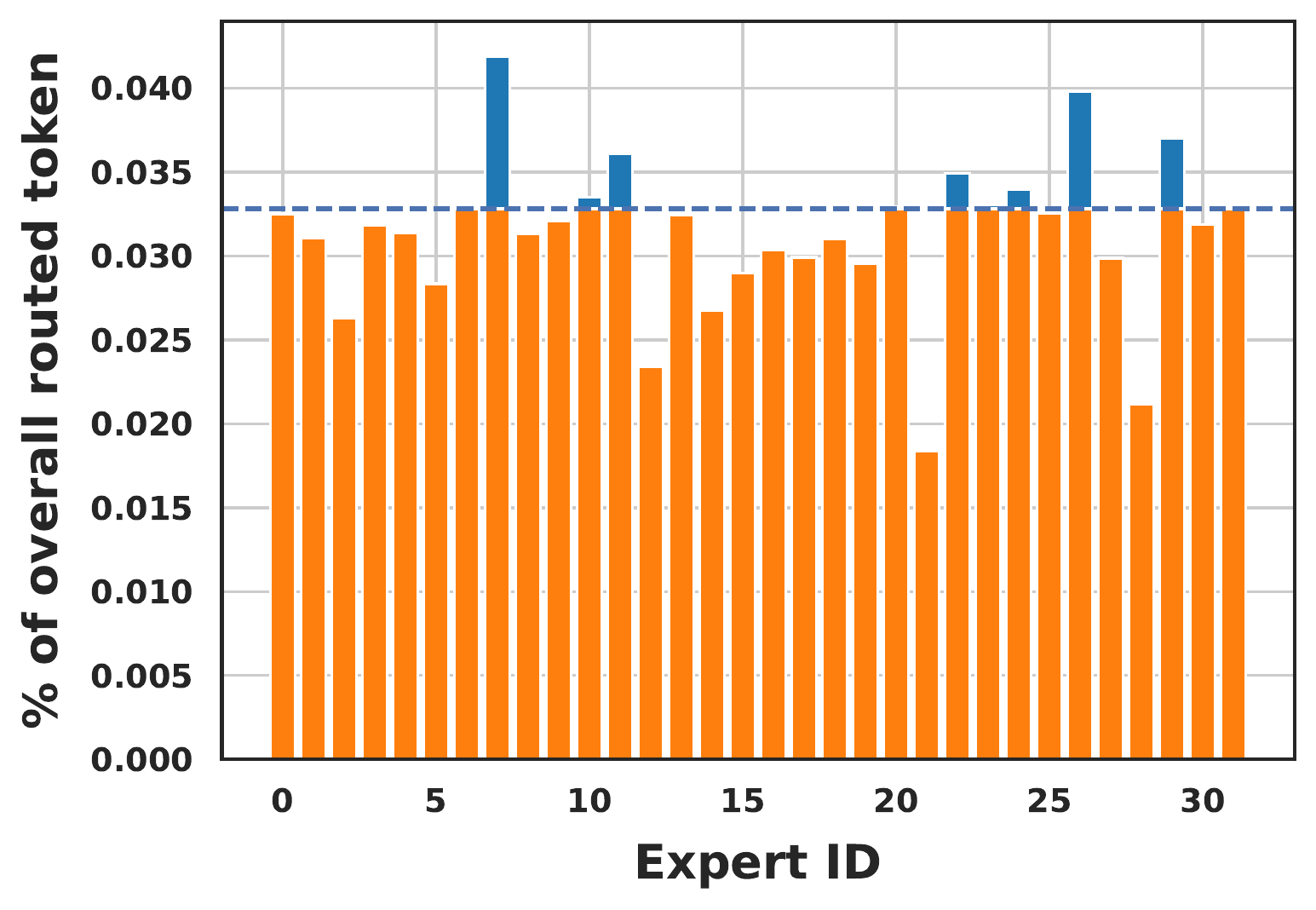} 
 \\
		(a) MLM T-MoE Layer 2 \vspace{2mm} 
		&
		(b) MLM T-MoE Layer 4 \vspace{2mm} 
		&
  		(c) MLM T-MoE Layer 6 \vspace{2mm} 
		&
		(d) MLM T-MoE Layer 8  \hspace{-0mm}  \\ 
		\hspace{-3mm}
		\includegraphics[height=3.0cm]{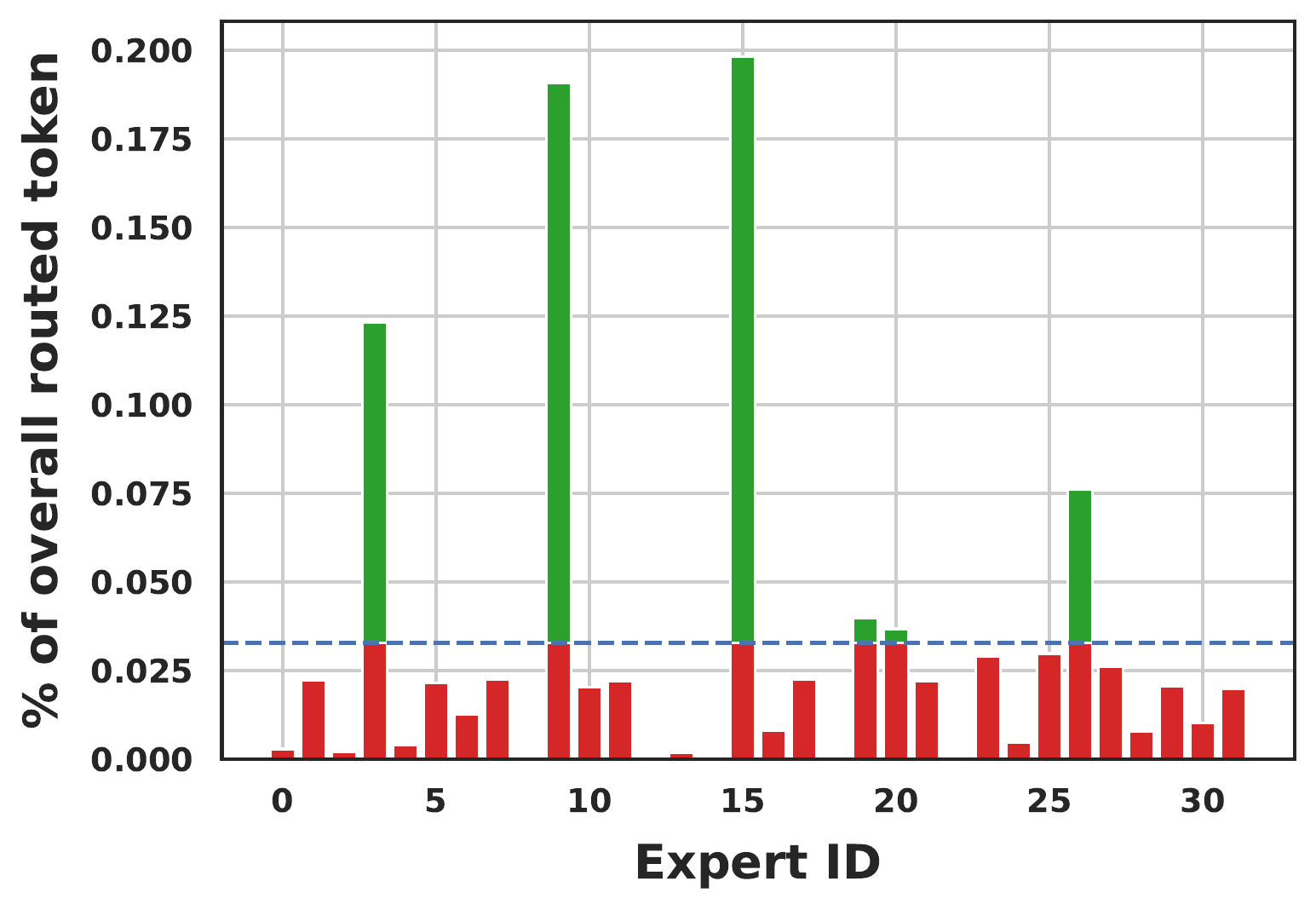}
		&
  \hspace{-4mm}
		\includegraphics[height=3.0cm]{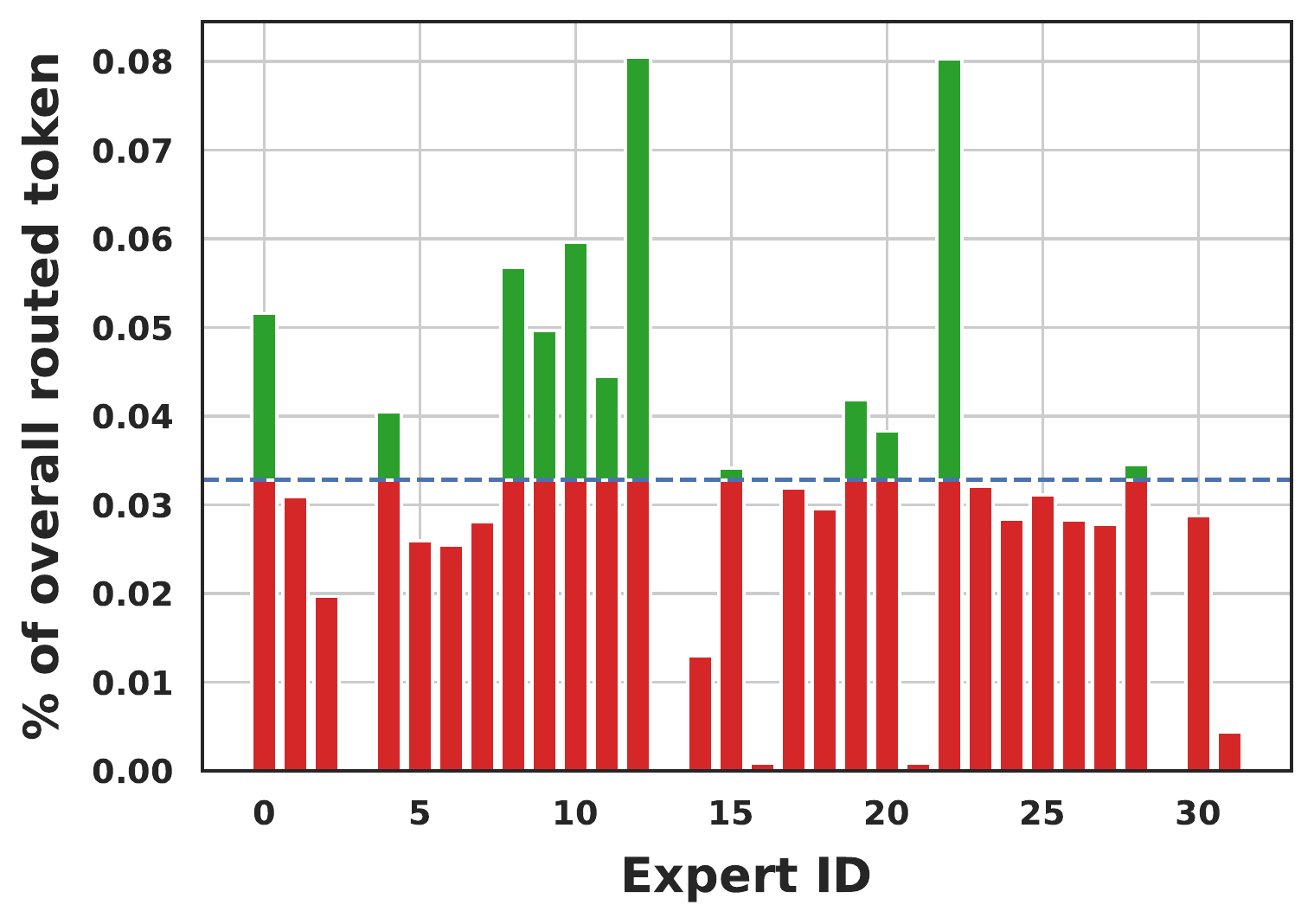}
 		&
  \hspace{-4mm}
		\includegraphics[height=3.0cm]{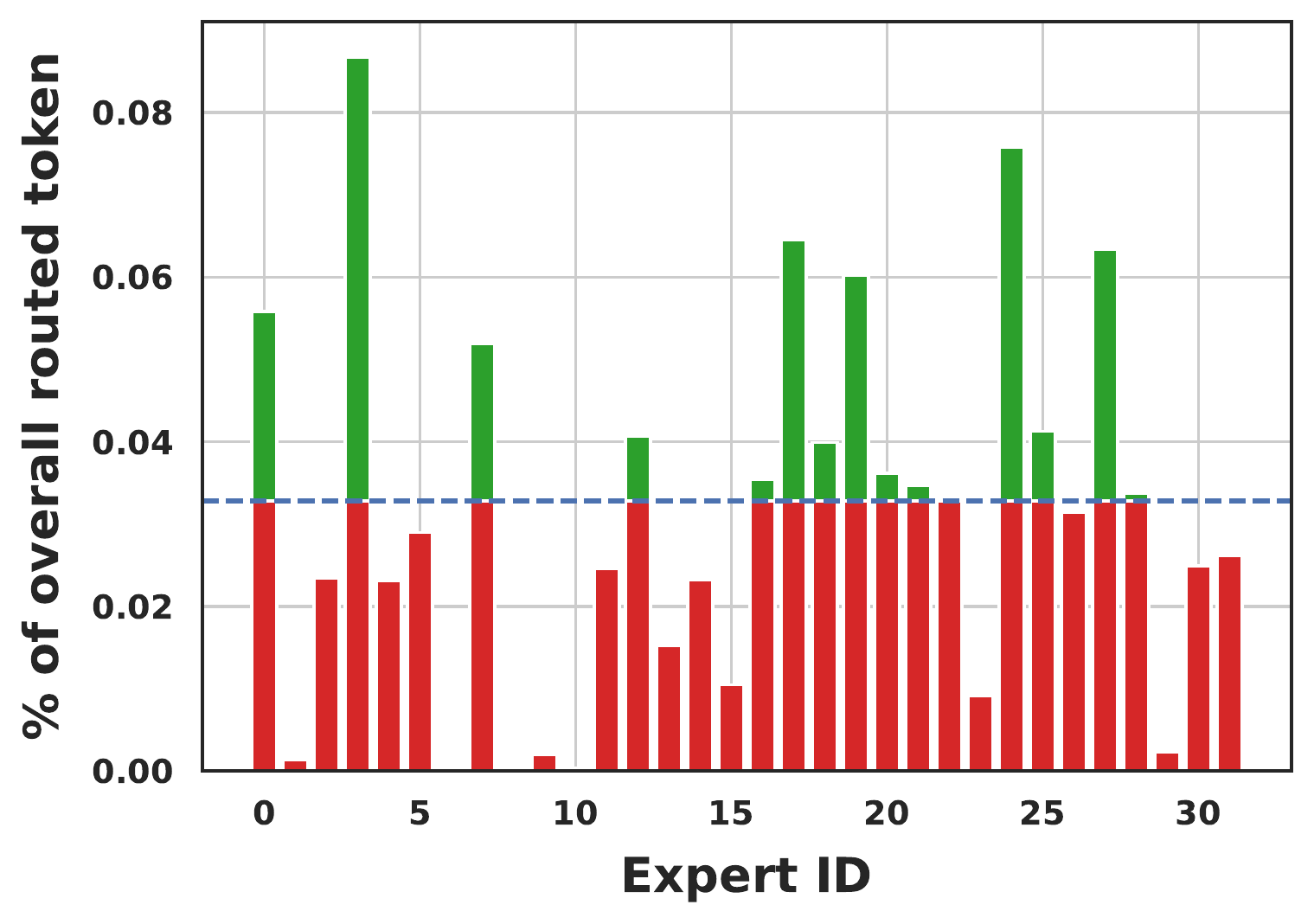} 
            &
            \hspace{-5mm}
  				\includegraphics[height=3.0cm]{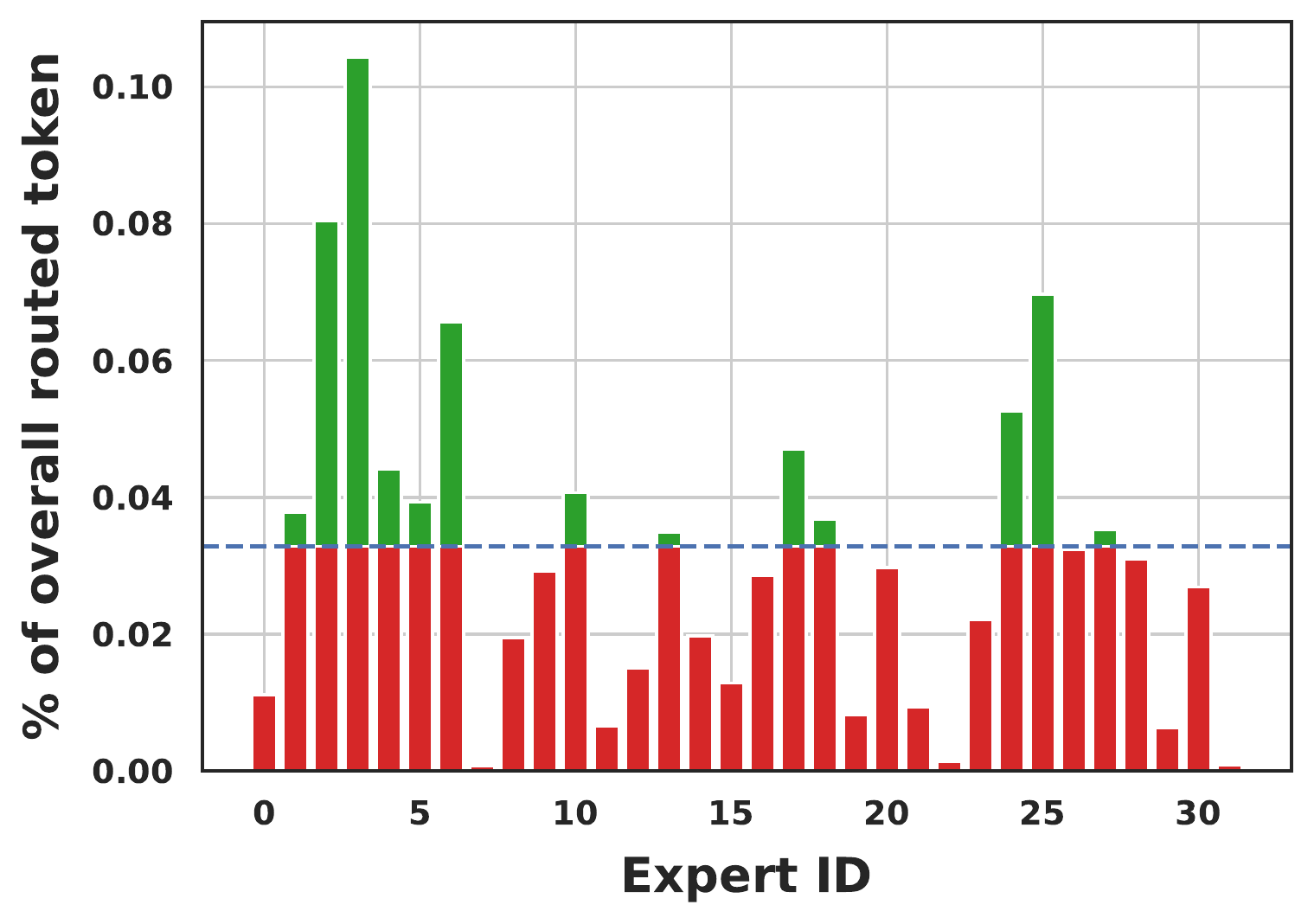} 
 \\
		(e) MIM V-MoE Layer 2 \vspace{2mm} 
		&
		(f) MIM V-MoE Layer 4 \vspace{2mm} 
		&
  		(g) MIM V-MoE Layer 6 \vspace{2mm} 
		&
		(h) MIM V-MoE Layer 8  \hspace{-0mm}  \\
		\hspace{-3mm}
		\includegraphics[height=3.0cm]{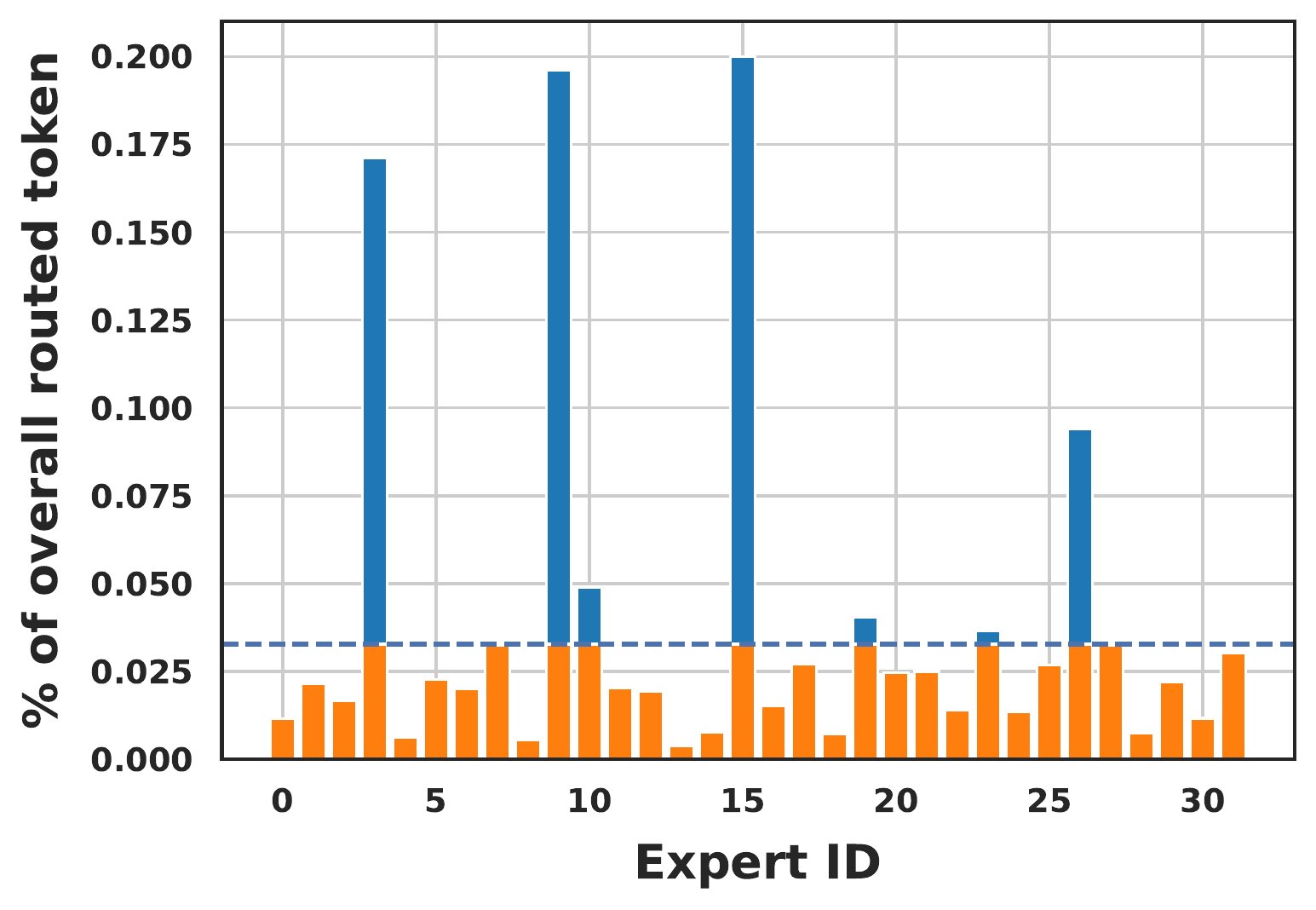}
		&
  \hspace{-4mm}
		\includegraphics[height=3.0cm]{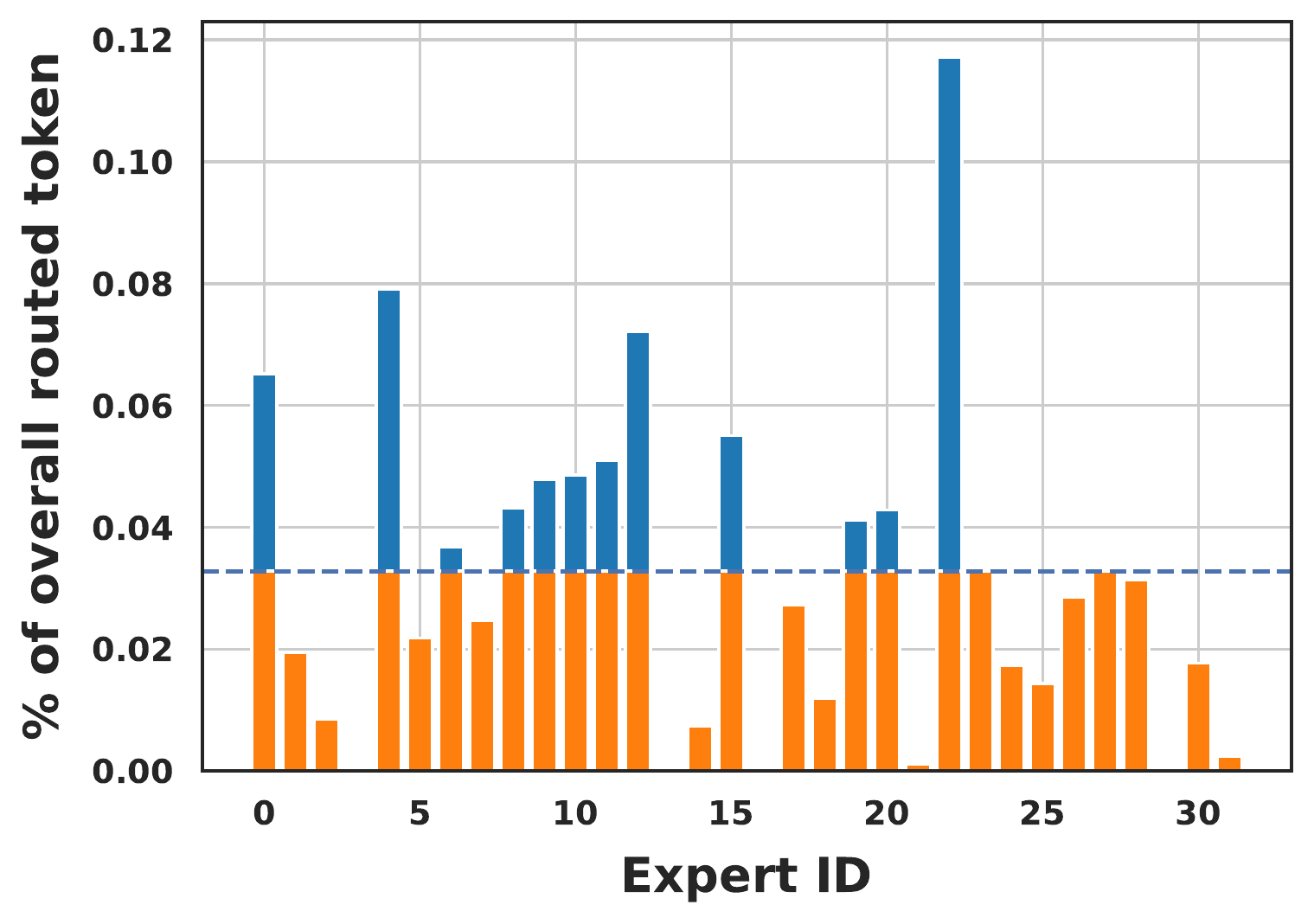}
 		&
  \hspace{-4mm}
		\includegraphics[height=3.0cm]{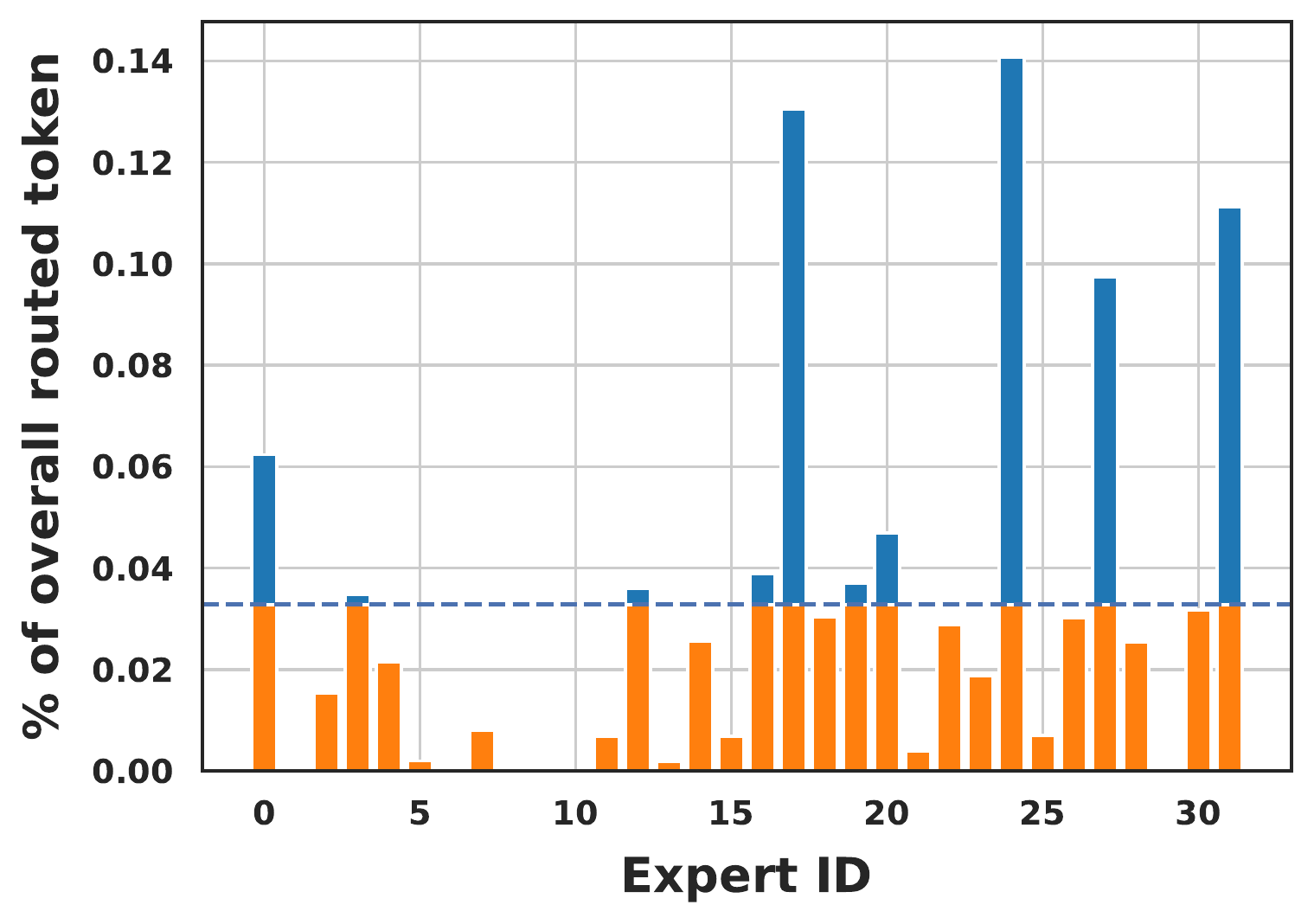} 
            &
            \hspace{-5mm}
  				\includegraphics[height=3.0cm]{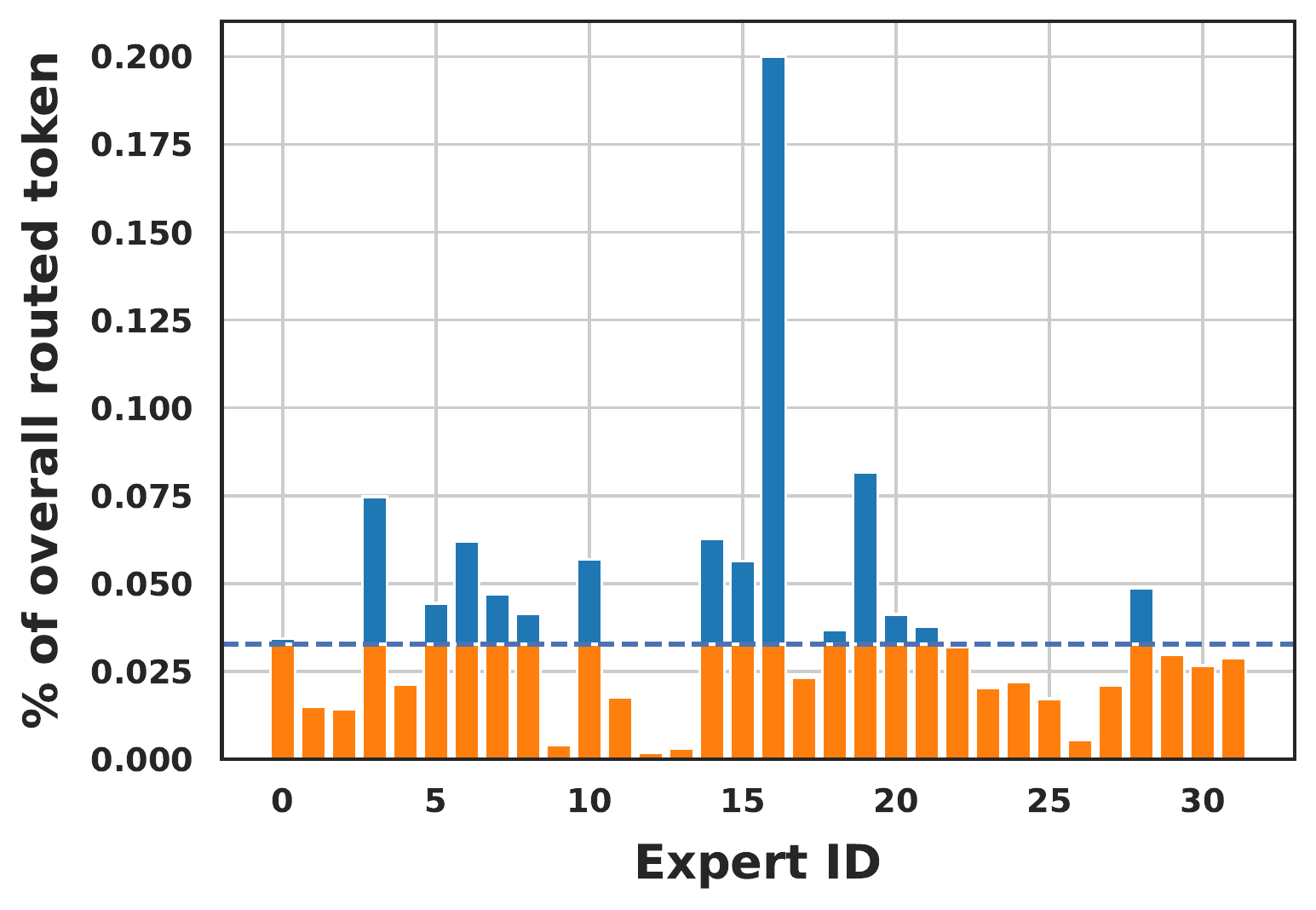} 
 \\
		(i) VLM T-MoE Layer 2 \vspace{2mm} 
		&
		(j) VLM T-MoE Layer 4 \vspace{2mm} 
		&
  		(k) VLM T-MoE Layer 6 \vspace{2mm} 
		&
		(l) VLM T-MoE Layer 8  \hspace{-0mm}  \\ 
		\hspace{-3mm}
		\includegraphics[height=3.0cm]{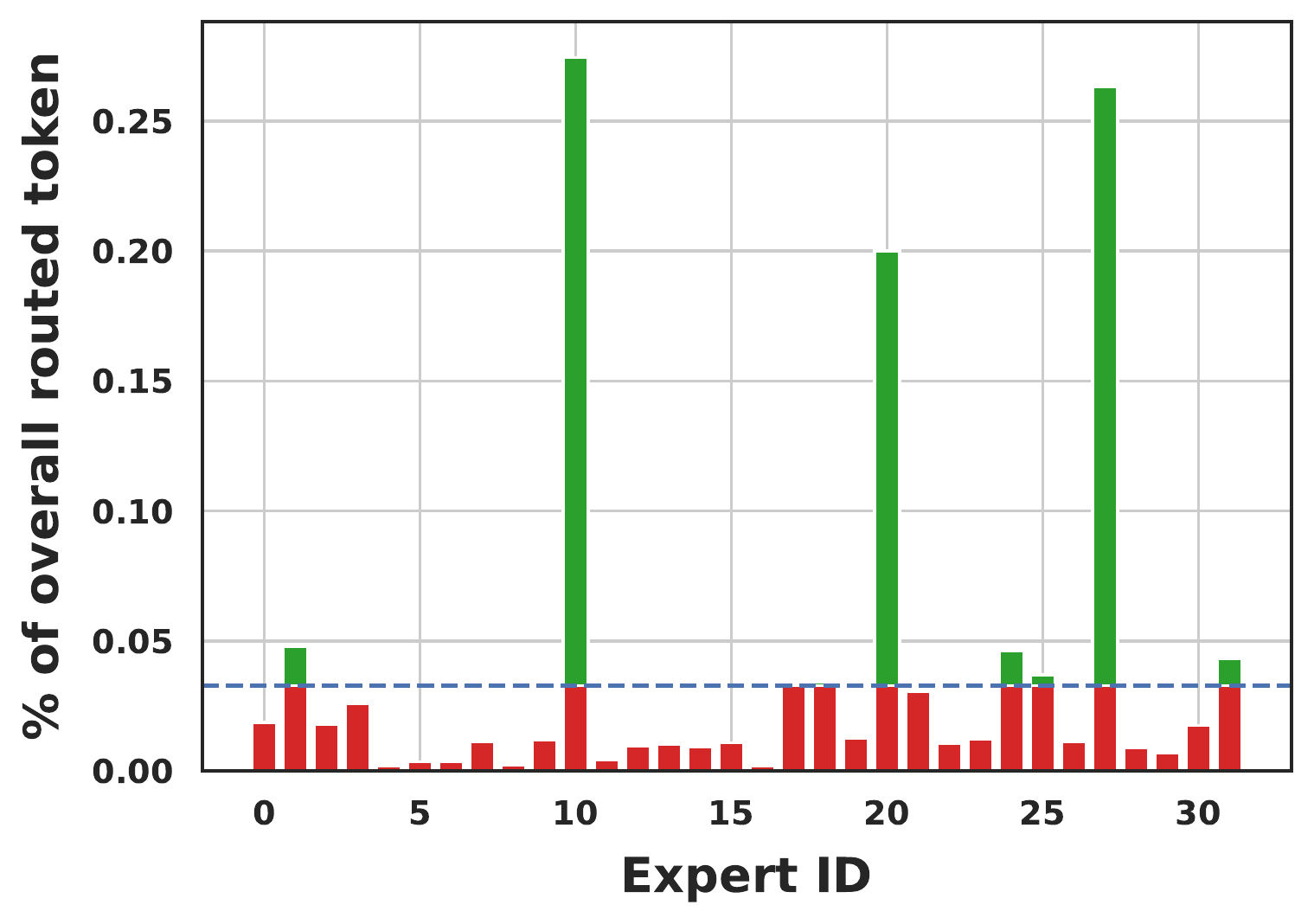}
		&
  \hspace{-4mm}
		\includegraphics[height=3.0cm]{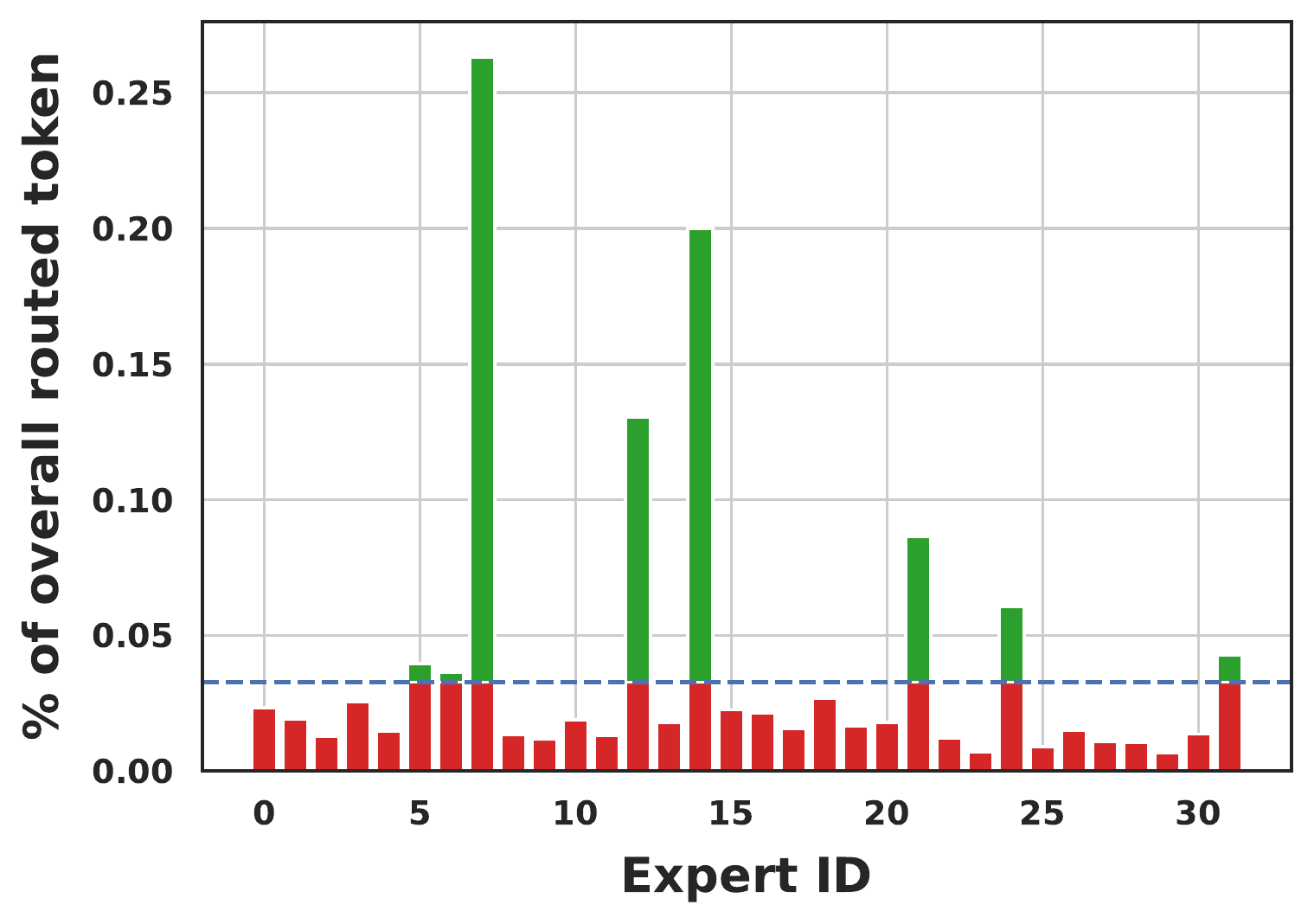}
 		&
  \hspace{-4mm}
		\includegraphics[height=3.0cm]{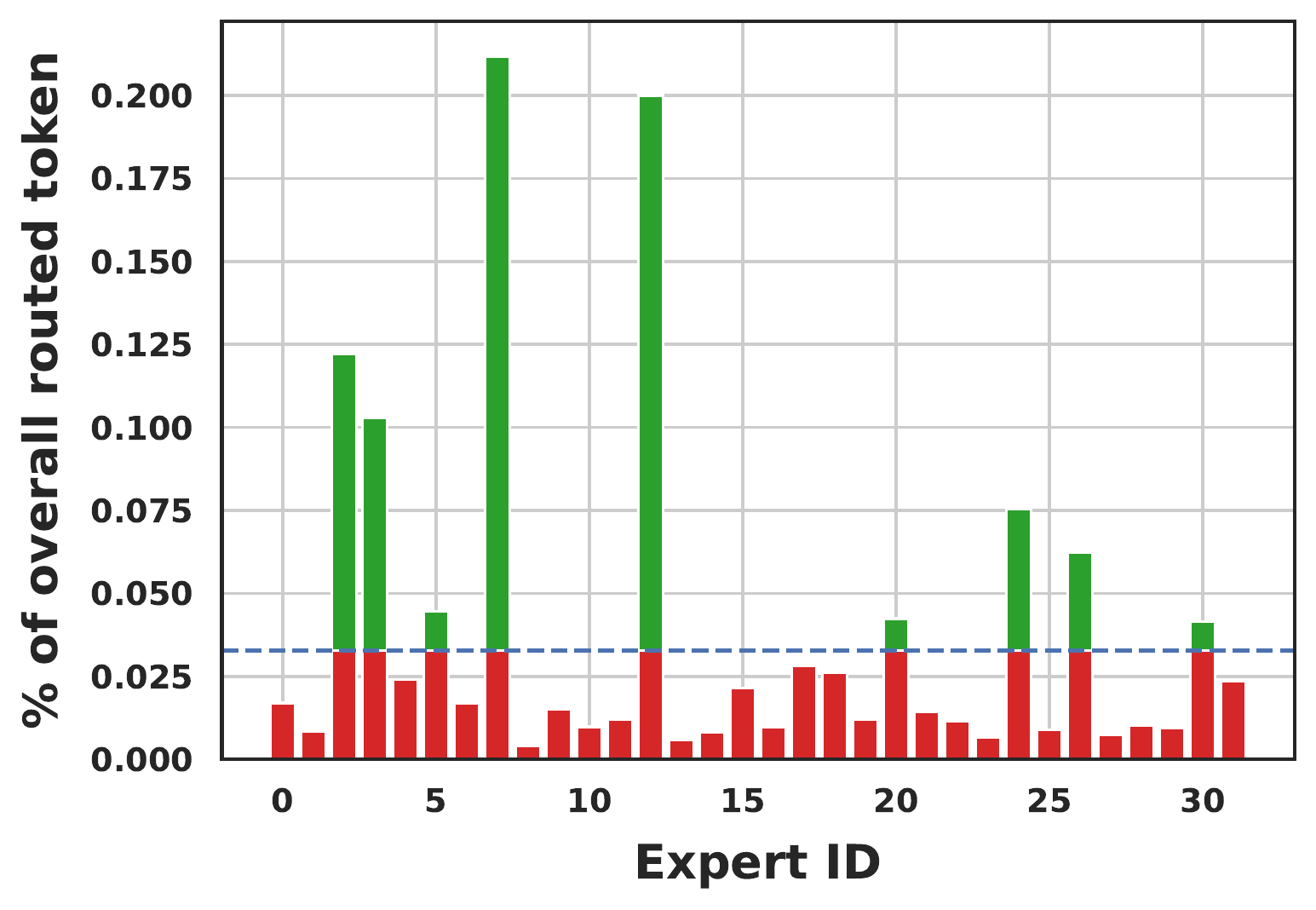} 
            &
            \hspace{-5mm}
  				\includegraphics[height=3.0cm]{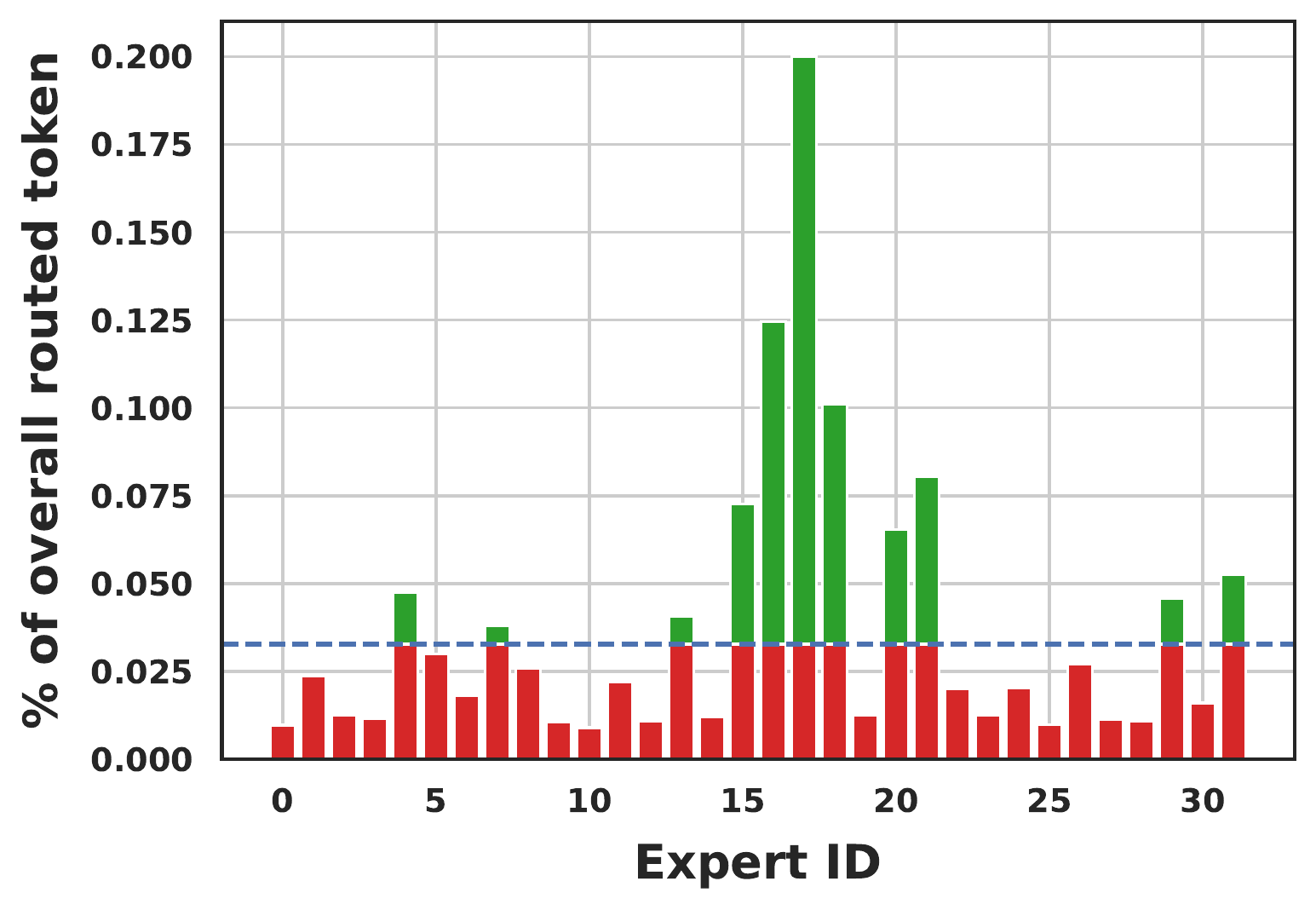} 
 \\
		(m) VLM V-MoE Layer 2 \vspace{2mm} 
		&
		(n) VLM V-MoE Layer 4 \vspace{2mm} 
		&
  		(o) VLM V-MoE Layer 6 \vspace{2mm} 
		&
		(p) VLM V-MoE Layer 8  \hspace{-0mm}  \\
		
	\end{tabular}
	\vspace{-6mm}
	\caption{``Dropped'' Token analyses for \ours{}$_\textsc{base/32E}$ with three mask language modeling (MLM), mask image modeling (MIM), and masked vision-language modeling (VLM) pre-training tasks. 
	Above the dashed line denotes the ratio of tokens that exceed the expert capacity and will be dropped. 
	 }
	\vspace{-0mm}
	\label{fig:drop_token}
\end{figure*}

In MoE training, the issue of "Dropped Tokens" is inherited~\cite{gshard,moe,limoe,vmoe,expertchoice} and caused by the limited capacity of each MoE expert, which can lead to instability. To provide a detailed analysis of this issue, we present Figure~\ref{fig:drop_token}, which illustrates the distribution of dropped tokens in \ours{}$_\textsc{base/32E}$ across different pre-training tasks. The figure shows that MLM and MIM tasks exhibit a more balanced distribution of tokens compared to VLM task, which may explain the improved performance of using MoEs in the former two pre-training tasks, as depicted in Figure~\ref{fig:train_loss_beit}. Additionally, the problem of dropped imag tokens is more severe compared to dropped text tokens, which aligns with the results of different scaling strategies presented in Section~\ref{sec:ablation} and the findings in~\cite{limoe,vmoe}. 

\paragraph{Pretrain Losses for Different Scaling Strategies.}

\begin{figure*}[h!]
	\vspace{-0mm}\centering
	\begin{tabular}{c c c c}
		&
		\hspace{-48mm}
  \includegraphics[height=0.54cm]{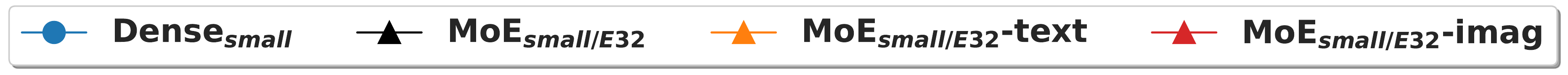}
  \hspace{-80mm}
  & & 
		 
		 \\
		\hspace{-3mm}
		\includegraphics[height=3.0cm]{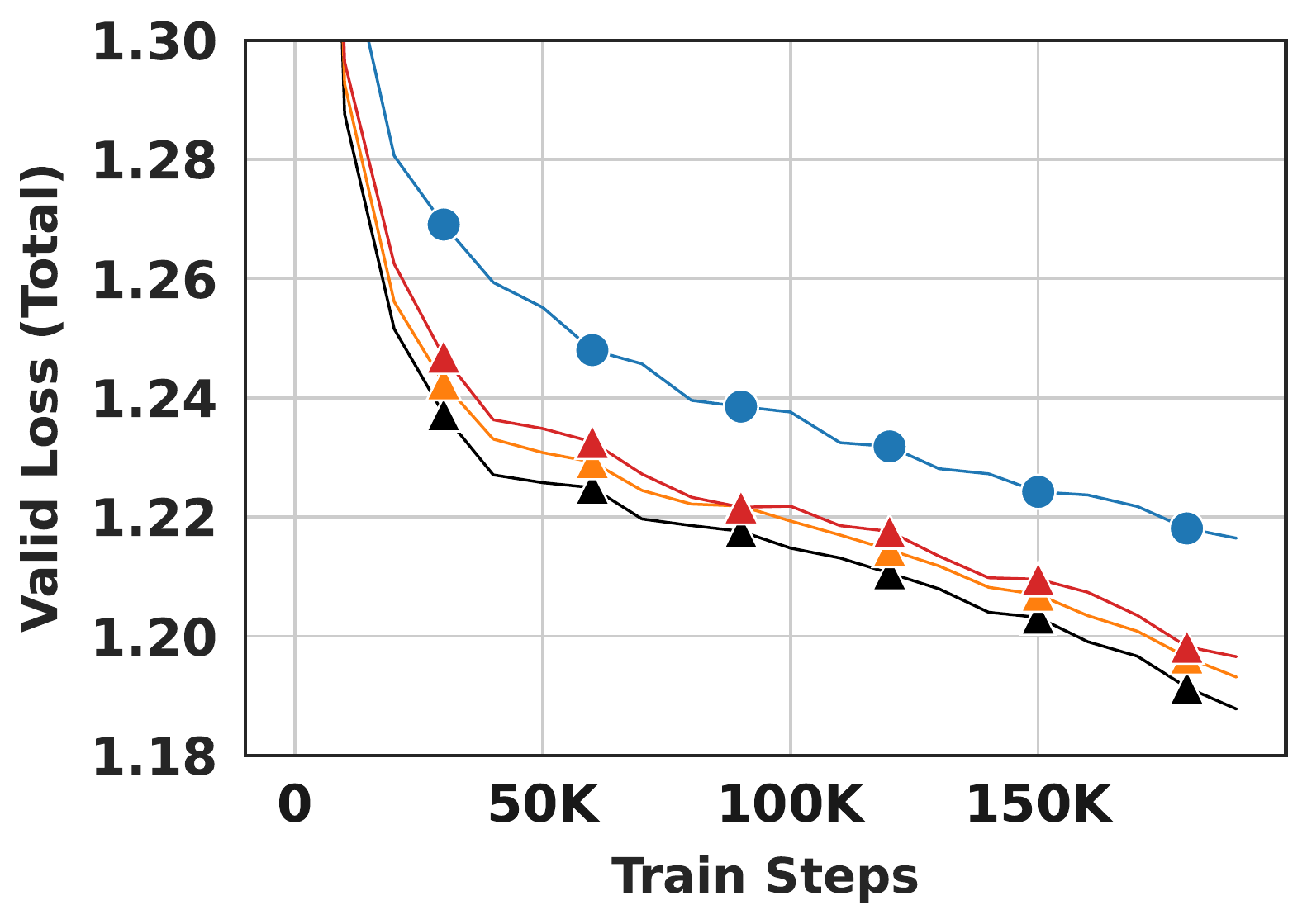}
		&
  \hspace{-4mm}
		\includegraphics[height=3.0cm]{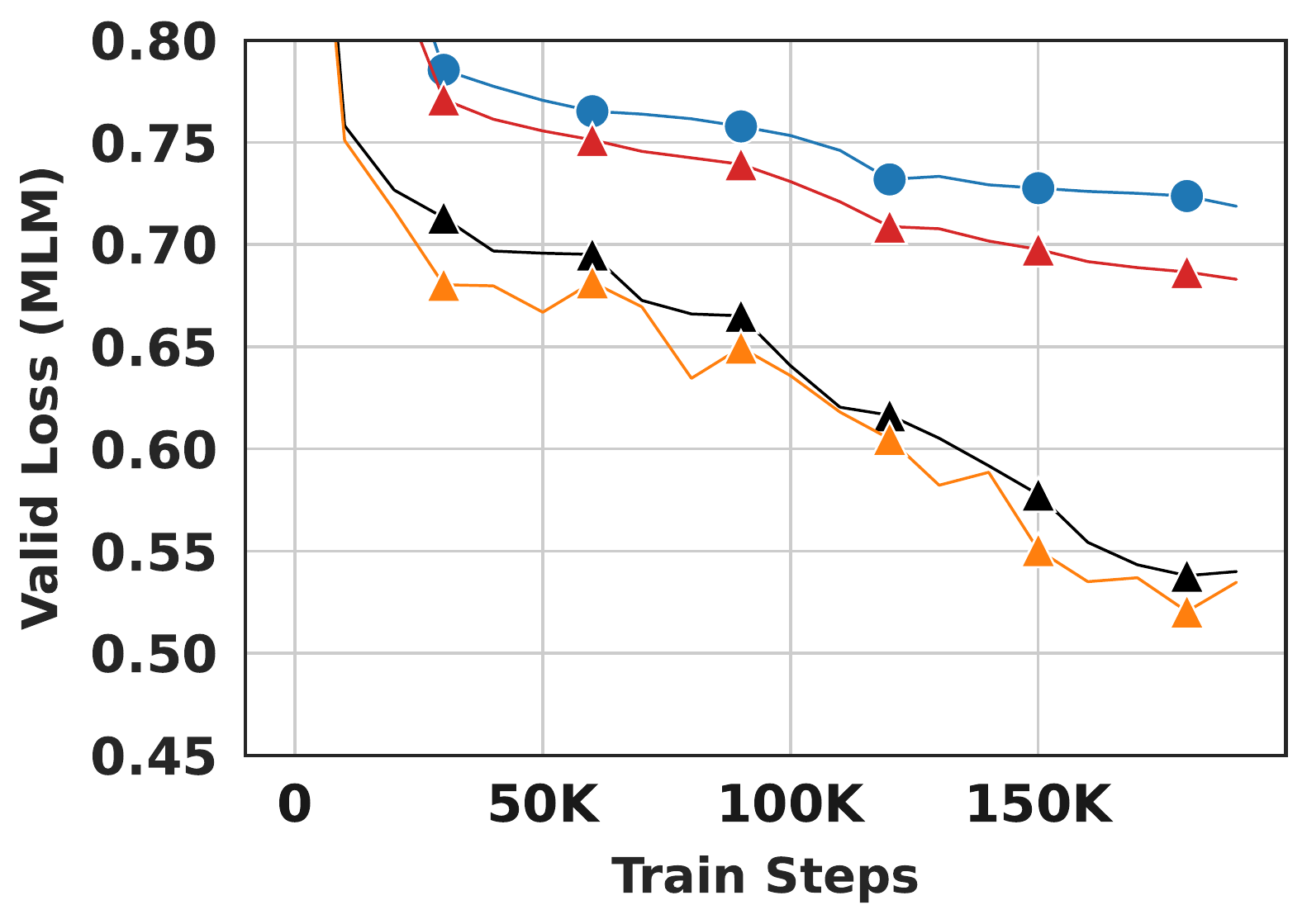}
 		&
  \hspace{-4mm}
		\includegraphics[height=3.0cm]{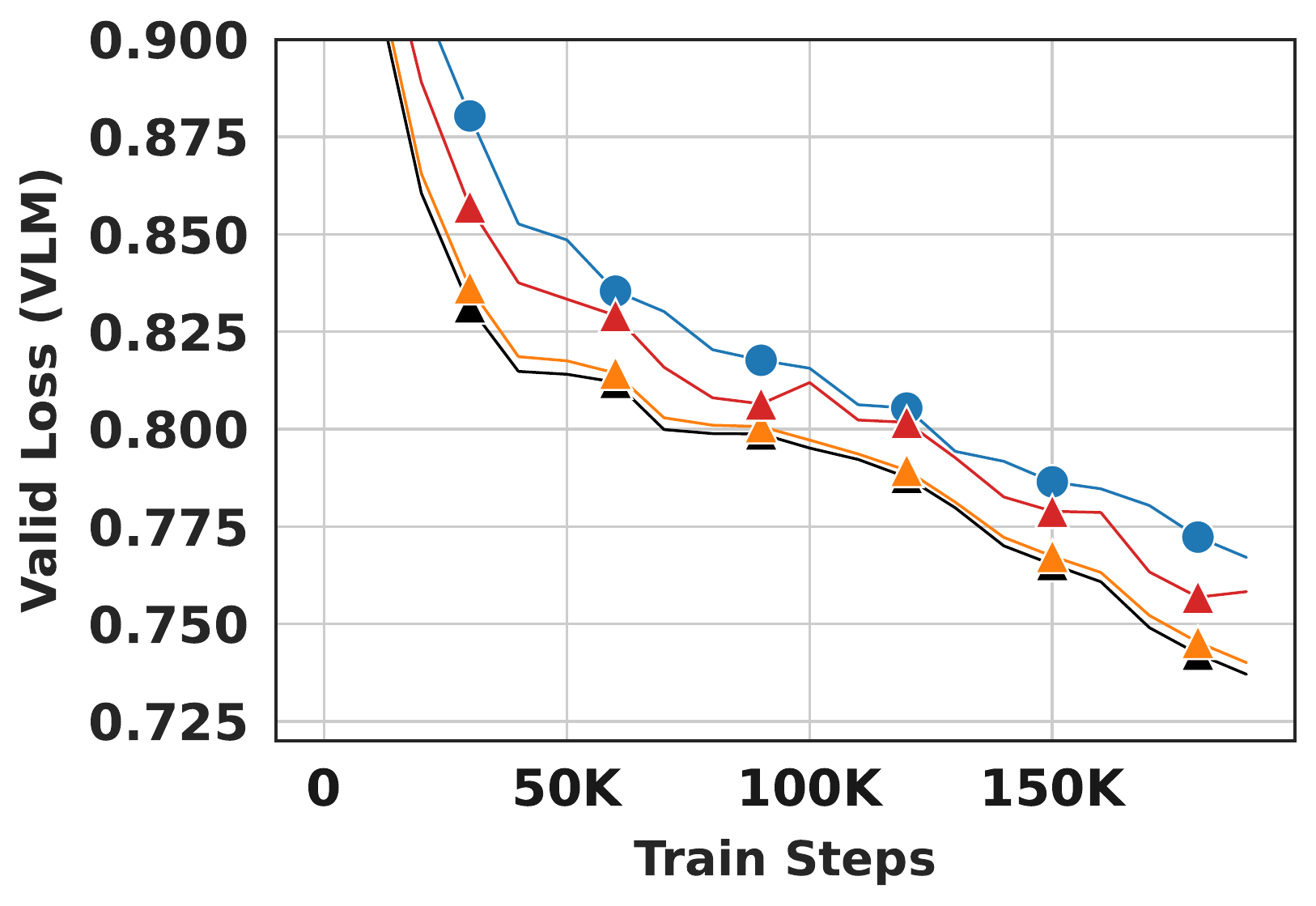} 
            &
            \hspace{-5mm}
  		\includegraphics[height=3.0cm]{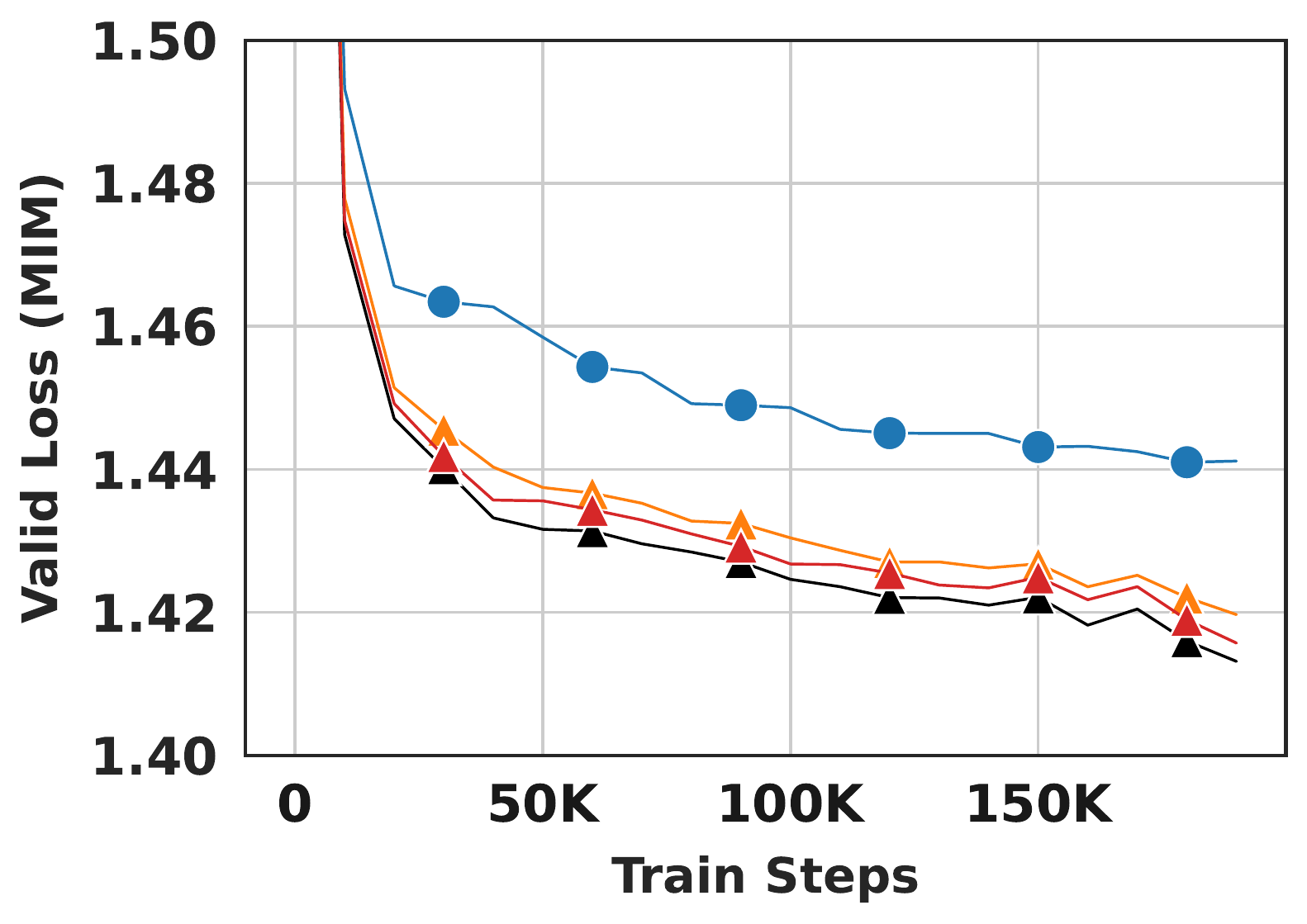} \\
		(a) Total Validation Loss \vspace{2mm} 
		&
		(b) MLM Validation Loss \vspace{2mm} 
		&
  		(c) VLM Validation Loss \vspace{2mm} 
		&
		(d) MIM Validation Loss \hspace{-0mm}  \\
	\end{tabular}
	\vspace{-6mm}
	\caption{Effect of different scaling strategy in Section~\ref{sec:ablation} for \ours{}$_\textsc{small/32E}$ scaling on three mask language modeling (MLM), mask image modeling (MIM), and masked vision-language modeling (VLM) pre-training tasks across training steps. 
	 }
	\vspace{-0mm}
	\label{fig:train_loss_beit_small_scaling}
\end{figure*}

We additionaly report the effect of different scaling strategy in Section~\ref{sec:ablation} for \ours{}$_\textsc{small/32E}$ scaling on three mask language modeling (MLM), mask image modeling (MIM), and masked vision-language modeling (VLM) pre-training tasks across training steps in Figure~\ref{fig:train_loss_beit_small_scaling}. 
The results support our hypothesis that using three distinct pretraining objectives for each modality and scaling each modality leads to improved optimization of both the specific modality pretraining loss and the VLM loss. 

\paragraph{Comparision with \limoe.} 
\begin{figure}[t]
    \centering
    \begin{subfigure}{.24\textwidth}
        \centering
        \includegraphics[width=\textwidth]{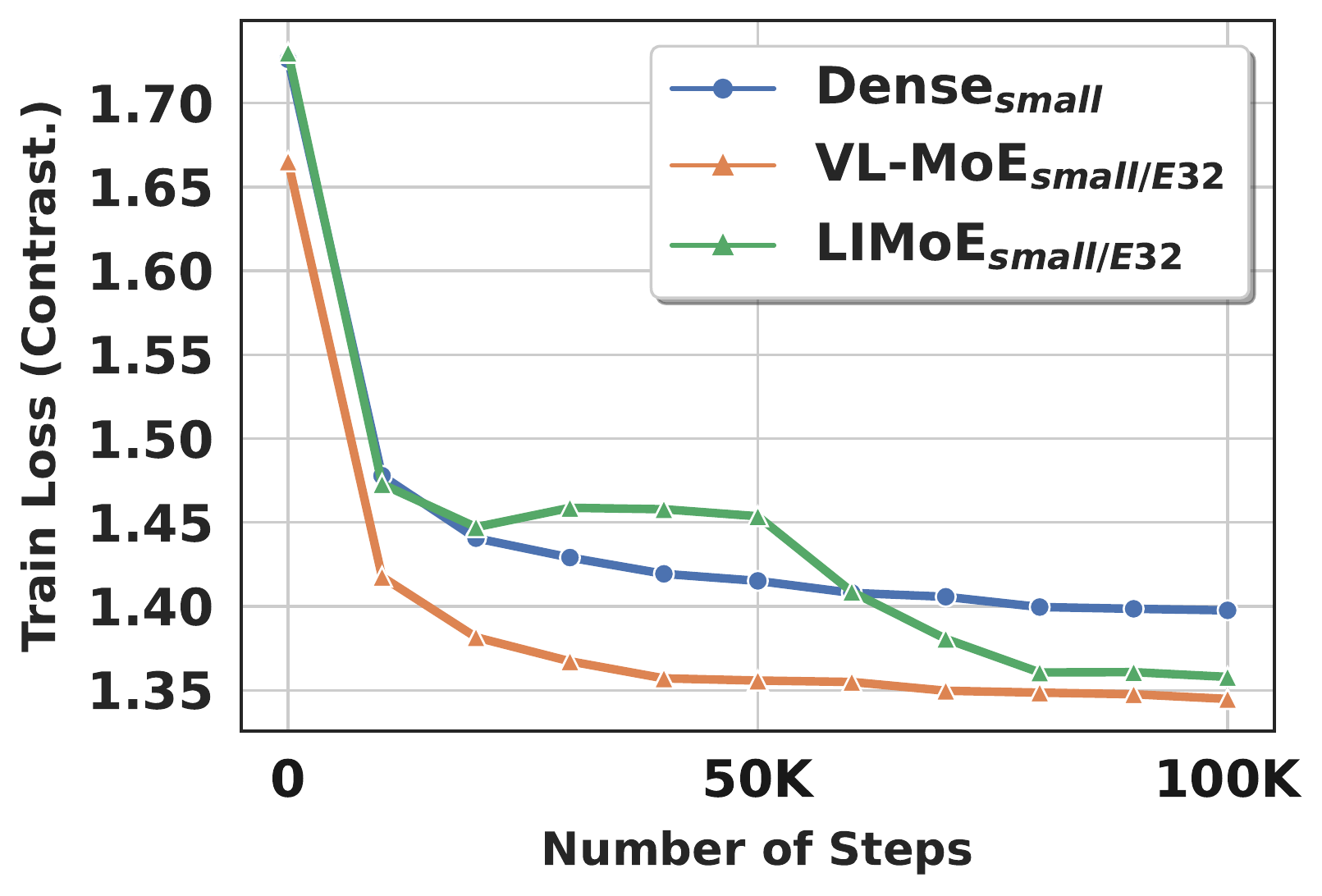}
        \caption{
        Contrastive Train Loss}
    \end{subfigure}%
    \begin{subfigure}{.24\textwidth}
        \centering
        \includegraphics[width=\textwidth]{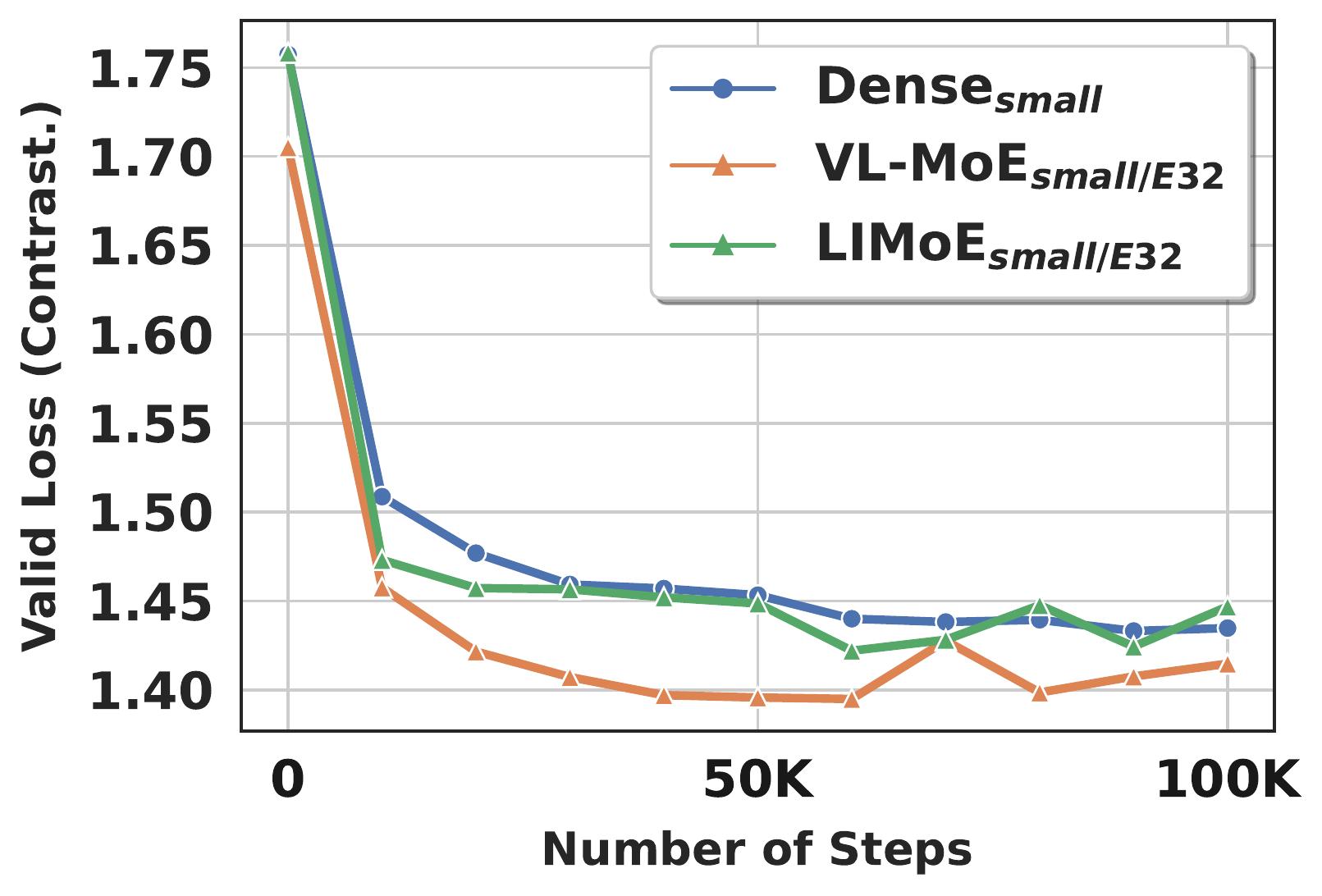}
        \caption{Contrastive Validation Loss }
    \end{subfigure}
    \caption{
    Comparision of Dense, \ours{}, and \limoe{} on contrastive pre-training task across training steps. 
    }
    \label{fig:train_loss_limoe}
\end{figure}

In \limoe~\cite{limoe}, the single-modality MoE architecture and the employed contrastive loss are the two main building blocks. To directly compare the two components of multimodal \limoe{} under our setting, we thoroughly experimented with optimizing either the single-modality MoE architecture or \ours{} with contrastive or masked data modeling (MDM) loss. However, we found that the models fail to converge when optimizing the \limoe{} architecture with the MDM loss, likely due to the fact that the MDM losses consist of three losses aiming for different modalities, which may exacerbate the modality imbalance problem and make it difficult to optimize MoEs even equipped with the entropy balancing loss in~\cite{limoe}.

Therefore, we focused on optimizing \ours{} and \limoe{} with the contrastive loss, as it yielded more stable results. However, it should be noted that while \limoe{} uses $1.8$B image-text pairs, our setting only has $4$M. 
We then report the training and validation loss across steps by optimizing \ours{} or \limoe{} with the contrastive loss in Figure~\ref{fig:train_loss_limoe}. 
The batch size is set to be $2$k. 
It can be seen that both models quickly overfit to the 4M image-text pairs, but the single modality MoE architecture in~\limoe{} inherits more instability. 

\subsection{Hyperparameter}
\paragraph{Visual Question Answering (VQA).}
We fine-tune the base-size models for $10$ epochs with $128$ batch size.
The peak learning rate is 3e-5.
Following VLMO~\cite{vlmo}, the input image resolution is $480 \times 480$. 

\paragraph{Natural Language for Visual Reasoning (NLVR2).}

For results of Table~\ref{tbl:results:vl_tasks}, the base-size models are fine-tuned for $10$ epochs with $128$ batch size.
The peak learning rate of the base-size models is set to 5e-5.
The input image resolution is $384 \times 384$.
For ablation experiments, we fine-tune the models for $10$ epochs with $128$ batch size, and choose learning rates from \{5e-5, 1e-4\}.
The input image resolution is $224 \times 224$. All the ablation results of NLVR2 are averaged over $3$ runs.

\paragraph{COCO.}

We fine-tune the base-size model for $20$ epochs with $2048$ batch size.
The peak learning rate is 2e-5 and the input image resolution is $384 \times 384$.

\paragraph{Flickr30K.}

For results of Table~\ref{tbl:results:vl_tasks}, the base-size models are fine-tuned for $40$ epochs with a batch size of $2048$ and a peak learning rate of 1e-5. We use the fine-tuned model on COCO as the initialization. The input image resolution is $384 \times 384$.
For all ablation experiments, we fine-tune the models for $10$ epochs with $1024$ batch size.
The peak learning rate is set to 5e-5, and the input image resolution is $224 \times 224$.

\paragraph{ImageNet-1k.} We fine-tune the base-size \ours{} with V-MoE and V-FFN only for $15$ epochs with $2048$ batch size.
The peak learning rate is 3e-5 and the input image resolution is $384 \times 384$.

\paragraph{MNLI.} We fine-tune the base-size \ours{} with T-MoE and T-FFN only for $10$ epochs with $32$ batch size.
The peak learning rate is 3e-5.

\subsection{Formula of Auxiliary Loss}

Given a token $\vx \in \R^D$, we denote by $g(\vx) = \texttt{softmax}(\mW \vx) \in \R^E$ the gating weights across the $E$ experts, with $\mW \in \R^{E \times D}$ being the routing parameters. When we deal with a batch of multiple tokens $\{\vx_i\}_{i=1}^n$, we use the notation $\mX\in\R^{n \times D}$.

\textbf{Importance loss.} We follow the definition from~\cite{vmoe,limoe}. 
The importance loss $\Omega_\text{imp}$ ensures that the gating weights are evenly distributed among the experts, maintaining a balanced profile.  
For any expert $e \in \{1,\dots, E\}$, we have
$$
\text{imp}_e(\mX) = \sum_{\vx \in \mX} g(\vx)_e
$$
and the loss $\Omega_\text{imp}$ is defined via the squared coefficient of variation for $\text{imp}(\mX) = \{\text{imp}_e(\mX)\}_{e=1}^E$
$$
\Omega_\text{imp}(\mX) = \left(
\frac{\texttt{std}(\text{imp}(\mX))}{\texttt{mean}(\text{imp}(\mX))}
\right)^2.
$$

\textbf{Load loss.}
Like previously, we follow~\cite{vmoe}. 
We assume the gating weights $g_\text{noisy}(\vx)$ are obtained by perturbing the routing function with noise, i.e., $g_\text{noisy}(\vx) = \texttt{softmax}(\mW \vx + \varepsilon)$ with $\varepsilon \sim \mathcal{N}(\vzero, \sigma^2 \mI)$ and $\sigma=1/E$. 
We denote $\eta_k$ the $k$-th largest entry of $\mW \vx + \varepsilon$.
The importance loss $\Omega_\text{imp}$ aims to balance the selection probability of experts by focusing on the likelihood of choosing them, as assigning tasks to experts is a discrete process. The load loss $\Omega_\text{load}$ complements this by striving to even out the number of assignments among the experts. To calculate the selection probability, the expert $e \in \{1,\dots, E\}$ is assumed to be among the top-$k$ even when resampling only the noise as
$$
p_e(\vx) = 1 - \Phi\Big( \frac{\eta_k - (\mW \vx)_e}{\sigma}  \Big) 
$$
with $\Phi$ the cumulative distribution function of a Gaussian distribution.
The load loss $\Omega_\text{load}$ is eventually defined by
\begin{align}
&\Omega_\text{load}(\mX) = \left(
\frac{\texttt{std}(\text{load}(\mX))}{\texttt{mean}(\text{load}(\mX))}
\right)^2 \nonumber
 \\
\text{where}
\ 
\text{load}(\mX) &= \{\text{load}_e(\mX)\}_{e=1}^E
\ 
, 
\ 
\text{load}_e(\mX) = \sum_{\vx \in \mX} p_e(\vx). \nonumber
\end{align}

\textbf{Z-loss.} The z-loss $\Omega_{\text{zloss}}$ introduced in~\cite{stmoe} aims at controlling the maximum magnitude of the router activations $\mA = \{\mW \vx_i\}_{i=1}^n \in\R^{n \times E}$ with entries $a_{i,e} = (\mW \vx_i)_e$. The loss is defined by
$$
\Omega_{\text{zloss}}(\mX) = \frac{1}{n} \sum_{i=1}^n \left( \log\left(\sum_{e=1}^E \exp{(a_{i,e})} \right)  \right)^2.
$$

\textbf{v-loss.} The notation ``v-loss'' we used in Section~\ref{sec:ablation} is essentially the final employed loss in \vmoe~\cite{vmoe}, where $\Omega_{\text{vloss}}(\mX) = 0.5*\Omega_{\text{imp}}(\mX) + 0.5*\Omega_{\text{load}}(\mX)$.

\end{document}